\documentclass{article}

 \usepackage[main, final]{neurips_2026}

\usepackage[utf8]{inputenc} %
\usepackage[T1]{fontenc}    %
\usepackage{hyperref}       %
\usepackage{url}            %
\usepackage{booktabs}       %
\usepackage{amsfonts}       %
\usepackage{nicefrac}       %
\usepackage{microtype}      %
\usepackage{xcolor}         %

\usepackage{amsmath}
\usepackage{amssymb}
\usepackage{amsthm}
\usepackage{mathtools}
\usepackage{bm}
 
\theoremstyle{plain}
\newtheorem{theorem}{Theorem}
\newtheorem{proposition}[theorem]{Proposition}
\newtheorem{lemma}[theorem]{Lemma}
\newtheorem{corollary}[theorem]{Corollary}
\theoremstyle{definition}

\newtheorem{assumption}[theorem]{Assumption}
\theoremstyle{remark}
\newtheorem{remark}[theorem]{Remark}
 
\newcommand{\Real}{\mathbb{R}}
\newcommand{\Pop}{\mathcal{P}}
\newcommand{\Hcal}{\mathcal{H}}
\newcommand{\Tcal}{\mathcal{T}}
\newcommand{\Zcal}{\mathcal{Z}}
\newcommand{\Scal}{\mathcal{S}}
\newcommand{\Aenc}{A^{\mathrm{enc}}}
\newcommand{\Sopt}{S^{\star}}
\newcommand{\Rpop}{R^{\Pop}_\Hcal}
\newcommand{\Rpopt}{R^{\Pop}_\Tcal}
\newcommand{\Remp}{\widehat{R}_\Hcal}
\newcommand{\Rempt}{\widehat{R}_\Tcal}
\newcommand{\Rdag}{R^{\Pop,\dagger}_\Hcal}
\newcommand{\Rempdag}{\widehat{R}^{\dagger}_\Tcal}
\newcommand{\Shat}{\widehat{S}}
\newcommand{\GammaH}{\Gamma_\Hcal}
\newcommand{\GammaT}{\Gamma_\Tcal}
\newcommand{\GammaTH}{\Gamma_{\Tcal\Hcal}}
\newcommand{\eapp}{\varepsilon_{\mathrm{app}}}
\newcommand{\eopt}{\varepsilon_{\mathrm{opt}}}
\newcommand{\Vstat}{\mathcal{V}^{\mathrm{stat}}_n}
\newcommand{\En}{\mathcal{E}_n}
\newcommand{\muB}{\mu_B}
\newcommand{\Deltas}{\Delta_s}
\newcommand{\score}[1]{s_{#1}}
\newcommand{\Otilde}{\widetilde{\mathcal{O}}}
\newcommand{\Wmax}{W_{\max}}
\newcommand{\bhat}{\widehat{b}}
\newcommand{\What}{\widehat{W}}
\newcommand{\Bmodel}{B_{\mathrm{model}}}
\newcommand{\indic}{\mathbf{1}}
\DeclareMathOperator{\logit}{logit}
\DeclareMathOperator{\Ber}{Ber}
\DeclareMathOperator{\Bin}{Bin}
\DeclareMathOperator{\rank}{rank}
\DeclareMathOperator{\op}{op}
\providecommand{\E}{\mathbb{E}}

\usepackage{enumitem}

\usepackage{longtable}
\usepackage{multirow}

\title{From Link Prediction to Linear Community Detection: A Finite-Sample Guarantee for Graph Pretraining}

\author{%
  Ying Chen \\
  Arizona State University \\
  \texttt{yche1422@asu.edu} \\
  \And
  Zhangyang Wang \\
  University of Texas at Austin \\
  \texttt{atlaswang@utexas.edu} \\
  \And
  Yixuan He\thanks{Corresponding author.} \\
  Arizona State University \\
  \texttt{Yixuan.He@asu.edu} \\
}

\begin{document}

\maketitle

\begin{abstract}

Link prediction (LP) is a standard pretraining objective for graph neural networks (GNNs): hide edges, train an encoder--decoder to predict them, and reuse the node embeddings for downstream node tasks. Despite its empirical success, there is little quantitative theory explaining why an edge-level objective should produce embeddings that are linearly useful for node classification. We study this question in the finite-sample stochastic block model (SBM). For any bounded GNN encoder paired with a symmetric bilinear decoder, we prove a conditional reduction: small population held-out LP excess risk implies small community-classification error for a transductive linear classifier on the learned embeddings. The bound decomposes into an optimization residual, an architectural approximation residual, and explicit statistical terms of order $\Otilde(\sqrt{d/n})$ in bounded dense regimes. We further give a constructive few-label classifier: nearest-centroid classification under the positive semidefinite quadratic form \(G = W Z^\top Z W\), whose label error is coupled to LP quality. The framework is architecture-agnostic through the approximation residual, and we instantiate this residual as \(O(\log n / n)\) for a graph isomorphism network (GIN) with spectral positional encoding and a bilinear decoder. Synthetic SBM experiments validate the theorem chain from LP risk to score error to classification error, and show when the worst-case constants become conservative.
\end{abstract}

\section{Introduction}

Link prediction (LP) is a standard self-supervised objective for graph representation learning: hide part of the edge set, train an encoder--decoder to predict those edges, and reuse the node embeddings for downstream tasks. This idea appears in graph autoencoders with inner-product decoders \citep{kipf2016variational}, in proximity-based unsupervised objectives such as GraphSAGE \citep{hamilton2017inductive}, and as a core component of generative graph pretraining frameworks such as GPT-GNN \citep{hu2020gptgnn}.  It is also central to graph foundation models and the pretrain--finetune/prompt paradigm \citep{liu2025gfmsurvey,sun2022gppt,liu2023graphprompt}, where LP can help align pretraining with downstream node tasks. Empirically, this recipe is effective even with few labels.

The theoretical picture is less clear. LP is trained on \emph{edges}, while the downstream task is defined over \emph{nodes}; the LP loss does not use community labels and does not define a linear decision boundary in the embedding space. This leads to our central question: \textbf{why should low LP loss imply that a linear classifier on the learned embeddings recovers communities, and how is the classification error controlled by LP-training quality?} Similar questions have been answered for contrastive self-supervised learning under latent-class or augmentation-graph assumptions \citep{arora2019theoretical,haochen2021provable}. To our knowledge, a finite-sample guarantee of this form is still missing for LP-based graph pretraining.

Any guarantee of this kind needs an assumption that links the graph to the node labels. LP never sees labels, so if the labels were independent of the graph, no classifier built on graph-derived embeddings could beat the trivial classifier, however well the embeddings predict edges. This is analogous to contrastive-learning theory, where positive pairs are assumed to share latent classes \citep{arora2019theoretical} or to lie in the same component of an augmentation graph \citep{haochen2021provable}. For graphs, the stochastic block model (SBM) \citep{holland1983stochastic,snijders1997estimation} is the canonical model of such a link: edge probabilities are determined by latent community memberships. It is also the standard setting of community-detection theory \citep{abbe2018community,lei2015consistency}. Therefore, in our analysis,  the SBM plays the role of the latent-class assumption. We study finite-sample SBMs with identifiable communities and bounded bilinear encoder--decoder models.

In this setting, our main result directly connects downstream classification error to LP risk. For masked LP pretraining with a bounded GNN encoder and a symmetric bilinear decoder, small LP excess risk implies that the learned embeddings admit a transductive linear classifier that recovers the communities. The classifier can also be built explicitly from a small labeled set. The bound takes the schematic form
\begin{equation}
\frac{1}{n}\sum_{i=1}^{n}\indic\{\widehat h(z_i)\neq c_i\}
\;\lesssim\;
\frac{C}{\Deltas^2}\cdot \bigl[\eopt+\eapp\bigr]
\;+\;
\Otilde\!\bigl(\Deltas^{-2}\sqrt{d/n}\bigr),
\label{eq:intro-main}
\end{equation}
where $n$ is the number of nodes, $d$ the embedding dimension, and $\Deltas$ a community-separation margin. $\eopt+\eapp$ is the population LP excess-risk decomposition: $\eopt$ is an in-class optimization residual, and $\eapp$ is an architectural approximation residual. Full notation is given in Sec.~\ref{sec:strategy} and App.~\ref{app:notation}.

\paragraph{Contributions.} Our three main contributions are:

\textbf{(1) A finite-sample LP-risk-to-classification bound (Thm.~\ref{thm:main}, Cor.~\ref{cor:half-empirical}, Cor.~\ref{cor:train-side}).}
For any bounded GNN encoder paired with a symmetric bilinear decoder, LP excess risk controls linear community-classification error on the SBM through \eqref{eq:intro-main}. The population LP quantity can be replaced by the held-out empirical loss (Cor.~\ref{cor:half-empirical}), and Cor.~\ref{cor:train-side} further relates it to the train-side residual, the quantity directly optimized by gradient descent during training. Thus the guarantee is stated in terms of LP losses that can be monitored during pretraining, rather than an unobservable population measure of embedding quality.

\textbf{(2) A constructive few-label classifier with LP-coupled label complexity (Thm.~\ref{thm:main}~Part~II, Prop.~\ref{prop:mahalanobis}).}
Beyond showing that a low-error linear classifier exists when the LP risk is small, we construct one explicitly from a few labeled nodes per community, and show it is exactly the classical nearest-centroid rule in embedding space, under the data-dependent Mahalanobis metric $G\!=\!WZ^\top ZW$ \citep{hastie2009elements} induced by the learned decoder and embeddings. The guarantee therefore covers an explicit linear probe with closed-form weights, applied at $O(d^2)$ cost per node. Its error bound contains an LP-quality floor that more labels cannot remove, and the number of labels the bound requires decreases with the LP risk. The analysis also explains the choice of decoder: for $K\!\ge\!3$, inner-product$+$bias decoders cannot realize some SBM oracle score matrices, while bilinear decoders can under a mild capacity condition.

\textbf{(3) Architecture instantiation and empirical evidence (Prop.~\ref{prop:eapp-rate}, Sec.~\ref{sec:experiments}).}
The reduction is architecture-agnostic through the approximation residual $\eapp$, which can be instantiated for specific encoder families. We show that $\eapp\!=\!\Otilde(\log n/n)$ for a graph isomorphism network (GIN) \citep{xu2019gin} with spectral positional encoding (PE) (Prop.~\ref{prop:eapp-rate}); by the in-class infimum structure, the same rate applies to any encoder--decoder class containing this construction. Synthetic SBM experiments validate the theoretical predictions from LP risk to score error and classification error, and show that the worst-case curvature factor $\muB^{-1}$ can be replaced in practice by a much tighter effective curvature $\mu_{\mathrm{eff}}^{-1}$.

\paragraph{Scope and limitations.}
Our main result is conditional on the LP quality that the trained model achieves. This is a deliberate boundary of the theory. Proving that gradient descent globally optimizes a nonconvex GNN encoder--decoder is a separate problem, open for most deep-learning models, and we do not assume that training automatically reaches a good solution. 
Our theorem instead answers: once a model reaches a certain LP-risk level, what can we guarantee about the downstream linear usability of its embeddings. We do not claim to improve classical SBM recovery rates (App.~\ref{app:rate}), nor that LP pretraining outperforms methods designed specifically for SBM recovery, such as spectral clustering \citep{lei2015consistency} or adjacency spectral embedding \citep{sussman2012consistent}. Our goal is a guarantee for reusable embeddings, not a new recovery algorithm. Finally, $\eapp$ has no uniform rate across encoder families. We instantiate it for spectral-PE GIN. Encoder families that do not contain this construction need separate arguments. In sparse regimes the bound remains valid but uninformative, since $\muB^{-1}$ grows and the dense-regime assumption of the construction fails (App.~\ref{app:rho-impossible}).

\section{Related Work}\label{sec:related-work}

\textbf{Community recovery in the SBM.}
Classical SBM theory characterizes how accurately specific procedures recover communities from the adjacency matrix, including spectral clustering, semidefinite programming, and adjacency spectral embedding \citep{lei2015consistency,hajek2016achieving,lyzinski2014perfect,abbe2018community}, with sharp exact-recovery thresholds \citep{abbe2016exact,mossel2015consistency} and minimax misclassification rates \citep{zhang2016minimax,gao2017achieving}. SBM structure has also guided GNN-based recovery \citep{bruna2017community}. These analyses fix the recovery algorithm and use its node-wise statistics to obtain sharp rates. Our question comes from the pretraining pipeline instead: without specifying the algorithm or imposing conditions on individual embeddings, we ask what the LP risk achieved by a bounded encoder--decoder, an observable black-box quantity, guarantees about its embeddings. Spectral PE enters our analysis only to instantiate $\eapp$. The bilinear decoder itself fits a low-rank matrix of logits, a learned, logit-scale analogue of the generalized random dot product graph, whose latent positions spectral embedding estimates \citep{rubin2022statistical}.

\textbf{Link-prediction objectives and SBM-aware architectures.}
Link and neighborhood prediction are common ways to train graph embeddings, from node2vec and GraphSAGE \citep{grover2016node2vec,hamilton2017inductive} to variational graph autoencoders (VGAE) with an inner-product decoder \citep{kipf2016variational}. Modern graph pretraining and prompting methods also use LP-style or node--pair matching objectives, including GPT-GNN, GPPT, and GraphPrompt \citep{hu2020gptgnn,sun2022gppt,liu2023graphprompt}, and some architectures make the LP model explicitly block-aware, such as the variational-autoencoder GNN of \citet{mehta2019stochastic}. These methods motivate our setting but do not by themselves give a finite-sample statement of the form ``LP risk controls linear node-classification error''; our result supplies one under the SBM coupling.

\textbf{Self-supervised pretraining to downstream guarantees.}
Recent work has started to turn a self-supervised loss into a quantitative downstream-classification bound. \citet{arora2019theoretical} assume positive pairs share a latent class, and \citet{haochen2021provable} replace this with augmentation-graph connectivity; \citet{tosh2021contrastive} use multi-view redundancy, \citet{lee2021predicting} conditional independence between input parts, and \citet{wei2021why} analyze masked prediction in language models. Graph self-supervised learning has developed contrastive losses on augmented views \citep{velickovic2019deep,you2020graphcl,zhu2020deep}, but quantitative downstream guarantees remain rare. The closest result is by \citet{davison2024community}: for minimizers of the node2vec loss over free per-node embeddings, k-means consistently recovers communities in (degree-corrected) SBMs as $n$ grows. We instead study masked LP with GNN encoder--decoders, the form widely used in practical graph pretraining, and bound the finite-sample error of an explicit few-label linear classifier through the LP risk the model attains, not through optimality of the training objective.

\section{Theoretical Framework}\label{sec:theory}
 
\subsection{Setting and Definitions}\label{sec:setting}

Let $\mathcal{G}=(V,E)$ be an undirected, unweighted graph with node set $V=[n]:=\{1,\dots,n\}$ and edge set $E\subseteq\binom{V}{2}$; equivalently, $\mathcal{G}$ is encoded by a symmetric adjacency matrix $A\in\{0,1\}^{n\times n}$ with $A_{ij}=\indic\{\{i,j\}\in E\}$ and $A_{ii}=0$.

\textbf{Stochastic block model.} The SBM \citep{holland1983stochastic, snijders1997estimation} partitions $V$ into $K$ communities: each node $i\in V$ has an unobserved community label $c_i\in[K]$ with proportions $\pi_a=|\{i:c_i=a\}|/n\ge\pi_{\min}>0$, and edges are drawn independently as $A_{ij}\sim\Ber(P_{ij})$ for $i<j$, where $P_{ij}=p_{c_ic_j}$ is the $(c_i,c_j)$ entry of a symmetric block probability matrix $\Theta=(p_{ab})_{a,b\in[K]}\in(0,1)^{K\times K}$. We write $N:=\binom{n}{2}$ for the total number of node pairs.

\textbf{Masking.} Fix a held-out fraction $\tau\in(0,1)$, draw a set $\Hcal$ of $\lceil\tau N\rceil$ node pairs uniformly at random from $\{(i,j):i<j\}$, and let $\Tcal=\{(i,j):i<j\}\setminus\Hcal$ be the remaining pairs. The encoder input is the masked adjacency $\Aenc$, obtained from $A$ by zeroing the held-out entries: $\Aenc_{ij}=\Aenc_{ji}=A_{ij}\indic\{(i,j)\in\Tcal\}$ for $i<j$ and $\Aenc_{ii}=0$. Pairs in $\Tcal$ (edges and non-edges) are the visible/train pairs used in the LP training objective, while pairs in $\Hcal$ are hidden from the encoder and used for held-out evaluation. All $n$ nodes remain in the graph: only node pairs are split. Since $\Hcal$ and $\Tcal$ are disjoint and the entries $A_{ij}$ are independent, $\Aenc$ is independent of $\{A_{ij}:(i,j)\in\Hcal\}$ conditional on the split $(\Hcal,\Tcal)$ (App.~\ref{app:setting}).

\textbf{Encoder and general bilinear decoder.} A GNN encoder $f_\theta$ produces embeddings $z_i:=f_\theta(\Aenc,i)\in\Real^d$ for node $i$, where the embedding dimension $d$ is treated as fixed throughout, and the \emph{general bilinear decoder} outputs scores (estimates of $\logit P_{ij}$)
\begin{equation}\label{eq:score-def}
\score{ij}=\bhat+z_i^\top W z_j,\quad i\ne j,\qquad W=W^\top\in\Real^{d\times d},\ \bhat\in\Real,
\end{equation}
with the notational extension $\score{ii}:=\bhat+z_i^\top W z_i$, used only in Step~3 of the proof strategy (Sec.~\ref{sec:strategy}) and not in the LP objective. The inner-product decoder is the special case $W=I_d$, the $d\times d$ identity matrix. We retain the general bilinear $W$ because for $K\ge 3$ there exist SBM regimes in which the community-level oracle score matrix $M^\star\in\Real^{K\times K}$ with entries $m^\star_{ab}:=\logit(p_{ab})$ is not realizable by any inner-product$+$bias decoder, while the bilinear form recovers it exactly under a mild capacity condition (made precise after Assumption~\ref{ass:bounded}; App.~\ref{app:setting}). The map $f_\theta$ is left abstract: any GNN family is admissible, with $\|z_i\|\le R$ enforced via a final projection if needed.

\textbf{LP risks.} LP is commonly trained with the binary cross-entropy (BCE) loss $\ell(s;p)=-p\log\sigma(s)-(1-p)\log(1-\sigma(s))$ \citep{kipf2016variational}, with $\sigma(s)=(1+e^{-s})^{-1}$. We use this loss to define, for any pair set $\mathcal{E}\subseteq\{(i,j):i<j\}$, the empirical and population LP risks
\begin{equation*}
\widehat R_\mathcal{E}(S):=\tfrac{1}{|\mathcal{E}|}\!\sum_{(i,j)\in\mathcal{E}}\!\ell(\score{ij};A_{ij}),\qquad R^\Pop_\mathcal{E}(S):=\tfrac{1}{|\mathcal{E}|}\!\sum_{(i,j)\in\mathcal{E}}\!\ell(\score{ij};P_{ij})=\E_A\bigl[\widehat R_\mathcal{E}(S)\bigr].
\end{equation*}
The empirical $\widehat R_\bullet$ uses the observed Bernoulli targets $A_{ij}$, and the population $R^\Pop_\bullet$ replaces them by the underlying probabilities $P_{ij}$. Specializing to $\mathcal{E}=\Tcal$ gives the training objective $\widehat R_\Tcal$ that gradient descent optimizes directly, and $\mathcal{E}=\Hcal$ gives the held-out risks $\widehat R_\Hcal,R^\Pop_\Hcal$ used in the main theorem.

\begin{assumption}[Bounded encoder--decoder class]\label{ass:bounded}
We restrict the encoder--decoder family to parameters $(\theta,W,\bhat)$ with $\|z_i\|\le R$ for all $i$, $W=W^\top$, $\|W\|_{\mathrm{op}}\le\Wmax$, and $|\bhat|\le B_0$, for some constants $R,\Wmax,B_0<\infty$, and assume that the trained score matrix $\Shat$ comes from this family.
\end{assumption}
Cauchy--Schwarz bounds every learned score by $|\score{ij}|\le\Bmodel:=B_0+\Wmax R^2$, and the oracle scores $m^\star_{ab}$ are bounded by $B^\star:=\max_{a,b}|m^\star_{ab}|$. With $B:=\max(\Bmodel,B^\star,\log 2)$, all learned and oracle scores lie in $[-B,B]$, and the $\log 2$ floor ensures $\ell(s;p)\in[0,2B]$ (App.~\ref{app:train-side}). Boundedness caps the complexity of the encoder--decoder class and gives finite-sample concentration. We call $\Wmax R^2$ the \emph{bilinear capacity}: if $\|M^\star-b\,\mathbf 1\mathbf 1^\top\|_{\op}\le\Wmax R^2$ for some bias $|b|\le B_0$, then $M^\star$ is exactly realizable by the bilinear form within these bounds (App.~\ref{app:setting}).

\subsection{Key Quantities and Proof Strategy}\label{sec:strategy}
 
\textbf{Oracle, hypothesis class, and excess risk.} The oracle scores $m^\star_{ab}=\logit(p_{ab})$ induce the \emph{oracle score matrix} $\Sopt:=(m^\star_{c_ic_j})_{i,j\in[n]}\in\Real^{n\times n}$. Per-pair BCE is strictly convex with unique minimizer $\logit(P_{ij})=m^\star_{c_ic_j}$, so $\Sopt$ minimizes $\Rpop$, uniquely on $\Hcal$. The encoder--decoder hypothesis class is
\begin{equation*}
\Scal(A,\Tcal) := \bigl\{(s_{ij})_{i,j\in[n]} : s_{ij}=\bhat+z_i^\top W z_j,\ z_i=f_\theta(\Aenc,i),\ (\theta,W,\bhat)\text{ satisfies Assumption~\ref{ass:bounded}}\bigr\}.
\end{equation*}
Let $\Rdag:=\inf_{S\in\Scal(A,\Tcal)}\Rpop(S)$ denote the in-class population infimum. The population held-out excess risk decomposes \emph{exactly} as residuals
\begin{equation}\label{eq:excess-decomp}
\Rpop(\Shat)-\Rpop(\Sopt) \;=\; \eopt+\eapp,\qquad \eopt:=\Rpop(\Shat)-\Rdag,\quad \eapp:=\Rdag-\Rpop(\Sopt),
\end{equation}
both non-negative. The \textbf{held-out statistical deviation} is
\(
\GammaH \;:=\; \sup_{S\in\Scal(A,\Tcal)}\bigl|\Remp(S)-\Rpop(S)\bigr|.
\)

\textbf{Interpretation of the separation: in-class vs.\ out-of-class.} $\eopt$ measures how well the trained model approaches the in-class population optimum and is controllable through training (Cor.~\ref{cor:train-side}). $\eapp$ measures how closely the encoder--decoder class can in principle approach the unrestricted Bayes optimum: it is an architectural ceiling set by the class's expressiveness, irreducible by any training procedure (Sec.~\ref{sec:eapp}). All training-side statistics ($\widehat R_\bullet$ and uniform deviations $\GammaH,\GammaT,\GammaTH$) operate inside the class and therefore control only $\eopt$.

\textbf{Geometric quantities.} The held-out and full score mean squared errors (MSE) are
\[
\mathrm{ScoreMSE}_\Hcal := \frac{1}{|\Hcal|}\sum_{(i,j)\in\Hcal}\!\!(s_{ij}-m^\star_{c_i c_j})^2,\qquad
\mathrm{ScoreMSE}_{\mathrm{full}} := \frac{1}{n^2}\sum_{i,j\in[n]}(s_{ij}-m^\star_{c_i c_j})^2,
\]
the latter using the diagonal extension $s_{ii}$.
We further define two constants used in the bound: the strong-convexity parameter of BCE on $[-B,B]$, $\muB:=\sigma(B)(1-\sigma(B))=\min_{|s|\le B}\sigma'(s)>0$, entering Step~1; and the score-profile separation $\Deltas^2:=\min_{a\ne b}\sum_{c=1}^K\pi_c(m^\star_{ac}-m^\star_{bc})^2$, entering Step~3 ($\Deltas^2>0$ if and only if the rows of $M^\star$ are distinct, the SBM identifiability condition).

\paragraph{Four-step proof strategy.} Our proof of the classification-error bound is divided into four steps, $\Remp\to\Rpop\to\mathrm{ScoreMSE}_\Hcal\to\mathrm{ScoreMSE}_{\mathrm{full}}\to$ error. Detailed proofs are in Appendices~\ref{app:step0}--\ref{app:step3}.

\begin{lemma}[Step 0: empirical $\to$ population on $\Hcal$]\label{lem:step0}
Under Assumption~\ref{ass:bounded}, there exists a universal constant $C_0>0$ such that with probability at least $1-n^{-1}$,
\[
\GammaH \;:=\; \sup_{S\in\Scal(A,\Tcal)}\bigl|\Remp(S)-\Rpop(S)\bigr| \;\le\; C_0\,B\sqrt{\frac{d\log(nBR\Wmax)}{\tau n}}.
\]
\end{lemma}
\noindent\emph{Empirical-to-population bridge.} Since the encoder--decoder class is bounded, held-out empirical risks concentrate around their population values uniformly over the class, so $\GammaH=\Otilde(\sqrt{d/(\tau n)})\to0$ (App.~\ref{app:step0}). Without boundedness, the class can contain score matrices that fit the realized held-out entries $A_{ij}$, and for these the two risks can be arbitrarily far apart.

\begin{lemma}[Step 1: population strong convexity]\label{lem:step1}
Under Assumption~\ref{ass:bounded},
\[
\mathrm{ScoreMSE}_\Hcal \;\le\; \frac{2}{\muB}\bigl(\Rpop(\Shat)-\Rpop(\Sopt)\bigr) \;=\; \frac{2}{\muB}(\eopt+\eapp).
\]
\end{lemma}
\noindent\emph{Excess risk drives scores toward the oracle.} Per-pair BCE on $[-B,B]$ is $\muB$-strongly convex with Bayes minimum at $m^\star_{c_i c_j}=\logit(P_{ij})$, so excess risk upper-bounds score MSE quadratically.

\begin{lemma}[Step 2: held-out $\to$ full ScoreMSE]\label{lem:step2}
Under Assumption~\ref{ass:bounded}, there exists a universal constant $C_1>0$ such that with probability at least $1-n^{-1}$, simultaneously for all $S\in\Scal(A,\Tcal)$,
\[
\mathrm{ScoreMSE}_{\mathrm{full}} \;\le\; \mathrm{ScoreMSE}_\Hcal + \Vstat,\qquad \Vstat := \frac{C_1 B^2}{\sqrt\tau}\sqrt{\frac{d\log(nBR\Wmax)}{n}}+\frac{4B^2}{n}.
\]
\end{lemma}
\noindent\emph{Held-out averages transfer to all pairs.} $\Hcal$ is a uniform random $\tau$-fraction of node pairs; uniform sampling-without-replacement concentration over the bounded class transfers averages from $\Hcal$ to all $i<j$, plus a $4B^2/n$ correction for the diagonal entries.

\begin{proposition}[Step 3: ScoreMSE $\to$ classification error]\label{prop:step3}
Fix a trained $(Z,W,\bhat)$ satisfying Assumption~\ref{ass:bounded}, where $Z\in\Real^{n\times d}$ stacks the embeddings $z_i$ as rows. For each node $i$, let $\tilde s_i := (s_{ij})_{j\in[n]}\in\Real^n$ be the full score profile, and $\mu_a := (m^\star_{a,c_j})_{j\in[n]}\in\Real^n$ the \emph{oracle prototype} (the ideal profile under community $a$). We define the \emph{oracle classifier} $h^\star$ to assign each node to its nearest oracle prototype in score-profile space; using $W=W^\top$, the bilinear identity $\tilde s_i = \bhat\mathbf{1}_n + ZWz_i$ allows this rule to be rewritten as an equivalent linear classifier in the embedding $z_i$:
\[
h^\star(z_i) \;:=\; \arg\min_{a\in[K]}\|\tilde s_i - \mu_a\| \;=\; \arg\max_{a\in[K]}\bigl(w_a^{\star\top} z_i + b_a^\star\bigr),
\]
where $w_a^\star := WZ^\top\mu_a$ and $b_a^\star := \bhat\mu_a^\top\mathbf{1}_n - \tfrac12\|\mu_a\|^2$. Identifiability $\Deltas>0$ separates prototypes by $\sqrt n\,\Deltas$, and Markov's inequality on $\tilde e_i^2 := n^{-1}\|\tilde s_i-\mu_{c_i}\|^2$ bounds the error of $h^\star$:
\[
\frac{1}{n}\sum_{i=1}^n\indic\{h^\star(z_i)\ne c_i\} \;\le\; \frac{4\,\mathrm{ScoreMSE}_{\mathrm{full}}}{\Deltas^2}.
\]
\end{proposition}

\subsection{Main Results}\label{sec:main-results}
 
Combining Steps~0--3 yields a finite-sample classification guarantee in two forms. \textbf{Part~I (Oracle)} follows by combining the Step~0--2 envelope $\mathrm{ScoreMSE}_{\mathrm{full}} \le \En$ with the geometric reduction of Prop.~\ref{prop:step3}. \textbf{Part~II (Plug-in)} estimates the unobservable oracle prototypes $\mu_a$ from a small labeled set $L=\bigcup_{a=1}^K L_a$, giving a constructive linear classifier that uses only the trained model and $L$.

\begin{theorem}[LP-trained embeddings enable linear community detection]\label{thm:main}
Let $(Z,W,\bhat)$ be any trained model satisfying Assumption~\ref{ass:bounded}; it may depend on the data and on optimizer randomness. Assume $\Deltas>0$ and that $L$ is drawn independently of the trained model. Then, for every $n\ge2$, with probability $\ge 1-3n^{-1}$ over $(A,\Hcal,L)$:
 
\smallskip\noindent\textbf{Part~I (Oracle).} The oracle classifier $h^\star$ of Prop.~\ref{prop:step3} satisfies
\[
\begin{gathered}
\frac{1}{n}\sum_{i=1}^n\indic\{h^\star(z_i)\ne c_i\} \;\le\; \frac{4\,\En}{\Deltas^2},\qquad \En := \frac{2}{\muB}(\eopt+\eapp) + \Vstat,\\
\Vstat = \frac{C_1B^2}{\sqrt\tau}\sqrt{\frac{d\log(nBR\Wmax)}{n}}+\frac{4B^2}{n}.
\end{gathered}
\]
\smallskip\noindent\textbf{Part~II (Plug-in).} For each $a\in[K]$ sample a set $L_a$ of nodes in community $a$ uniformly without replacement, with $|L_a|=m_a$ and $m_{\min}:=\min_a m_a$; form the empirical prototype estimate $\hat\mu_a := |L_a|^{-1}\sum_{i\in L_a}\tilde s_i$. Then
\[
\eta_L := \max_a\tfrac{1}{\sqrt n}\|\hat\mu_a-\mu_a\| \;\le\; \sqrt{\frac{\En}{\pi_{\min}}}+\sqrt{\frac{2\En\lambda_n}{m_{\min}\pi_{\min}}}+\frac{4B\lambda_n}{3m_{\min}},\qquad \lambda_n := \log(2Kn);
\]  
on the event $\eta_L<\Deltas/2$, the linear \emph{plug-in classifier} $\hat h^{\mathrm{plug}}(z) := \arg\max_{a}(\hat w_a^\top z + \hat b_a)$ with $\hat w_a := WZ^\top\hat\mu_a$, $\hat b_a := \bhat\,\hat\mu_a^\top\mathbf 1_n - \tfrac12\|\hat\mu_a\|^2$, i.e., $h^\star$ with $\mu_a$ replaced by $\hat\mu_a$, satisfies
\[
\frac{1}{n}\sum_{i=1}^n\indic\{\hat h^{\mathrm{plug}}(z_i)\ne c_i\} \;\le\; \frac{4\,\En}{(\Deltas-2\eta_L)^2}.
\]
\end{theorem}
\noindent In the bound on $\eta_L$, the term $\sqrt{\En/\pi_{\min}}$ is an LP-quality floor: it does not vanish with more labels. Define the label complexity as the total number of labels $|L|=\sum_a m_a$ required to guarantee $\eta_L<\Deltas/2$. It is coupled with LP quality: smaller $\En$ lowers the label requirement, and when $\eta_L$ is small, the plug-in bound approaches the oracle bound $4\En/\Deltas^2$. Unlike VC-type linear-probing bounds, which scale with the embedding dimension, our label requirement depends on $d$ only through $\En$, that is, through the quality of the LP-trained embeddings.

\begin{proposition}[Mahalanobis nearest-centroid form of the plug-in classifier]\label{prop:mahalanobis}
Let $\bar z_a := |L_a|^{-1}\sum_{i\in L_a}z_i\in\Real^d$ be the empirical embedding centroid of class $a$, and $G := WZ^\top ZW \in \Real^{d\times d}$. Then $G$ is positive semi-definite (PSD; $G\succeq 0$), and
\[
\hat h^{\mathrm{plug}}(z) \;=\; \arg\min_{a\in[K]}(z-\bar z_a)^\top G(z-\bar z_a) \;=\; \arg\max_{a\in[K]}\bigl\{\bar z_a^\top G z - \tfrac12\bar z_a^\top G\bar z_a\bigr\}.
\]
\end{proposition}
 
\smallskip\noindent The plug-in classifier is therefore the classical nearest-centroid rule~\citep{mensink2013distance,snell2017prototypical} in the embedding space under the data-dependent PSD form $G$, with $\bar z_a$ the centroid of the labeled \emph{embeddings} (not score profiles). The inner-product case $W=I$ gives $G=Z^\top Z$, the embedding Gram matrix, and the general bilinear case is the same construction in a $W$-deformed metric. Once $G$ is formed from all embeddings, classifying a node needs only its own embedding and takes $O(d^2)$ time, with no need to construct the $n$-dim score profile $\tilde s_i$ explicitly.

Without labels, the same form gives an unsupervised guarantee: if $(16+8\kappa)\En/\Deltas^2<\pi_{\min}$, any $(1+\kappa)$-approximate $K$-means clustering of the embeddings under $G$ misclassifies at most a $(16+8\kappa)\En/\Deltas^2$ fraction of nodes, up to a label permutation (App.~\ref{app:kmeans}, based on \citealp[Lemma~5.3]{lei2015consistency}). Hence $\En$ and $\Deltas$ control the error bounds of the oracle, plug-in, and unsupervised classifiers alike.

\begin{corollary}[Half-empirical bound]\label{cor:half-empirical}
Replacing the held-out population risk $\Rpop(\Shat)$ in Theorem~\ref{thm:main} Part~I by the empirical loss $\widehat R_\Hcal(\Shat)$ adds the Step~0 edge-noise concentration term $\GammaH$:
\[
\frac{1}{n}\sum_{i=1}^n\indic\{h^\star(z_i)\ne c_i\} \;\le\; \frac{8}{\muB\Deltas^2}\bigl[\widehat R_\Hcal(\Shat) - \Rpop(\Sopt) + \GammaH\bigr] + \frac{4\Vstat}{\Deltas^2}.
\]
\end{corollary}
\noindent Here $\widehat R_\Hcal(\Shat)$ is read from held-out evaluation. The Bayes floor $\Rpop(\Sopt)$ is unknown but does not depend on the trained model, and both $\GammaH$ and $\Vstat$ vanish at the parametric rate $\Otilde(\sqrt{d/n})$ (App.~\ref{app:step0} and~\ref{app:step2}). The same replacement applies to Part~II (plug-in).

\paragraph{Train-side bridge.} 
$\eopt$ measures LP-training quality (Sec.~\ref{sec:strategy}). We bound it via the empirical train-set risk that gradient descent directly minimizes. Define the in-class empirical floor and residual
\[
\Rempdag \;:=\; \inf_{S\in\Scal(A,\Tcal)}\Rempt(S),\qquad
\widehat\rho \;:=\; \Rempt(\Shat)-\Rempdag\;\ge\;0.
\]
\begin{corollary}[Train-side bound]\label{cor:train-side}
Define the train-side and cross-side uniform deviations
\[
\GammaT := \!\!\sup_{S\in\Scal(A,\Tcal)}\!\!\bigl|\Rempt(S)-\Rpopt(S)\bigr|,\qquad
\GammaTH := \!\!\sup_{S\in\Scal(A,\Tcal)}\!\!\bigl|\Rpopt(S)-\Rpop(S)\bigr|,
\]
each bounded by $\Otilde\bigl(B^2\sqrt{d\log(nBR\Wmax)/n}\bigr)$ with probability $\ge 1-2n^{-1}$ (App.~\ref{app:train-side}). Then $\eopt \le \widehat\rho + 2(\GammaT+\GammaTH)$, and Theorem~\ref{thm:main} Part~I becomes
\[
\frac{1}{n}\sum_i\indic\{h^\star(z_i)\ne c_i\} \;\le\; \frac{8}{\muB\Deltas^2}\bigl[\widehat\rho+\eapp+2(\GammaT+\GammaTH)\bigr]+\frac{4\Vstat}{\Deltas^2}.
\]
\end{corollary}
\noindent\emph{Gradient descent targets $\widehat\rho$, while $\GammaT$ and $\GammaTH$ quantify overfitting.} Here $\widehat\rho\to0$ means that training reaches the in-class optimum, a nonconvex optimization question outside our scope. $\GammaT$ measures overfitting to the realized training edges, and $\GammaTH$ the mismatch between train and held-out pairs. Boundedness (Assumption~\ref{ass:bounded}) caps the complexity of the class, so both vanish at the parametric rate. At small $n$ they can dominate, and even $\widehat\rho=0$ then certifies only $\eopt\le2(\GammaT+\GammaTH)$.

\subsection{Architectural Realization of $\eapp$ and Scope}\label{sec:eapp}

The approximation residual $\eapp$ depends on the chosen architecture and admits no uniform rate, so we treat it as a \emph{half black box}: we verify the rate on one representative GNN family, and the in-class infimum structure transfers it to any class containing this family.

\begin{proposition}[$\eapp$ rate for spectral-PE GIN; informal]\label{prop:eapp-rate}
Consider a GIN encoder \citep{xu2019gin} with $L\ge1$ layers and hidden width at least $2K$, taking the top-$D_{\mathrm{PE}}$ absolute-eigenvalue eigenvectors of $\Aenc$ (rescaled by $\sqrt n$) as positional encoding for a fixed $D_{\mathrm{PE}}\ge K$, paired with the general bilinear decoder of Sec.~\ref{sec:setting} with embedding dimension $d\ge K$. Under \textbf{(A1)} dense regime $p_{ab}\in[p_{\min},1-p_{\min}]$ for some $p_{\min}\in(0,1/2)$, \textbf{(A2)} balanced communities $\pi_a\ge\pi_{\min}>0$, \textbf{(A3)} $\rank(\Theta)=K$ with $\min_{k\le K}|\lambda_k^\star|\ge c_\gamma n$ for the nonzero eigenvalues $\lambda_k^\star$ of $P$, \textbf{(A4)} nontrivial masking $\tau \in (0, 1-\tau_0]$, and the bilinear capacity condition $\Wmax R^2\ge\min_{|b|\le B_0}\|M^\star-b\mathbf{1}\mathbf{1}^\top\|_{\mathrm{op}}$, there exist constants $C,c>0$ depending only on $(K,p_{\min},\pi_{\min},c_\gamma,\tau_0,\Wmax,R)$ such that, with probability $\ge 1-n^{-c}$, the spectral-PE GIN class satisfies
\[
\eapp \;\le\; \frac{C\log n}{n} \;=\; \Otilde(1/n).
\]
\end{proposition}

\noindent Proposition~\ref{prop:eapp-rate} is an existence statement: it exhibits one admissible parameter setting, which may depend on $\Aenc$. The complete proof is in App.~\ref{app:eapp}.

\textbf{Scope.}
Thm.~\ref{thm:main} together with Cor.~\ref{cor:train-side} and Prop.~\ref{prop:eapp-rate} closes the loop. If gradient descent on $\widehat R_\Tcal$ successfully drives the empirical residual to $\widehat\rho=\Otilde(\sqrt{d/n})$ and the architecture realizes $\eapp=\Otilde(1/n)$, then $\eopt=\Otilde(\sqrt{d/n})$ by Cor.~\ref{cor:train-side}, and $\En$ decays at the parametric rate $\Otilde(\sqrt{d\log(nBR\Wmax)/n})$, dominated by the statistical terms. By Thm.~\ref{thm:main},
\[
\frac{1}{n}\sum_{i=1}^n\indic\{h^\star(z_i)\ne c_i\}\;=\; \Otilde\!\left(\tfrac{1}{\Deltas^2}\sqrt{d\log n / n}\right)\;\xrightarrow{n\to\infty}\; 0.
\]
However, the conditional clause is non-trivial: this rate antecedent requires both a sufficiently expressive architecture and successful optimization, neither of which is automatic. The result is therefore a theoretical guarantee on the \emph{achievable} error once these antecedents are met, not a claim about training dynamics.

\section{Experiments}\label{sec:experiments}

We empirically validate the finite-sample predictions of Sec.~\ref{sec:theory} on synthetic SBMs. We use symmetric assortative SBMs with $K\!\in\!\{2,3,4\}$, balanced communities, held-out fraction $\tau\!=\!0.2$, embedding dimension $d\!=\!16$, norm bound $R\!=\!1.5$, and the general bilinear decoder. The baseline is $K\!=\!3$, $p_{\mathrm{in}}\!=\!0.10$, $p_{\mathrm{out}}\!=\!0.04$ ($\Deltas\!=\!0.80$), and density multiplier $\rho_n\!=\!1$, which scales all block probabilities. We denote configurations by \texttt{architecture+features}. \textbf{GIN+PE} uses a GIN encoder \citep{xu2019gin} with spectral PE: the $D_{\mathrm{PE}}\!=\!8$ eigenvectors of $\Aenc$ with the largest absolute eigenvalues, rescaled by $\sqrt n$, a setting covered by Prop.~\ref{prop:eapp-rate}.  \textbf{GCN+PE} uses a graph convolutional network (GCN) encoder \citep{kipf2017gcn} with Laplacian PE: the $8$ eigenvectors of the normalized Laplacian of $\Aenc$ with the smallest nontrivial eigenvalues, each column standardized. \textbf{GIN+random} replaces the PE with i.i.d.\ Gaussian features, and \textbf{CP+random} uses a centered-power (CP) encoder, a power-iteration encoder inspired by ChebNet \citep{defferrard2016chebnet}. All main-text results are at $K\!=\!3$. Results for $K\!\in\!\{2,4\}$ and for GAT+PE (a graph attention network (GAT) \citep{velickovic2018gat} with Laplacian PE) are in App.~\ref{app:setup-and-K}, and full hyperparameters are in App.~\ref{app:hyperparams}.

\textbf{Norm rules and training.}
A sufficient condition for the bilinear form to realize $M^\star$ (Sec.~\ref{sec:setting}) is $\Wmax R^2\ge\mathrm{cap}:=\min_{|b|\le B_0}\|M^\star-b\,\mathbf 1\mathbf 1^\top\|_{\op}$. We set $B_0=|b^\star|+0.5$ for an unconstrained minimizer $b^\star$, so $\mathrm{cap}=|\logit p_{\mathrm{in}}-\logit p_{\mathrm{out}}|$ for our SBMs, and set $\Wmax R^2=(\text{cap-ratio})\cdot\mathrm{cap}$ with default cap-ratio $2$. The bound $\|W\|_{\op}\le\Wmax$ is enforced by clipping the eigenvalues of $W$. The $\Deltas$ sweep instead fixes $\Wmax\!=\!1.5$ to keep the bilinear capacity fixed, while the $\rho_n$ sweep follows the same rule as the $n$-sweep (App.~\ref{app:hyperparams}). We train with full-batch BCE using Adam (lr $10^{-3}$) for 2500 epochs, initialize the bias at zero, and report medians over 10 seeds with IQR shading. Code is available at \url{https://github.com/YChen1103/LP-to-Linear-Community-Detection}.

\begin{figure}[!htbp]
\centering
\includegraphics[width=\linewidth]{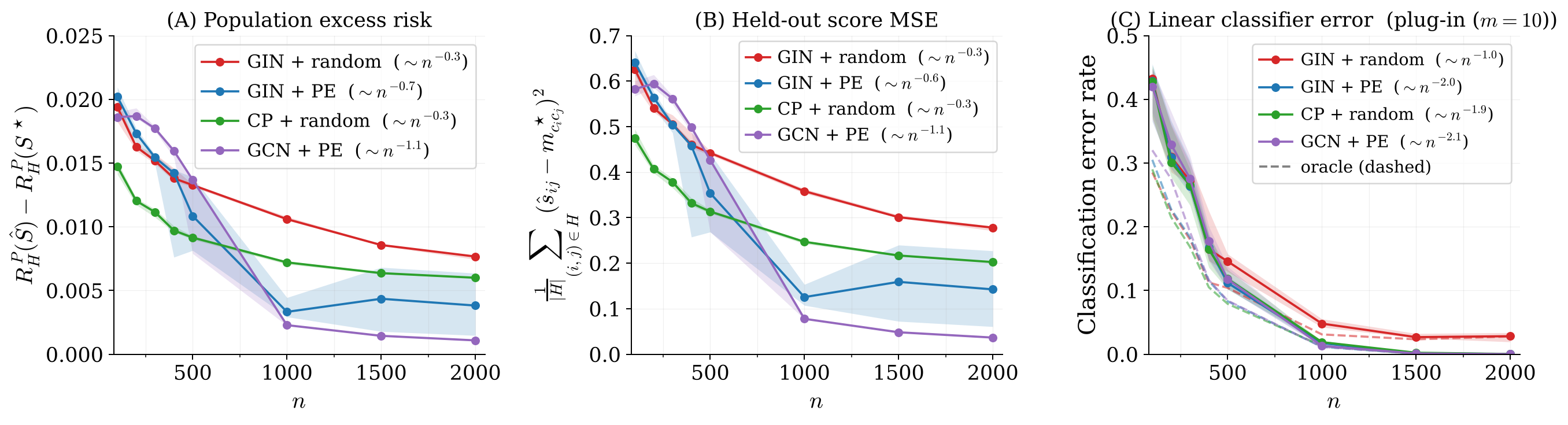}
\vspace{-1.4em}
\caption{\textbf{Theorem chain at $K\!=\!3$.}
\textbf{(A)} Population excess risk (held-out) $\Rpop(\Shat)\!-\!\Rpop(\Sopt)$.
\textbf{(B)} Held-out score MSE.
\textbf{(C)} Classification error of the plug-in classifier of Thm.~\ref{thm:main} Part~II ($m\!=\!10$, solid) and of the oracle classifier of each configuration (dashed, same color). Curves are medians over seeds with IQR shading. Slopes are log--log fits over $n\!\in\![100,2000]$, excluding errors below $10^{-3}$ in (C).}
\label{fig:chain}
\vspace{-1.2em}
\end{figure}

\paragraph{Theorem chain at $K\!=\!3$ (Fig.~\ref{fig:chain}).}
At this configuration, the three chain quantities (population excess risk, held-out score MSE, and classification error) decrease together as $n$ grows, validating the vanishing trend predicted by Thm.~\ref{thm:main}. Excess and score-MSE curves are nearly parallel for every architecture: the Step~1 ratio (excess $\to$ ScoreMSE) $\mathrm{ScoreMSE}/[\Rpop(\Shat)\!-\!\Rpop(\Sopt)]$ stays between $32$ and $39$, far below the worst-case $2/\muB\!\approx\!425$, a gap we revisit at the end of this section.

The decay rate is, however, architecture-dependent. Both PE-based configurations drive the excess risk toward zero, at empirical slopes of about $-0.7$ (GIN+PE) and $-1.1$ (GCN+PE). For GIN+PE this is broadly compatible with the $\Otilde(\log n/n)$ rate of Prop.~\ref{prop:eapp-rate}. Random-feature configurations decay only at slope $-0.3$ and, for $n\!\ge\!1000$, sit above the PE-based ones, consistent with an $\eapp$ floor absent an architectural argument. GIN+random also plateaus in error for large $n$, while the other three configurations reach near-zero error.

Classification error decays \emph{faster} than score MSE. Since $\Deltas$ is fixed, this gap shows that the Markov inequality in Step~3 (ScoreMSE $\to$ error) is loose deep inside the recovery regime (App.~\ref{app:delta-tight}). Direct logistic regression on the same labels closely matches our plug-in classifier (Tab.~\ref{tab:three-classifier} in App.~\ref{app:three-classifier}). Applied instead to the raw $8$-dimensional PE that GIN+PE receives, the same probe has higher median error at every $n\!\in\!\{500,1000,2000\}$ and $m\!\in\!\{1,5,10,20\}$, e.g., $0.392$ versus $0.109$ at $n\!=\!1000$ with one label per class, so LP training itself adds to its spectral input (App.~\ref{app:pe-baseline}).

\begin{figure}[!htbp]
\centering
\begin{minipage}[t]{0.42\linewidth}
\centering\includegraphics[width=\linewidth]{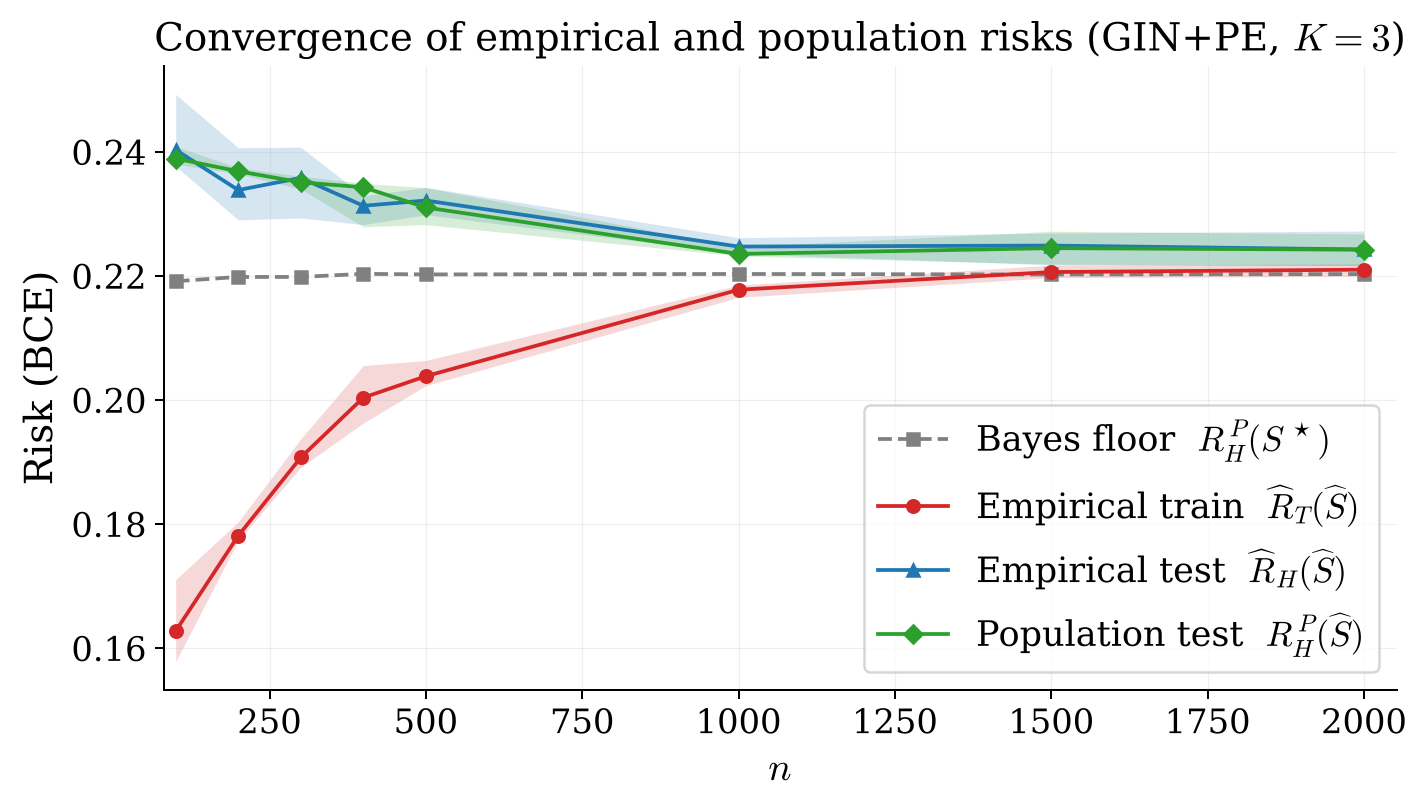}
\vspace{-1.4em}
\caption{Risk closure on the Bayes floor (GIN+PE, $K\!=\!3$): empirical train, empirical held-out, and population held-out risks all close on $\Rpop(\Sopt)$ as $n$ grows.}
\label{fig:bayes}
\end{minipage}\hfill
\begin{minipage}[t]{0.56\linewidth}
\centering\includegraphics[width=\linewidth]{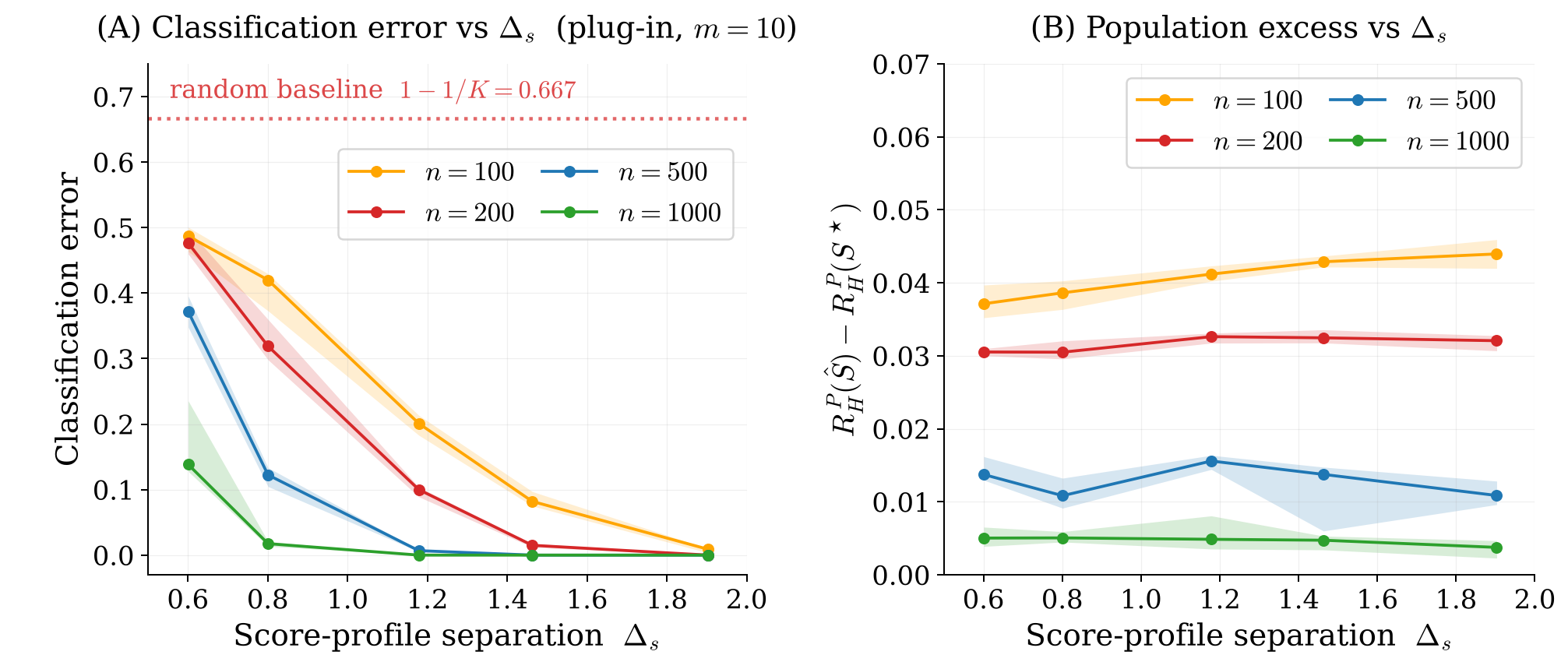}
\vspace{-1.4em}
\caption{$\Deltas$ sweep (GIN+PE, $K\!=\!3$). \textbf{(A)} Error decreases monotonically with $\Deltas$ at every $n$. \textbf{(B)} Population excess varies little across $\Deltas$ (CV $\le 15\%$).}
\label{fig:delta}
\end{minipage}
\vspace{-1em}
\end{figure}

\paragraph{Closure to the Bayes floor (Fig.~\ref{fig:bayes}).}
As $n$ increases, the empirical and population held-out risks closely track each other, and both are within $0.004$ of the Bayes floor $\Rpop(\Sopt)$ at $n=2000$. This supports the predicted decay of the statistical-noise term $\GammaH$ and justifies using the empirical held-out loss as a numerical monitor. The empirical train risk can be far below the floor for small $n$, because the encoder may overfit the realized Bernoulli edges. This is a finite-sample noise artifact: as $n$ grows, it diminishes because the bounded encoder--decoder class restricts overfitting, consistent with Cor.~\ref{cor:train-side}.

\paragraph{Geometric factor $\Deltas$ (Fig.~\ref{fig:delta}).}
We vary $\Deltas$ from $0.60$ to $1.90$ through $p_{\mathrm{in}}$ ($p_{\mathrm{out}}$ fixed) at fixed architecture and capacity. Classification error decreases monotonically as $\Deltas$ grows (Fig.~\ref{fig:delta}A), while the population excess risk changes only mildly (coefficient of variation (CV) $\le\!15\%$, Fig.~\ref{fig:delta}B). Thus, larger $\Deltas$ seems to help mainly through the geometric factor in
\(
\mathrm{Error}\le 4\En/\Deltas^2
\)
(Thm.~\ref{thm:main}), rather than by substantially reducing the LP excess risk. This matches the bound, since Steps~0--2 have no explicit $\Deltas$ dependence. The $\Deltas^{-2}$ factor is empirically tight near the recovery boundary, but looser deep inside the recovery regime (App.~\ref{app:delta-tight}).

\begin{figure}[!htbp]
\centering
\includegraphics[width=0.92\linewidth]{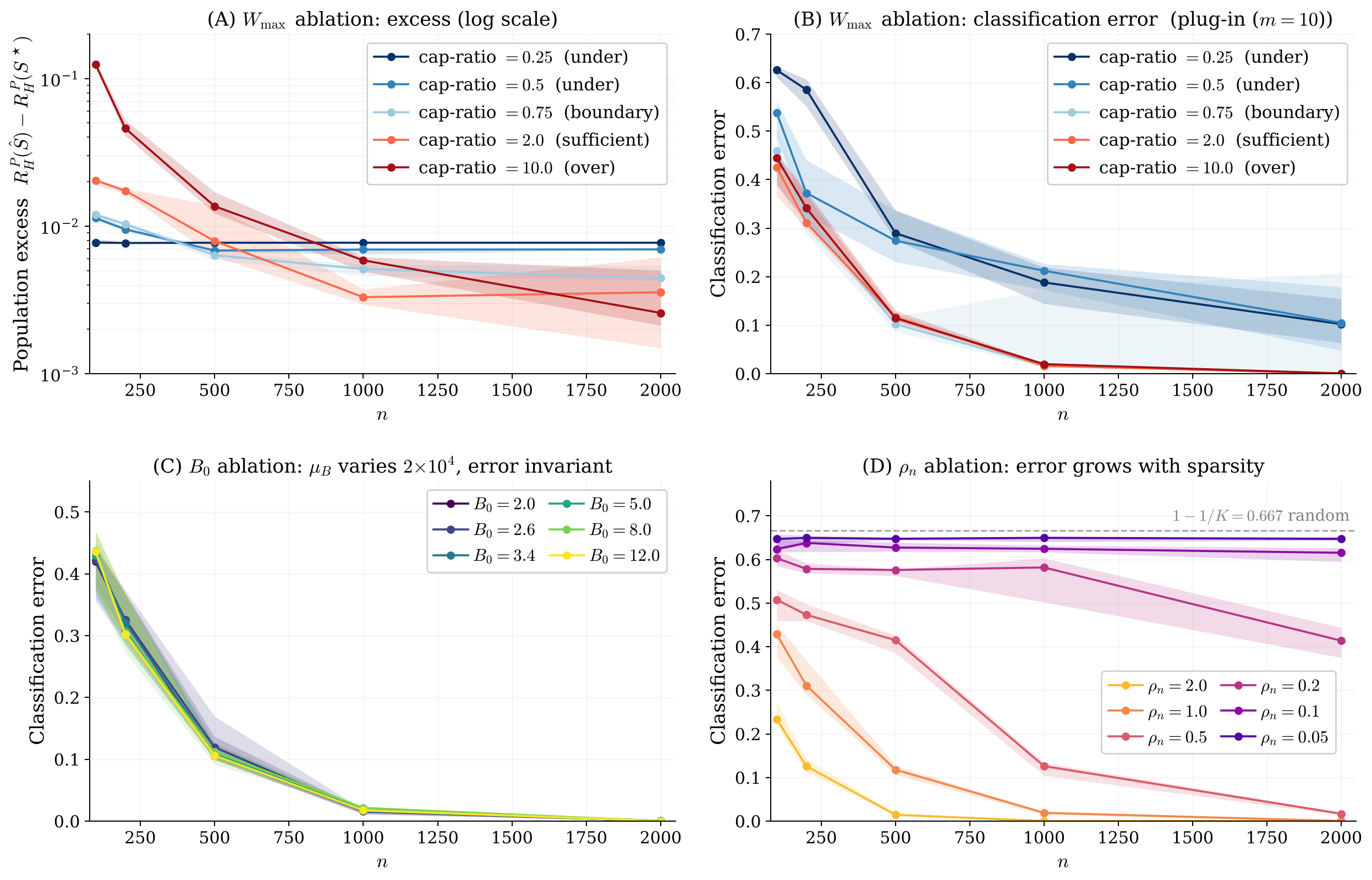}
\caption{\textbf{Ablations of the decoder bounds and the density (GIN+PE, $K\!=\!3$).}
\textbf{(A,B)} $\Wmax$ ablation across cap-ratio $\in\!\{0.25,0.5,0.75,2,10\}$: excess (log scale, panel A) and plug-in classification error ($m\!=\!10$, panel B).
\textbf{(C)} $B_0$ ablation: classification error across $B_0\!\in\!\{2,2.6,3.4,5,8,12\}$.
\textbf{(D)} $\rho_n$ ablation: classification error across $\rho_n\!\in\!\{0.05,0.1,0.2,0.5,1,2\}$.}
\label{fig:ablations}
\vspace{-1em}
\end{figure}

\paragraph{Ablations of the decoder bounds and the density (Fig.~\ref{fig:ablations}).}
The three parameters enter the bound differently: $\Wmax$ sets both the bilinear capacity $\Wmax R^2$ (hence $\eapp$) and the score range $B$ entering $\muB$, $B_0$ bounds the bias and enlarges $B$, and $\rho_n$ changes the SBM itself. \textbf{(A,B)} Cap-ratio $1$ in Prop.~\ref{prop:eapp-rate} is sufficient, and realizability needs at least cap-ratio $(K\!-\!1)/K\!=\!2/3$ (Prop.~\ref{prop:exact-capacity}). Below $2/3$, the lower bound on $\eapp$ is small: $0.0026$ at cap-ratio $0.25$ and $0.0004$ at $0.5$. The observed excess is larger and flat in $n$. For $n\!\ge\!500$ it equals the excess of the constant score $\bhat\!-\!\Wmax R^2$ ($0.0077$ and $0.0070$): the trained bias stays too high, and the bilinear term spends its whole budget lowering every score. Training thus stops at a trivial solution, and the plateau is mainly $\eopt$ (App.~\ref{app:cap-saddle}). From cap-ratio $0.55$ to $0.75$, the share of seeds that recover rises from $1/10$ to $7/10$ at $n\!=\!2000$, and some seeds recover below $2/3$. In our training setup, a small $\Wmax$ therefore acts mainly through optimization. Very large capacity instead hurts at small $n$ ($0.124$ at cap-ratio $10$ vs.\ $0.020$ at $2$, $n\!=\!100$), consistent with $\Gamma_T,\Gamma_{T\Hcal}\!\sim\!B^2\sqrt{d/n}$ in Cor.~\ref{cor:train-side}, and the gap disappears by $n\!=\!2000$. The error curves for cap-ratio $2$ and $10$ nearly coincide although $2/\muB$ changes by $2500\!\times$, so the worst-case curvature factor is not active at the trained solution. \textbf{(C)} Similarly, increasing $B_0$ from $2$ to $12$ inflates $2/\muB$ by $2\!\times\!10^4\!\times$ with no visible effect on error. \textbf{(D)} Varying $\rho_n$ changes $\Deltas$ by only $13\%$ but inflates $2/\muB$ by $32\!\times$. Error is small for $\rho_n\!\ge\!0.5$, partial at $\rho_n\!=\!0.2$ (plug-in error $0.41$, well below random), and near chance for $\rho_n\!\le\!0.1$. Since $\Deltas$ barely changes and (C) shows that a small $\muB$ alone does not hurt error, neither explains the failure. App.~\ref{app:rho-impossible} traces the gap at $\rho_n\!=\!0.2$ partly to optimization, and the failure at $\rho_n\!\le\!0.1$ to the sparse regime, where (A1) in Prop.~\ref{prop:eapp-rate} fails.

\paragraph{Effective vs.\ worst-case Hessian.}
The empirical Step-1 ratio is likewise not governed by the worst-case curvature $\muB$. In Lemma~\ref{lem:step1}, $\muB$ lower-bounds the BCE Hessian uniformly on the whole score range $[-B,B]$, which makes the formal bound conservative. The relevant curvature in practice is instead the curvature along the path between $m^\star_{c_ic_j}$ and $\widehat s_{ij}$. A simple SBM-level proxy is
\begin{equation}\label{eq:mu-eff}
\mu_{\mathrm{eff}}
:=
\E_{(i,j)\sim\Hcal}\bigl[P_{ij}(1-P_{ij})\bigr],
\end{equation}
which depends only on the block probabilities $\{p_{ab}\}$. At the $K=3$ baseline, $\mu_{\mathrm{eff}}\approx0.056$, so $2/\mu_{\mathrm{eff}}\approx36$, against an observed ratio of $38.5$. Across the $\Wmax$ and $B_0$ ablations, $2/\muB$ ranges from $80$ to $2\times10^6$, yet the ratio stays within $36.5$--$42$. When $\rho_n\in[0.1,2]$ moves $2/\mu_{\mathrm{eff}}$ from $19.5$ to $336$, the ratio stays $1$--$16\%$ above it. This small excess is expected: the exact ratio is an error-weighted curvature average, which grows when errors fall more on lower-curvature block pairs (App.~\ref{app:mu-eff}).

\section{Conclusion}\label{sec:conclusion}
 
We proved a finite-sample bound that turns the held-out LP risk of any
bounded GNN encoder with a bilinear decoder into a guarantee for linear
community detection on its embeddings. The guarantee is constructive. A
nearest-centroid rule under the learned Mahalanobis metric
$G\!=\!WZ^\top ZW$, built from a few labeled nodes per community, attains
it, and its label complexity improves with LP quality. On synthetic SBMs,
classification error follows LP risk in most configurations. The empirical
Step-1 ratio stays near the SBM-dependent scale $2/\mu_{\mathrm{eff}}$
rather than the worst-case $2/\muB$, and the Markov factor $\Deltas^{-2}$
is close to tight only near the recovery boundary.
 
Several directions remain open. Showing that gradient-based training
reaches a small LP risk is a separate optimization problem. We realize
$\eapp$ only for GIN with spectral PE. The $K\!=\!4$ collapse of GCN+PE
and GAT+PE (App.~\ref{app:failure-modes}), which change both the
aggregator and the PE, calls for constructions covering other aggregators
and PEs. Sparse SBMs need a different analysis. Finally, our
$\Otilde(n^{-1/2})$ rate is slower than the exponential rates of
SBM-specific procedures \citep{zhang2016minimax,gao2017achieving}. Our bound
depends on the trained model only through its average LP risk. This average
cannot exclude errors concentrated on a few nodes, so the linear dependence
on it is worst-case tight (App.~\ref{app:rate}). Faster rates need
node-level control, and thus a fixed architecture and training algorithm,
both left general by design. Empirically, the error decays faster than
the bound.

\begin{ack}
Yixuan He is supported by a Jetstream2 AI Fellowship~\citep{hancock2021jetstream2} and an NVIDIA Academic Grant Program award. Zhangyang Wang is supported in part by NSF Awards 2145346 (CAREER), 2523383 (DMS), and the NSF AI Institute for Foundations of Machine Learning (IFML).
\end{ack}

\bibliographystyle{plainnat}
\bibliography{refs}

\begin{thebibliography}{46}
\providecommand{\natexlab}[1]{#1}
\providecommand{\url}[1]{\texttt{#1}}
\expandafter\ifx\csname urlstyle\endcsname\relax
  \providecommand{\doi}[1]{doi: #1}\else
  \providecommand{\doi}{doi: \begingroup \urlstyle{rm}\Url}\fi

\bibitem[Abbe(2018)]{abbe2018community}
Emmanuel Abbe.
\newblock Community detection and stochastic block models: recent developments.
\newblock \emph{Journal of Machine Learning Research}, 18\penalty0
  (177):\penalty0 1--86, 2018.

\bibitem[Abbe et~al.(2016)Abbe, Bandeira, and Hall]{abbe2016exact}
Emmanuel Abbe, Afonso~S. Bandeira, and Georgina Hall.
\newblock Exact recovery in the stochastic block model.
\newblock \emph{IEEE Transactions on Information Theory}, 62\penalty0
  (1):\penalty0 471--487, 2016.

\bibitem[Bruna and Li(2017)]{bruna2017community}
Joan Bruna and Xiang Li.
\newblock Community detection with graph neural networks.
\newblock \emph{stat}, 1050:\penalty0 27, 2017.

\bibitem[Cape et~al.(2019)Cape, Tang, and Priebe]{cape2019twotoinfinity}
Joshua Cape, Minh Tang, and Carey~E Priebe.
\newblock The two-to-infinity norm and singular subspace geometry with
  applications to high-dimensional statistics.
\newblock \emph{The Annals of Statistics}, 47\penalty0 (5):\penalty0
  2405--2439, 2019.

\bibitem[Davison et~al.(2024)Davison, Morgan, and Ward]{davison2024community}
Andrew Davison, S~Carlyle Morgan, and Owen~G Ward.
\newblock Community detection guarantees using embeddings learned by node2vec.
\newblock \emph{Advances in Neural Information Processing Systems},
  37:\penalty0 685--738, 2024.

\bibitem[Defferrard et~al.(2016)Defferrard, Bresson, and
  Vandergheynst]{defferrard2016chebnet}
Micha{\"e}l Defferrard, Xavier Bresson, and Pierre Vandergheynst.
\newblock Convolutional neural networks on graphs with fast localized spectral
  filtering.
\newblock \emph{Advances in neural information processing systems (NeurIPS)},
  29, 2016.

\bibitem[Dwivedi and Bresson(2020)]{dwivedi2020generalization}
Vijay~Prakash Dwivedi and Xavier Bresson.
\newblock A generalization of transformer networks to graphs.
\newblock \emph{arXiv preprint arXiv:2012.09699}, 2020.

\bibitem[Gao et~al.(2017)Gao, Ma, Zhang, and Zhou]{gao2017achieving}
Chao Gao, Zongming Ma, Anderson~Y. Zhang, and Harrison~H. Zhou.
\newblock Achieving optimal misclassification proportion in stochastic block
  models.
\newblock \emph{Journal of Machine Learning Research}, 18\penalty0
  (60):\penalty0 1--45, 2017.

\bibitem[Grover and Leskovec(2016)]{grover2016node2vec}
Aditya Grover and Jure Leskovec.
\newblock node2vec: Scalable feature learning for networks.
\newblock In \emph{Proceedings of the 22nd ACM SIGKDD international conference
  on Knowledge discovery and data mining}, pages 855--864, 2016.

\bibitem[Hajek et~al.(2016)Hajek, Wu, and Xu]{hajek2016achieving}
Bruce Hajek, Yihong Wu, and Jiaming Xu.
\newblock Achieving exact cluster recovery threshold via semidefinite
  programming.
\newblock \emph{IEEE Transactions on Information Theory}, 62\penalty0
  (5):\penalty0 2788--2797, 2016.

\bibitem[Hamilton et~al.(2017)Hamilton, Ying, and
  Leskovec]{hamilton2017inductive}
Will Hamilton, Zhitao Ying, and Jure Leskovec.
\newblock Inductive representation learning on large graphs.
\newblock \emph{Advances in Neural Information Processing Systems (NeurIPS)},
  30, 2017.

\bibitem[Hancock et~al.(2021)Hancock, Fischer, Lowe, Snapp-Childs, Pierce,
  Marru, Coulter, Vaughn, Beck, Merchant, Skidmore, and
  Jacobs]{hancock2021jetstream2}
David~Y. Hancock, Jeremy Fischer, John~Michael Lowe, Winona Snapp-Childs,
  Marlon Pierce, Suresh Marru, J.~Eric Coulter, Matthew Vaughn, Brian Beck,
  Nirav Merchant, Edwin Skidmore, and Gwen Jacobs.
\newblock Jetstream2: Accelerating cloud computing via {Jetstream}.
\newblock In \emph{Practice and Experience in Advanced Research Computing
  (PEARC '21)}, pages 1--8. ACM, 2021.
\newblock \doi{10.1145/3437359.3465565}.

\bibitem[HaoChen et~al.(2021)HaoChen, Wei, Gaidon, and Ma]{haochen2021provable}
Jeff~Z HaoChen, Colin Wei, Adrien Gaidon, and Tengyu Ma.
\newblock Provable guarantees for self-supervised deep learning with spectral
  contrastive loss.
\newblock \emph{Advances in neural information processing systems (NeurIPS)},
  34:\penalty0 5000--5011, 2021.

\bibitem[Hastie(2009)]{hastie2009elements}
Trevor Hastie.
\newblock The elements of statistical learning: data mining, inference, and
  prediction, 2009.

\bibitem[Hoeffding(1963)]{hoeffding1963probability}
Wassily Hoeffding.
\newblock Probability inequalities for sums of bounded random variables.
\newblock \emph{Journal of the American statistical association}, 58\penalty0
  (301):\penalty0 13--30, 1963.

\bibitem[Holland et~al.(1983)Holland, Laskey, and
  Leinhardt]{holland1983stochastic}
Paul~W Holland, Kathryn~Blackmond Laskey, and Samuel Leinhardt.
\newblock Stochastic blockmodels: First steps.
\newblock \emph{Social networks}, 5\penalty0 (2):\penalty0 109--137, 1983.

\bibitem[Hu et~al.(2020)Hu, Dong, Wang, Chang, and Sun]{hu2020gptgnn}
Ziniu Hu, Yuxiao Dong, Kuansan Wang, Kai-Wei Chang, and Yizhou Sun.
\newblock Gpt-gnn: Generative pre-training of graph neural networks.
\newblock In \emph{Proceedings of the 26th ACM SIGKDD international conference
  on knowledge discovery \& data mining (KDD)}, pages 1857--1867, 2020.

\bibitem[Joag-Dev and Proschan(1983)]{joag1983negative}
Kumar Joag-Dev and Frank Proschan.
\newblock Negative association of random variables with applications.
\newblock \emph{The Annals of Statistics}, pages 286--295, 1983.

\bibitem[Kipf and Welling(2016{\natexlab{a}})]{kipf2016variational}
Thomas~N. Kipf and Max Welling.
\newblock Variational graph auto-encoders.
\newblock \emph{CoRR}, abs/1611.07308, 2016{\natexlab{a}}.
\newblock URL \url{http://arxiv.org/abs/1611.07308}.

\bibitem[Kipf and Welling(2016{\natexlab{b}})]{kipf2017gcn}
Thomas~N Kipf and Max Welling.
\newblock Semi-supervised classification with graph convolutional networks.
\newblock \emph{arXiv preprint arXiv:1609.02907}, 2016{\natexlab{b}}.

\bibitem[Lee et~al.(2021)Lee, Lei, Saunshi, and Zhuo]{lee2021predicting}
Jason~D Lee, Qi~Lei, Nikunj Saunshi, and Jiacheng Zhuo.
\newblock Predicting what you already know helps: Provable self-supervised
  learning.
\newblock \emph{Advances in Neural Information Processing Systems (NeurIPS)},
  34:\penalty0 309--323, 2021.

\bibitem[Lei and Rinaldo(2015)]{lei2015consistency}
Jing Lei and Alessandro Rinaldo.
\newblock Consistency of spectral clustering in stochastic block models.
\newblock \emph{The Annals of Statistics}, pages 215--237, 2015.

\bibitem[Liu et~al.(2025)Liu, Yang, Lu, Chen, Li, Zhang, Bai, Fang, Sun, Yu,
  et~al.]{liu2025gfmsurvey}
Jiawei Liu, Cheng Yang, Zhiyuan Lu, Junze Chen, Yibo Li, Mengmei Zhang, Ting
  Bai, Yuan Fang, Lichao Sun, Philip~S Yu, et~al.
\newblock Graph foundation models: Concepts, opportunities and challenges.
\newblock \emph{IEEE Transactions on Pattern Analysis and Machine
  Intelligence}, 2025.

\bibitem[Liu et~al.(2023)Liu, Yu, Fang, and Zhang]{liu2023graphprompt}
Zemin Liu, Xingtong Yu, Yuan Fang, and Xinming Zhang.
\newblock {GraphPrompt}: Unifying pre-training and downstream tasks for graph
  neural networks.
\newblock In \emph{Proceedings of the ACM web conference 2023}, pages 417--428,
  2023.

\bibitem[Loshchilov and Hutter(2019)]{loshchilov2019decoupled}
Ilya Loshchilov and Frank Hutter.
\newblock Decoupled weight decay regularization.
\newblock In \emph{International Conference on Learning Representations}, 2019.

\bibitem[Lyzinski et~al.(2014)Lyzinski, Sussman, Tang, Athreya, and
  Priebe]{lyzinski2014perfect}
Vince Lyzinski, Daniel~L. Sussman, Minh Tang, Avanti Athreya, and Carey~E.
  Priebe.
\newblock Perfect clustering for stochastic blockmodel graphs via adjacency
  spectral embedding.
\newblock \emph{Electronic Journal of Statistics}, 8\penalty0 (2):\penalty0
  2905--2922, 2014.

\bibitem[Mehta et~al.(2019)Mehta, Duke, and Rai]{mehta2019stochastic}
Nikhil Mehta, Lawrence~Carin Duke, and Piyush Rai.
\newblock Stochastic blockmodels meet graph neural networks.
\newblock In \emph{International Conference on Machine Learning (ICML)}, pages
  4466--4474. PMLR, 2019.

\bibitem[Mensink et~al.(2013)Mensink, Verbeek, Perronnin, and
  Csurka]{mensink2013distance}
Thomas Mensink, Jakob Verbeek, Florent Perronnin, and Gabriela Csurka.
\newblock Distance-based image classification: Generalizing to new classes at
  near-zero cost.
\newblock \emph{IEEE Transactions on Pattern Analysis and Machine
  Intelligence}, 35\penalty0 (11):\penalty0 2624--2637, 2013.

\bibitem[Mossel et~al.(2015)Mossel, Neeman, and Sly]{mossel2015consistency}
Elchanan Mossel, Joe Neeman, and Allan Sly.
\newblock Consistency thresholds for the planted bisection model.
\newblock In \emph{Proceedings of the 47th Annual ACM Symposium on Theory of
  Computing (STOC)}, pages 69--75, 2015.

\bibitem[Rubin-Delanchy et~al.(2022)Rubin-Delanchy, Cape, Tang, and
  Priebe]{rubin2022statistical}
Patrick Rubin-Delanchy, Joshua Cape, Minh Tang, and Carey~E Priebe.
\newblock A statistical interpretation of spectral embedding: the generalised
  random dot product graph.
\newblock \emph{Journal of the Royal Statistical Society Series B: Statistical
  Methodology}, 84\penalty0 (4):\penalty0 1446--1473, 2022.

\bibitem[Saunshi et~al.(2019)Saunshi, Plevrakis, Arora, Khodak, and
  Khandeparkar]{arora2019theoretical}
Nikunj Saunshi, Orestis Plevrakis, Sanjeev Arora, Mikhail Khodak, and
  Hrishikesh Khandeparkar.
\newblock A theoretical analysis of contrastive unsupervised representation
  learning.
\newblock In \emph{International conference on machine learning (ICML)}, pages
  5628--5637. PMLR, 2019.

\bibitem[Serfling(1974)]{serfling1974probability}
Robert~J Serfling.
\newblock Probability inequalities for the sum in sampling without replacement.
\newblock \emph{The Annals of Statistics}, pages 39--48, 1974.

\bibitem[Snell et~al.(2017)Snell, Swersky, and Zemel]{snell2017prototypical}
Jake Snell, Kevin Swersky, and Richard Zemel.
\newblock Prototypical networks for few-shot learning.
\newblock In \emph{Advances in Neural Information Processing Systems}, pages
  4077--4087, 2017.

\bibitem[Snijders and Nowicki(1997)]{snijders1997estimation}
Tom A.~B. Snijders and Krzysztof Nowicki.
\newblock Estimation and prediction for stochastic blockmodels for graphs with
  latent block structure.
\newblock \emph{Journal of Classification}, 14:\penalty0 75--100, 1997.

\bibitem[Sun et~al.(2022)Sun, Zhou, He, Wang, and Wang]{sun2022gppt}
Mingchen Sun, Kaixiong Zhou, Xin He, Ying Wang, and Xin Wang.
\newblock {GPPT}: Graph pre-training and prompt tuning to generalize graph
  neural networks.
\newblock In \emph{Proceedings of the 28th ACM SIGKDD conference on knowledge
  discovery and data mining (KDD)}, pages 1717--1727, 2022.

\bibitem[Sussman et~al.(2012)Sussman, Tang, Fishkind, and
  Priebe]{sussman2012consistent}
Daniel~L Sussman, Minh Tang, Donniell~E Fishkind, and Carey~E Priebe.
\newblock A consistent adjacency spectral embedding for stochastic blockmodel
  graphs.
\newblock \emph{Journal of the American Statistical Association}, 107\penalty0
  (499):\penalty0 1119--1128, 2012.

\bibitem[Tosh et~al.(2021)Tosh, Krishnamurthy, and Hsu]{tosh2021contrastive}
Christopher Tosh, Akshay Krishnamurthy, and Daniel Hsu.
\newblock Contrastive learning, multi-view redundancy, and linear models.
\newblock In \emph{Algorithmic learning theory (ALT)}, pages 1179--1206. PMLR,
  2021.

\bibitem[Tropp(2012)]{tropp2012user}
Joel~A Tropp.
\newblock User-friendly tail bounds for sums of random matrices.
\newblock \emph{Foundations of computational mathematics}, 12\penalty0
  (4):\penalty0 389--434, 2012.

\bibitem[Veli{\v{c}}kovi{\'c} et~al.(2018)Veli{\v{c}}kovi{\'c}, Cucurull,
  Casanova, Romero, Li{\`o}, and Bengio]{velickovic2018gat}
Petar Veli{\v{c}}kovi{\'c}, Guillem Cucurull, Arantxa Casanova, Adriana Romero,
  Pietro Li{\`o}, and Yoshua Bengio.
\newblock Graph attention networks.
\newblock In \emph{International Conference on Learning Representations
  (ICLR)}, 2018.

\bibitem[Velickovic et~al.(2018)Velickovic, Fedus, Hamilton, Li{\`o}, Bengio,
  and Hjelm]{velickovic2019deep}
Petar Velickovic, William Fedus, William~L Hamilton, Pietro Li{\`o}, Yoshua
  Bengio, and R~Devon Hjelm.
\newblock Deep graph infomax.
\newblock \emph{stat}, 1050:\penalty0 21, 2018.

\bibitem[Wedin(1972)]{wedin1972perturbation}
Per-{\AA}ke Wedin.
\newblock Perturbation bounds in connection with singular value decomposition.
\newblock \emph{BIT Numerical Mathematics}, 12\penalty0 (1):\penalty0 99--111,
  1972.

\bibitem[Wei et~al.(2021)Wei, Xie, and Ma]{wei2021why}
Colin Wei, Sang~Michael Xie, and Tengyu Ma.
\newblock Why do pretrained language models help in downstream tasks? an
  analysis of head and prompt tuning.
\newblock \emph{Advances in Neural Information Processing Systems (NeurIPS)},
  34:\penalty0 16158--16170, 2021.

\bibitem[Xu et~al.(2019)Xu, Hu, Leskovec, and Jegelka]{xu2019gin}
Keyulu Xu, Weihua Hu, Jure Leskovec, and Stefanie Jegelka.
\newblock How powerful are graph neural networks?
\newblock In \emph{International Conference on Learning Representations
  (ICLR)}, 2019.

\bibitem[You et~al.(2020)You, Chen, Sui, Chen, Wang, and Shen]{you2020graphcl}
Yuning You, Tianlong Chen, Yongduo Sui, Ting Chen, Zhangyang Wang, and Yang
  Shen.
\newblock Graph contrastive learning with augmentations.
\newblock \emph{Advances in neural information processing systems (NeurIPS)},
  33:\penalty0 5812--5823, 2020.

\bibitem[Zhang and Zhou(2016)]{zhang2016minimax}
Anderson~Y. Zhang and Harrison~H. Zhou.
\newblock Minimax rates of community detection in stochastic block models.
\newblock \emph{The Annals of Statistics}, 44\penalty0 (5):\penalty0
  2252--2280, 2016.

\bibitem[Zhu et~al.(2020)Zhu, Xu, Yu, Liu, Wu, and Wang]{zhu2020deep}
Yanqiao Zhu, Yichen Xu, Feng Yu, Qiang Liu, Shu Wu, and Liang Wang.
\newblock Deep graph contrastive representation learning.
\newblock \emph{arXiv preprint arXiv:2006.04131}, 2020.

\end{thebibliography}

\clearpage

\appendix

\section{Notation Table}\label{app:notation}

This appendix collects the symbols used throughout the paper, grouped by role.

\begin{longtable}{l@{\quad}p{0.6\linewidth}}
\toprule
Symbol & Description \\
\midrule
\endfirsthead
\toprule
Symbol & Description \\
\midrule
\endhead
\bottomrule
\endfoot
\multicolumn{2}{l}{\textit{Graph and stochastic block model}}\\
\midrule
$\mathcal{G}=(V,E)$ & undirected, unweighted graph; $V=[n]:=\{1,\dots,n\}$, $E\subseteq\binom{V}{2}$ \\
$n$, $K$ & number of nodes, number of communities \\
$A\in\{0,1\}^{n\times n}$ & symmetric adjacency matrix, $A_{ii}=0$ \\
$c_i\in[K]$ & (unobserved) community label of node $i$ \\
$\pi_a$, $\pi_{\min}$ & community proportion, minimum proportion across $a\in[K]$ \\
$p_{ab}\in(0,1)$ & SBM block probability for communities $a,b$ \\
$\Theta=(p_{ab})_{a,b\in[K]}\in(0,1)^{K\times K}$ & symmetric block probability matrix \\
$P=(P_{ij})_{i,j\in[n]}$, $P_{ij}=p_{c_ic_j}$ & edge-probability matrix (diagonal included) \\
$N=\binom{n}{2}$ & total number of node pairs \\
$\mathcal{C}_a:=\{i:c_i=a\}$ & set of nodes in community $a$ \\
\midrule
\multicolumn{2}{l}{\textit{Masking and pair sets}}\\
\midrule
$\tau\in(0,1)$ & held-out fraction \\
$\Hcal$, $\Tcal$ & held-out and visible (train) pair sets, $|\Hcal|=\lceil\tau N\rceil$ \\
$\Aenc$ & masked adjacency input to encoder; $\Aenc_{ij}=A_{ij}\indic\{(i,j)\in\Tcal\}$ \\
\midrule
\multicolumn{2}{l}{\textit{Encoder, decoder, and capacity}}\\
\midrule
$f_\theta$ & GNN encoder (any bounded family) \\
$z_i:=f_\theta(\Aenc,i)\in\Real^d$ & node embedding \\
$Z\in\Real^{n\times d}$ & embedding matrix with rows $z_i^\top$ \\
$d$, $R$ & embedding dimension, $\ell_2$-ball radius ($\|z_i\|\le R$) \\
$W=W^\top\in\Real^{d\times d}$, $\Wmax$ & symmetric bilinear weight; $\|W\|_{\op}\le\Wmax$ \\
$\bhat\in\Real$, $B_0$ & decoder bias and its cap, $|\bhat|\le B_0$ \\
$\Wmax R^2$ & bilinear capacity (Sec.~\ref{sec:setting}) \\
$s_{ij}=\bhat+z_i^\top W z_j$ & decoder score for pair $(i,j)$ \\
$m^\star_{ab}=\logit(p_{ab})$ & oracle community-level score \\
$M^\star=(m^\star_{ab})_{a,b\in[K]}$ & oracle score matrix at community level \\
$S=(s_{ij})_{i,j\in[n]}$, $\Sopt$, $\Shat$ & score matrix; oracle $\Sopt:=(m^\star_{c_ic_j})_{i,j\in[n]}$, trained \\
$\Scal(A,\Tcal)$ & class of score matrices realized by the bounded encoder--decoder family \\
$\Zcal$, $\Scal^{\Zcal}$ & \raggedright deterministic envelopes of all norm-bounded $(Z,W,b)$ and their score matrices (App.~\ref{app:step0}, \ref{app:train-side}) \tabularnewline
$\Bmodel$, $B^\star$ & learned-score bound $B_0+\Wmax R^2$; oracle score bound $\max_{a,b}|m^\star_{ab}|$ \\
$B$ & score-range bound, $B=\max(\Bmodel,\ B^\star,\ \log 2)$ \\
$b^\star$ & bias minimizing $\|M^\star-b\,\mathbf 1\mathbf 1^\top\|_{\op}$ (App.~\ref{app:decoder-real}, Sec.~\ref{sec:experiments}) \\
\midrule
\multicolumn{2}{l}{\textit{Loss, risks, and residuals}}\\
\midrule
$\sigma(s)=(1+e^{-s})^{-1}$ & sigmoid \\
$\ell(s;p)$ & per-pair BCE, $-p\log\sigma(s)-(1-p)\log(1-\sigma(s))$ \\
$\widehat R_\Tcal,\,R^\Pop_\Tcal,\,\widehat R_\Hcal,\,R^\Pop_\Hcal$ & training / held-out empirical and population risks \\
$\Rdag$, $\Rempdag$ & in-class population risk infimum on $\Hcal$; in-class empirical risk infimum on $\Tcal$ \\
$\eopt$ & in-class population optimization residual on $\Hcal$, $\Rpop(\Shat)-\Rdag$ \\
$\eapp$ & architectural approximation residual on $\Hcal$, $\Rdag-\Rpop(\Sopt)$ \\
$\widehat\rho$ & empirical training residual, $\widehat R_\Tcal(\Shat)-\Rempdag$ \\
$\mathrm{ScoreMSE}_\Hcal$ & mean of $(s_{ij}-m^\star_{c_ic_j})^2$ over $\Hcal$ \\
$\mathrm{ScoreMSE}_{\mathrm{full}}$ & mean of $(s_{ij}-m^\star_{c_ic_j})^2$ over all $i,j\in[n]$ (with $s_{ii}$) \\
\midrule
\multicolumn{2}{l}{\textit{Statistical deviations}}\\
\midrule
$\GammaH$ & held-out edge-noise concentration (Step~0) \\
$\GammaT,\GammaTH$ & train-side and cross-side uniform deviations (Cor.~\ref{cor:train-side}) \\
$\Vstat$ & held-out $\to$ full statistical bound (Step~2) \\
$\En$ & bound on $\mathrm{ScoreMSE}_{\mathrm{full}}$, $\tfrac{2}{\muB}(\eopt+\eapp)+\Vstat$ \\
$\log(nBR\Wmax)$ & \raggedright logarithmic factor, with $R,\Wmax$ read as $\max(R,1),\max(\Wmax,1)$ \tabularnewline
\midrule
\multicolumn{2}{l}{\textit{Geometry and curvature}}\\
\midrule
$\muB=\sigma(B)(1-\sigma(B))$ & worst-case BCE strong-convexity constant on $[-B,B]$ \\
$\mu_{\mathrm{eff}}=\E_{(i,j)\sim\Hcal}[P_{ij}(1-P_{ij})]$ & \raggedright typical BCE curvature at the oracle scores; $2/\mu_{\mathrm{eff}}$ tracks the Step-1 ratio (Sec.~\ref{sec:experiments}, App.~\ref{app:mu-eff}) \tabularnewline
$\Deltas$ & score-profile separation, $\Deltas^2=\min_{a\ne b}\sum_c\pi_c(m^\star_{ac}-m^\star_{bc})^2$ \\
\midrule
\multicolumn{2}{l}{\textit{Classifiers, prototypes, labels}}\\
\midrule
$\tilde s_i$, $\tilde e_i$ & score profile $(s_{ij})_{j\in[n]}$ of node $i$; its error $n^{-1/2}\|\tilde s_i-\mu_{c_i}\|$ \\
$\mu_a,\,\hat\mu_a$ & score-profile prototype: oracle $(m^\star_{a,c_j})_{j\in[n]}$, empirical \\
$\bar z_a=|L_a|^{-1}\sum_{i\in L_a}z_i$ & empirical embedding centroid of community $a$ \\
$h^\star$ & oracle classifier (Prop.~\ref{prop:step3}) \\
$\hat h^{\mathrm{plug}}$ & plug-in classifier (Thm.~\ref{thm:main}~Part~II) \\
$G=WZ^\top ZW\in\Real^{d\times d}$ & PSD Mahalanobis form (Prop.~\ref{prop:mahalanobis}) \\
$\kappa$ & \raggedright approximation factor of a $(1+\kappa)$-approximate $K$-means solution (App.~\ref{app:kmeans}) \tabularnewline
$L=\bigcup_{a=1}^K L_a$ & labeled set; $L_a\subset\mathcal{C}_a$ uniform without replacement \\
$m_a=|L_a|$, $m_{\min}$, $|L|$ & \raggedright per-community label count, minimum, total $|L|=\sum_a m_a$; in experiments, $m$ is the number of labels per community \tabularnewline
$\eta_L$ & prototype-estimation error, $\max_a n^{-1/2}\|\hat\mu_a-\mu_a\|$ \\
\midrule
\multicolumn{2}{l}{\textit{Spectral-PE GIN (Prop.~\ref{prop:eapp-rate}) and experiments}}\\
\midrule
$L$, $D_{\mathrm{PE}}$ & \raggedright number of GIN layers (in Prop.~\ref{prop:eapp-rate} and App.~\ref{app:eapp} only), PE dimension \tabularnewline
$p_{\min}$, $\tau_0$ & density floor in (A1), masking margin in (A4) \\
$\lambda^\star_k$, $c_\gamma$ & \raggedright nonzero eigenvalues of $P$, eigenvalue scale in (A3), $|\lambda^\star_k|\ge c_\gamma n$ \tabularnewline
$p_{\mathrm{in}}$, $p_{\mathrm{out}}$, $\rho_n$ & \raggedright within- and between-community probabilities, density multiplier (Sec.~\ref{sec:experiments}) \tabularnewline
$\mathrm{cap}$, cap-ratio & $\min_{|b|\le B_0}\|M^\star-b\,\mathbf 1\mathbf 1^\top\|_{\op}$; $\Wmax R^2/\mathrm{cap}$ (Sec.~\ref{sec:experiments}) \\
\midrule
\multicolumn{2}{l}{\textit{Abbreviations}}\\
\midrule
ASE & adjacency spectral embedding \\
BCE & binary cross-entropy \\
CP & centered-power encoder (App.~\ref{app:hyperparams}) \\
CV & coefficient of variation \\
ERM & empirical risk minimization \\
GAT & graph attention network \\
GCN & graph convolutional network \\
GIN & graph isomorphism network \\
GNN & graph neural network \\
IQR & interquartile range \\
KL & Kullback--Leibler divergence \\
LogReg & logistic regression \\
LP & link prediction \\
MLP & multilayer perceptron \\
MPNN & message-passing neural network \\
MSE & mean squared error \\
PE & \raggedright positional encoding; spectral PE for GIN, Laplacian PE for GCN and GAT (App.~\ref{app:hyperparams}) \tabularnewline
PSD & positive semidefinite \\
ReLU & rectified linear unit \\
SBM & stochastic block model \\
SVD & singular value decomposition \\
X+PE, X+random & \raggedright encoder X with PE or i.i.d.\ Gaussian input features (App.~\ref{app:hyperparams}) \tabularnewline
\end{longtable}

\section{Architectural Scope and Decoder Realizability}\label{app:setting}

\subsection{Architectural Scope}\label{app:arch-scope}

\paragraph{Encoder and decoder admissibility.} Any GNN family (message-passing, attention-based, spectral) is admissible for $f_\theta$, provided the boundedness $\|z_i\|\le R$ of Assumption~\ref{ass:bounded} holds; a final normalization or projection layer suffices when needed. The decoder parameter $W$ is jointly learnable with $(\theta,\bhat)$ and bounded by $\|W\|_{\mathrm{op}}\le\Wmax$. Beyond $W=W^\top$, no sign constraint is imposed, so $W$ may be indefinite.

\paragraph{Feature conventions.} Auxiliary inputs to $f_\theta$ are admissible under two conventions, both keeping held-out outcomes out of the encoder:
\begin{itemize}
\item \emph{Graph-derived features.} Constant features ($X=\mathbf 1$) and graph-derived features (Laplacian or random-walk positional encodings) must be computed from $\Aenc$, not from $A$. They are then measurable in $(\Aenc,\Tcal)$ and may be absorbed into the first layer of $f_\theta$ without changing the notation $z_i=f_\theta(\Aenc,i)$.
\item \emph{Random features.} Random features $X$ independent of $(A,\Hcal)$ are treated as sampled once and held fixed; probability statements are conditional on $X$ (equivalently, $X$ is absorbed into $\theta$).
\end{itemize}

\begin{proposition}[Held-out independence]\label{prop:held-out-indep}
Under the conventions above, $\{A_{ij}:(i,j)\in\Hcal\}$ is independent of $\Aenc$ given $\Hcal$.
\end{proposition}

\begin{proof}
Given $\Hcal$, the partition $(\Tcal,\Hcal)$ is deterministic and $\Aenc$ is measurable in $\{A_{ij}:(i,j)\in\Tcal\}$. Disjointness of $\Tcal,\Hcal$ and independence of the underlying Bernoullis give the claim. Graph-derived features (measurable in $(\Aenc,\Tcal)$) and fixed random features add no dependence on $\{A_{ij}:(i,j)\in\Hcal\}$.
\end{proof}

Proposition~\ref{prop:held-out-indep} is a property of the training protocol: the encoder never sees held-out outcomes. The concentration steps do not rely on it. Steps~0 and~2 (Lemmas~\ref{lem:step0} and~\ref{lem:step2}) hold uniformly over the deterministic class $\Zcal$ of all $(Z,W,b)$ obeying the bounds of Assumption~\ref{ass:bounded} (App.~\ref{app:step0}), and use only that $\Hcal$ is drawn uniformly and independently of $A$. The resulting event therefore covers the trained model however it depends on $(A,\Hcal)$, including optimizer and model-selection randomness, which enter only through $\eopt$. Architectural choices (GNN family, depth, hidden width, features) leave the form of the main theorem unchanged: they enter through $\eopt$ and $\eapp$, while the embedding dimension $d$ and the norm budgets $(R,\Wmax,B_0)$ also enter through $\Vstat$ and $B$.

\subsection{Decoder Realizability}\label{app:decoder-real}

We compare the inner-product$+$bias and general bilinear decoders at the $K\times K$ community level: do there exist $\{z_a\}_{a\in[K]}$ and $(W,b)$ with $z_a^\top W z_b + b = M^\star_{ab}$ for every $(a,b)$? A negative answer gives a structural floor on $\eapp$ that no encoder can erase for large $n$. The Bayes-optimal scores $\logit(p_{ij}) = M^\star_{c_i c_j}$ depend only on the community pair. Let $\zeta_a$ be the mean embedding of community $a$, so $\|\zeta_a\|\le R$. Averaging $s_{ij} = \bhat + z_i^\top W z_j$ over $\mathcal C_a\times\mathcal C_b$ gives $\bhat+\zeta_a^\top W\zeta_b$, up to $O(\Wmax R^2/(n\pi_{\min}))$ from the diagonal terms. Scores that match $M^\star$ therefore force an approximate community-level realization. Inner-product$+$bias and general bilinear give different answers at $K\ge 3$.

\begin{proposition}[Decoder realizability: inner-product$+$bias vs.\ general bilinear]\label{prop:bilinear-realizability}
Let $M^\star=(\logit(p_{ab}))_{a,b\in[K]}$.
\begin{enumerate}
\item[\emph{(a)}] \emph{(Inner-product$+$bias is locked to PSD on $\mathbf 1^\perp$.)} If $M^\star = UU^\top + b\,\mathbf 1\mathbf 1^\top$ for some $U\in\Real^{K\times d}$ and $b\in\Real$, then $x^\top M^\star x \ge 0$ for every $x\perp\mathbf 1$. Consequently the obstruction
\begin{equation}\label{eq:ip-obstruction}
\exists\, x\perp\mathbf 1\ \text{with}\ x^\top M^\star x < 0
\end{equation}
forbids exact inner-product$+$bias realization, regardless of $d$ and $b$.

\item[\emph{(b)}] \emph{(The obstruction is realized at every $K\ge 3$.)} The hub SBM $p_{11}=p_{22}=p_{13}=0.1$, $p_{33}=0.2$, $p_{12}=p_{23}=0.9$ (full-rank, hence compatible with the standing assumption $\mathrm{rank}(\Theta)=K$) satisfies~\eqref{eq:ip-obstruction} via $x=(1,-2,1)^\top$, with $x^\top M^\star x = -15\log 9 - 2\log 2 < 0$. For $K>3$, padding $x$ with zeros extends the witness to any larger community count.

\item[\emph{(c)}] \emph{(General bilinear realizes $M^\star$ under the bilinear capacity condition.)} For every symmetric $M^\star$ and every $d\ge K$, there exist $Z\in\Real^{K\times d}$ with $\|z_a\|\le R$, symmetric $W\in\Real^{d\times d}$ with $\|W\|_{\mathrm{op}}\le\Wmax$, and $|b|\le B_0$ such that $M^\star = ZWZ^\top + b\,\mathbf 1\mathbf 1^\top$, provided
\begin{equation}\label{eq:bilinear-capacity}
\Wmax R^2 \;\ge\; \min_{|b|\le B_0}\bigl\|M^\star - b\,\mathbf 1\mathbf 1^\top\bigr\|_{\mathrm{op}}.
\end{equation}
\end{enumerate}
\end{proposition}

\begin{proof}
\emph{(a).} For $x\perp\mathbf 1$,
\[
x^\top M^\star x \;=\; \|U^\top x\|^2 + b\,(x^\top\mathbf 1)^2 \;=\; \|U^\top x\|^2 \;\ge\; 0;
\]
the contrapositive proves the obstruction direction.

\emph{(b).} Direct expansion at $x=(1,-2,1)^\top$, using $\logit(0.1)=-\log 9$, $\logit(0.2)=-\log 4$, and $\logit(0.9)=\log 9$, gives $x^\top M^\star x = 7\logit(0.1)+\logit(0.2)-8\logit(0.9) = -15\log 9 - 2\log 2$. Padding $x$ with zeros to length $K>3$ leaves the value unchanged, since entries outside the top-left $3\times 3$ block do not contribute.

\emph{(c).} Pick $b^\star\in\arg\min_{|b|\le B_0}\|M^\star-b\,\mathbf 1\mathbf 1^\top\|_{\mathrm{op}}$ (attained, since the objective is continuous on a compact interval) and set $\widetilde M:=M^\star-b^\star\mathbf 1\mathbf 1^\top$. Spectrally decompose $\widetilde M=\sum_{k=1}^K\lambda_k v_k v_k^\top$ with orthonormal $\{v_k\}\subset\Real^K$. Form $Z\in\Real^{K\times d}$ by stacking $(v_1,\ldots,v_K)$ into the first $K$ columns and zeros elsewhere; set $W=\mathrm{diag}(\lambda_1,\ldots,\lambda_K,0,\ldots,0)$ and $b=b^\star$. Then $ZWZ^\top+b\,\mathbf 1\mathbf 1^\top = M^\star$ with $\|z_a\|\le 1$ and $\|W\|_{\mathrm{op}}=\|\widetilde M\|_{\mathrm{op}}$. The rescaling $(Z,W)\mapsto(\alpha Z, W/\alpha^2)$ preserves the identity and redistributes the norm budgets to any $(R,\Wmax)$ satisfying~\eqref{eq:bilinear-capacity}.
\end{proof}

\paragraph{Geometric picture.} Inner-product$+$bias decomposes scores as $UU^\top+b\,\mathbf 1\mathbf 1^\top$: a PSD term plus a rank-one correction supported on $\mathbf 1$. The bias acts only on $\mathbf 1$; the action on $\mathbf 1^\perp$ is locked to PSD, independently of $b$. At $K=2$, $\mathbf 1^\perp$ is one-dimensional, and any assortative regime ($p_{aa}>p_{12}$) gives $\logit(p_{11})+\logit(p_{22})-2\logit(p_{12})>0$ on the single direction $(1,-1)$, so the restriction is never active. At $K\ge 3$, $\mathbf 1^\perp$ has dimension $\ge 2$ and $M^\star|_{\mathbf 1^\perp}$ can be indefinite; the hub SBM in (b) realizes this. General bilinear lifts the PSD restriction by allowing an indefinite symmetric $W$: $ZWZ^\top$ can have eigenvalues of either sign. The remaining requirement is capacity: by~\eqref{eq:bilinear-capacity}, it suffices that $R^2\Wmax$ dominates $\|M^\star-b\,\mathbf 1\mathbf 1^\top\|_{\mathrm{op}}$ for some admissible bias $b$. Bilinear is therefore not free: it pays for indefinite expressivity with operator-norm capacity.

\paragraph{Architectural implication.} For any $K\ge 3$ SBM satisfying~\eqref{eq:ip-obstruction}, inner-product$+$bias has a decoder-level floor on $\eapp$ that no GNN encoder can erase, whatever its depth, width, or expressivity. General bilinear removes this floor under the bilinear capacity condition~\eqref{eq:bilinear-capacity}. The only remaining architectural source of $\eapp$ is then the encoder, a GNN approximation question treated in App.~\ref{app:eapp}. We adopt the bilinear form for this reason: it separates the structural decoder limit from encoder expressivity.

\begin{remark}[Operator vs.\ entrywise norm in~\eqref{eq:bilinear-capacity}]\label{rmk:op-vs-entrywise}
The bound~\eqref{eq:bilinear-capacity} is in operator norm, while the score range $B$ (Sec.~\ref{sec:setting}) controls $M^\star$ entrywise via $B^\star=\|M^\star\|_{\max}$; the two differ by at most a factor of $K$. A sufficient entrywise replacement (take $b=0$ in~\eqref{eq:bilinear-capacity}) is
\[
\Wmax R^2 \;\ge\; KB^\star,
\]
which is easy to check but typically loose for structured $M^\star$: the symmetric planted partition $M^\star=(\alpha-\beta)I_K+\beta\,\mathbf 1\mathbf 1^\top$ admits $b^\star=\beta$ (when $B_0\ge|\beta|$) and $\|M^\star-b^\star\,\mathbf 1\mathbf 1^\top\|_{\mathrm{op}}=|\alpha-\beta|$, with no $K$ factor. Capacity enters the bounds through $\eapp$ and through the score range $B$.
\end{remark}
\paragraph{A necessary condition for symmetric SBMs.}
Condition~\eqref{eq:bilinear-capacity} is sufficient. For symmetric SBMs, a trace argument also gives a necessary condition, and a floor on $\eapp$ when it fails.

\begin{proposition}[Necessary capacity for symmetric SBMs]\label{prop:exact-capacity}
Let $p_{ab}\!=\!p_{\mathrm{in}}$ for $a\!=\!b$ and $p_{ab}\!=\!p_{\mathrm{out}}$ for $a\!\ne\!b$, with $p_{\mathrm{in}}\!>\!p_{\mathrm{out}}$. Write $\gamma:=\logit p_{\mathrm{in}}-\logit p_{\mathrm{out}}$, $\bar m:=\frac1K\logit p_{\mathrm{in}}+\frac{K-1}{K}\logit p_{\mathrm{out}}$, and $C:=\Wmax R^2$. For $u_1,\dots,u_K\in\Real^d$ with $\|u_a\|\le R$, symmetric $W$ with $\|W\|_{\op}\le\Wmax$, and $b\in\Real$, let $\bar s_{ab}:=b+u_a^\top Wu_b$.
\begin{enumerate}
\item[\emph{(a)}] \emph{(Necessary condition.)} Every such $\bar s$ satisfies
\[
\frac1K\sum_a\bar s_{aa}-\frac1{K^2}\sum_{a,b}\bar s_{ab}+\Bigl|\frac1{K^2}\sum_{a,b}\bar s_{ab}-b\Bigr|\;\le\;C.
\]
Hence $\bar s\!=\!M^\star$ with $|b|\le B_0$ requires
\begin{equation}\label{eq:exact-capacity}
C\;\ge\;\tfrac{K-1}{K}\,\gamma+\mathrm{dist}\bigl(\bar m,[-B_0,B_0]\bigr).
\end{equation}
\item[\emph{(b)}] \emph{(Community-level floor.)} For symmetric $T\in\Real^{K\times K}$, let $\Phi(T):=\frac1{K^2}\sum_{a,b}\mathrm{KL}\bigl(p_{ab}\,\|\,\sigma(T_{ab})\bigr)$, where $\mathrm{KL}$ is the Bernoulli divergence. Every symmetric $T$ with $\sum_aT_{aa}-\frac1K\sum_{a,b}T_{ab}\le KC$ satisfies $\Phi(T)\ge F_K(C)$, where
\[
F_K(C):=\min_{s_{\mathrm{in}}-s_{\mathrm{out}}\le KC/(K-1)}\Bigl[\tfrac1K\,\mathrm{KL}\bigl(p_{\mathrm{in}}\,\|\,\sigma(s_{\mathrm{in}})\bigr)+\tfrac{K-1}{K}\,\mathrm{KL}\bigl(p_{\mathrm{out}}\,\|\,\sigma(s_{\mathrm{out}})\bigr)\Bigr].
\]
By (a), this applies to every $\bar s$. Moreover, $F_K(C)\!>\!0$ if and only if $C\!<\!\frac{K-1}{K}\gamma$.
\item[\emph{(c)}] \emph{(Floor on $\eapp$.)} Let $n_{\min}:=\min_a|\mathcal C_a|\ge2$, $\kappa_n:=\frac{K^2n_{\min}(n_{\min}-1)}{n(n-1)}$, and $C_n:=C\bigl(1+\frac{2(K-1)}{K(n_{\min}-1)}\bigr)$. With probability at least $1-n^{-1}$ over the draw of $\Hcal$, every $(Z,W,\bhat)$ obeying Assumption~\ref{ass:bounded} satisfies
\[
\Rpop(S)-\Rpop(\Sopt)\;\ge\;\kappa_n\,F_K(C_n)-\Gamma^\circ_n,\qquad \Gamma^\circ_n:=\frac{C_2B}{\sqrt\tau}\sqrt{\frac{d\log(nBR\Wmax)}{n}},
\]
for a universal constant $C_2>0$. In particular, $\eapp$ obeys the same lower bound for every encoder family. For balanced communities, $\kappa_n\to1$ and $C_n\to C$.
\end{enumerate}
\end{proposition}

\begin{proof}
Let $U\!=\![u_1,\dots,u_K]\in\Real^{d\times K}$, $\bar u\!:=\!U\mathbf 1/K$, $\Pi\!:=\!I_K-\mathbf 1\mathbf 1^\top/K$, and $G\!:=\!U^\top WU\!=\!\bar s-b\,\mathbf 1\mathbf 1^\top$.

\emph{(a).} Since $\Pi\mathbf 1\!=\!0$, the left side equals $\frac1K\mathrm{tr}(\Pi G)+|\bar u^\top W\bar u|$. The matrix $U\Pi U^\top$ is positive semidefinite, so $\mathrm{tr}(\Pi G)=\mathrm{tr}(WU\Pi U^\top)\le\Wmax\,\mathrm{tr}(U\Pi U^\top)=\Wmax\bigl(\sum_a\|u_a\|^2-K\|\bar u\|^2\bigr)$. Also $|\bar u^\top W\bar u|\le\Wmax\|\bar u\|^2$. The sum is at most $\frac{\Wmax}{K}\sum_a\|u_a\|^2\le C$. For $\bar s\!=\!M^\star$, the two terms equal $\frac{K-1}{K}\gamma$ and $|\bar m-b|\ge\mathrm{dist}(\bar m,[-B_0,B_0])$, which proves~\eqref{eq:exact-capacity}.

\emph{(b).} The map $\Phi$ is convex. Both $\Phi$ and the constraint are invariant under $T\mapsto P T P^\top$ for every permutation matrix $P$, because $P\Pi P^\top\!=\!\Pi$ and $\sum_aT_{aa}-\frac1K\sum_{a,b}T_{ab}=\mathrm{tr}(\Pi T)$. Averaging over all $K!$ permutations therefore gives a feasible $\bar T=s_{\mathrm{out}}\mathbf 1\mathbf 1^\top+(s_{\mathrm{in}}-s_{\mathrm{out}})I_K$ with $\Phi(\bar T)\le\Phi(T)$. Its constraint reads $(K\!-\!1)(s_{\mathrm{in}}-s_{\mathrm{out}})\le KC$, and $\Phi(\bar T)$ is the bracket in $F_K$. The bracket vanishes only at $(\logit p_{\mathrm{in}},\logit p_{\mathrm{out}})$, which is feasible if and only if $\gamma\le KC/(K-1)$. For $T\!=\!\bar s$, the constraint holds because $\mathrm{tr}(\Pi\bar s)\!=\!\mathrm{tr}(\Pi G)\le KC$ by (a).

\emph{(c).} Per pair, $\ell(s;p)-\ell(\logit p;p)\!=\!\mathrm{KL}(p\,\|\,\sigma(s))$. So $\Rpop(S)-\Rpop(\Sopt)$ is the average over $\Hcal$ of $g(i,j):=\mathrm{KL}(P_{ij}\,\|\,\sigma(s_{ij}))$. Let $\bar g$ be the average of $g$ over all $N$ pairs $i<j$.

\emph{Step 1: from $\Hcal$ to all pairs.} The function $g$ takes values in $[0,2B]$ and is $1$-Lipschitz in $s_{ij}$, since $|\partial_s\ell(s;p)|\!=\!|\sigma(s)-p|\le1$. Repeating the proof of Lemma~\ref{lem:step2-restated} (App.~\ref{app:step2}) with this $g$ shows that, with probability at least $1-n^{-1}$, the average of $g$ over $\Hcal$ is at least $\bar g-\Gamma^\circ_n$ for all $(Z,W,\bhat)$ obeying Assumption~\ref{ass:bounded}.

\emph{Step 2: block averaging.} Let $n_a\!:=\!|\mathcal C_a|$, $u_a\!:=\!\frac1{n_a}\sum_{i\in\mathcal C_a}z_i$, and $\bar s_{ab}\!:=\!\bhat+u_a^\top Wu_b$. Since $W$ is symmetric, the average of $s_{ij}$ over the $n_an_b$ pairs between communities $a\!\ne\!b$ equals $T_{ab}\!:=\!\bar s_{ab}$. Over the $n_a(n_a-1)/2$ pairs inside community $a$, it equals $T_{aa}\!:=\!\bar s_{aa}+\delta_a$, where $\delta_a\!:=\!\frac1{n_a-1}\bigl(u_a^\top Wu_a-\frac1{n_a}\sum_{i\in\mathcal C_a}z_i^\top Wz_i\bigr)$, so $|\delta_a|\le2C/(n_{\min}-1)$. Jensen's inequality within each block, applied to the convex map $s\mapsto\mathrm{KL}(p\,\|\,\sigma(s))$, gives
\[
\bar g\;\ge\;\sum_a\frac{n_a(n_a-1)}{2N}\,\mathrm{KL}\bigl(p_{\mathrm{in}}\,\|\,\sigma(T_{aa})\bigr)+\sum_{a<b}\frac{n_an_b}{N}\,\mathrm{KL}\bigl(p_{\mathrm{out}}\,\|\,\sigma(T_{ab})\bigr)\;\ge\;\kappa_n\,\Phi(T).
\]
The last step uses $\frac{n_a(n_a-1)}{2N}\ge\frac{\kappa_n}{K^2}$ and $\frac{n_an_b}{N}\ge\frac{2\kappa_n}{K^2}$, with $N\!=\!n(n-1)/2$.

\emph{Step 3: the constraint.} By (a), $\mathrm{tr}(\Pi T)\!=\!\mathrm{tr}(\Pi\bar s)+\frac{K-1}{K}\sum_a\delta_a\le KC+\frac{2(K-1)C}{n_{\min}-1}\!=\!KC_n$. Hence $\Phi(T)\ge F_K(C_n)$ by (b). Combining the three steps proves the bound. It holds for every $(Z,W,\bhat)$ obeying Assumption~\ref{ass:bounded}, so it also holds for the infimum over $\Scal(A,\Tcal)$, which gives the bound on $\eapp$.
\end{proof}

For these SBMs, \eqref{eq:bilinear-capacity} reads $C\ge\gamma$ when $B_0\ge|b^\star|$ (Remark~\ref{rmk:op-vs-entrywise}), and \eqref{eq:exact-capacity} reads $C\ge\frac{K-1}{K}\gamma$ when $B_0\ge|\bar m|$. The capacity at which $M^\star$ becomes realizable therefore lies between the two, a range of the factor $K/(K-1)$. In our experiments ($K\!=\!3$, $\gamma\!=\!0.981$, $\bar m\!=\!-2.85$, $B_0\!=\!3.39$), it lies between cap-ratio $2/3$ and $1$. Below $2/3$, $F_3\!=\!0.0026$ at cap-ratio $0.25$ and $0.0004$ at $0.5$. The minimizer of $F_K$ then still has $s_{\mathrm{in}}-s_{\mathrm{out}}\!=\!KC/(K\!-\!1)\!>\!0$, so a positive floor does not by itself prevent recovery.

\section{Proof of Step 0 (Lemma~\ref*{lem:step0})}\label{app:step0}

\begin{lemma}[Step 0, restated]\label{lem:step0-restated}
Under Assumption~\ref{ass:bounded}, with probability at least $1-n^{-1}$ over $A$,
\[
\GammaH \;\le\; C_0\,B\sqrt{\frac{d\log(nBR\Wmax)}{\tau n}}
\]
for a universal $C_0>0$.
\end{lemma}

\begin{proof}
Define $\Zcal:=\{(Z,W,b):\|z_i\|\le R\ \forall i,\ W=W^\top,\ \|W\|_{\mathrm{op}}\le\Wmax,\ |b|\le B_0\}$, $s_{ij}(Z,W,b):=b+z_i^\top W z_j$, and
\[
U_\Hcal(Z,W,b):=\frac{1}{|\Hcal|}\sum_{(i,j)\in\Hcal}\bigl(\ell(s_{ij};A_{ij})-\ell(s_{ij};P_{ij})\bigr).
\]
The exact identity $\ell(s;a)-\ell(s;p)=(p-a)\,s$ gives
\begin{equation}\label{eq:UH-linear}
U_\Hcal(Z,W,b) \;=\; \frac{1}{|\Hcal|}\sum_{(i,j)\in\Hcal}(P_{ij}-A_{ij})\,s_{ij}(Z,W,b),
\end{equation}
so $\GammaH\le\sup_{\Zcal}|U_\Hcal|$. Throughout, $|\Hcal|=\lceil\tau N\rceil$ is deterministic.

\smallskip\noindent\textbf{(a) Hoeffding.} Conditional on the partition $(\Tcal,\Hcal)$, the held-out $\{A_{ij}:(i,j)\in\Hcal\}$ are independent of $A_\Tcal$ and of each other (since SBM edges are globally independent and $\Hcal\perp A$), with $\mathbb{E}[A_{ij}\mid\Hcal]=P_{ij}$. The summands $(P_{ij}-A_{ij})\,s_{ij}$ of~\eqref{eq:UH-linear} are therefore independent, mean-zero, bounded in $[-B,B]$ (since $|s_{ij}|\le B$, $|P_{ij}-A_{ij}|\le 1$). Hoeffding gives
\begin{equation}\label{eq:hoeffding-stepB}
\mathbb{P}\bigl[|U_\Hcal(Z,W,b)|>t\,\big|\,\Hcal\bigr] \;\le\; 2\exp\!\bigl(-|\Hcal|t^2/(2B^2)\bigr).
\end{equation}

\smallskip\noindent\textbf{(b) Covering.} Standard $\varepsilon$-nets of $\{\|z\|\le R\}\subset\Real^d$, of symmetric $W$ with $\|W\|_{\mathrm{op}}\le\Wmax$, and of $|b|\le B_0$ have sizes $(3R/\varepsilon)^d$, $(C_0'\Wmax/\varepsilon)^{d(d+1)/2}$, $3B_0/\varepsilon$ for a universal $C_0'$. Their product (over $n$ embedding copies, $W$, and $b$) gives an $\varepsilon$-net $\mathcal{N}_\varepsilon$ of $\Zcal$ with
\begin{equation}\label{eq:cover-stepB}
|\mathcal{N}_\varepsilon| \;\le\; (3R/\varepsilon)^{nd}\,(C_0'\Wmax/\varepsilon)^{d(d+1)/2}\,(3B_0/\varepsilon).
\end{equation}

\smallskip\noindent\textbf{(c) Lipschitz.} For $(Z,W,b), (Z',W',b')\in\Zcal$ agreeing to $\varepsilon$ in each component, the telescope
\[
z_i^\top W z_j-z_i'^\top W' z_j' \;=\; (z_i-z_i')^\top W z_j+z_i'^\top(W-W')z_j+z_i'^\top W'(z_j-z_j')
\]
with Cauchy--Schwarz gives $\sup_{i,j}|s_{ij}-s_{ij}'|\le(1+2R\Wmax+R^2)\varepsilon=:L\varepsilon$, hence $|U_\Hcal(Z,W,b)-U_\Hcal(Z',W',b')|\le L\varepsilon$ via $|P_{ij}-A_{ij}|\le 1$.

\smallskip\noindent\textbf{(d) Assembly.} Fix $\varepsilon := 1/(Ln^2)\le 1$. In the net sizes~\eqref{eq:cover-stepB} we read $R,\Wmax,B_0$ as $\max(\cdot,1)$, which keeps~\eqref{eq:cover-stepB} valid for $\varepsilon\le1$, and likewise $R,\Wmax$ inside $\log(nBR\Wmax)$, so that $\log(nBR\Wmax)\ge\log(2\log 2)>0$ for $n\ge2$. We may assume $d<2n$: otherwise the claimed bound exceeds $B$ for $C_0\ge2$, while $|U_\Hcal|\le B$ always. Then $d(d+1)/2\le nd$, and~\eqref{eq:cover-stepB} gives $\log|\mathcal{N}_\varepsilon| = O(nd\log(nBR\Wmax))$. Set
\begin{equation}\label{eq:t-stepB}
t \;:=\; B\sqrt{2\bigl(\log|\mathcal{N}_\varepsilon|+\log(2n)\bigr)/|\Hcal|}.
\end{equation}
A union bound of~\eqref{eq:hoeffding-stepB} over $\mathcal{N}_\varepsilon$ gives total failure probability $\le n^{-1}$. On this event, for any $(Z,W,b)\in\Zcal$ with nearest net point in $\mathcal{N}_\varepsilon$, $|U_\Hcal(Z,W,b)|\le t+L\varepsilon$ by (c). Since $\tau n^2/4\le|\Hcal|\le n^2$ and $B\ge\log 2$,
\[
L\varepsilon \;=\; \frac{1}{n^2} \;\le\; \frac{B\sqrt{2\log(2n)}}{n} \;\le\; t \;\le\; \frac{C_0}{2}\,B\sqrt{\frac{d\log(nBR\Wmax)}{\tau n}},
\]
so $|U_\Hcal|\le 2t\le C_0B\sqrt{d\log(nBR\Wmax)/(\tau n)}$ uniformly on $\Zcal$ for every $n\ge2$. The same bound holds for $\GammaH$, hence for the trained $(\widehat Z,\What,\bhat)\in\Zcal$.
\end{proof}

\section{Proof of Step 1 (Lemma~\ref*{lem:step1})}\label{app:step1}

We split the argument: a population strong-convexity statement for any score matrix in the bounded class (Lemma~\ref{lem:risk-sc}), then application to $\Shat$ via the excess-risk identity.

\begin{lemma}[Risk-level strong convexity]\label{lem:risk-sc}
For any $S=(s_{ij})$ with $|s_{ij}|\le B$ on $\Hcal$,
\begin{equation}\label{eq:risk-sc}
\Rpop(S)-\Rpop(\Sopt) \;\ge\; \frac{\muB}{2}\,\mathrm{ScoreMSE}_\Hcal(S).
\end{equation}
\end{lemma}

\begin{proof}
Fix $(i,j)\in\Hcal$. Since $\ell''(s;P_{ij})=\sigma(s)(1-\sigma(s))\ge\sigma(B)(1-\sigma(B))=\muB$ on $[-B,B]$, the per-pair BCE $\ell(\cdot;P_{ij})$ is $\muB$-strongly convex on $[-B,B]$ with minimizer $m^\star_{c_i c_j}=\logit(P_{ij})\in[-B,B]$. Taylor's theorem yields
\begin{equation}\label{eq:per-pair-sc}
\ell(s_{ij};P_{ij})-\ell(m^\star_{c_i c_j};P_{ij}) \;\ge\; \frac{\muB}{2}(s_{ij}-m^\star_{c_i c_j})^2.
\end{equation}
The population target $P_{ij}$ is essential: $\ell(\cdot;A_{ij})$ for $A_{ij}\in\{0,1\}$ is monotone in $s$ with no interior minimizer, so strong convexity on $[-B,B]$ would not yield a useful Taylor bound. Averaging~\eqref{eq:per-pair-sc} over $\Hcal$ gives~\eqref{eq:risk-sc}.
\end{proof}

\begin{lemma}[Step 1, restated]\label{lem:step1-restated}
Under Assumption~\ref{ass:bounded},
\[
\mathrm{ScoreMSE}_\Hcal \;\le\; \frac{2}{\muB}\bigl(\Rpop(\Shat)-\Rpop(\Sopt)\bigr) \;=\; \frac{2}{\muB}(\eopt+\eapp).
\]
\end{lemma}

\begin{proof}
$|\widehat s_{ij}|\le B$ by Assumption~\ref{ass:bounded} (no clipping), so Lemma~\ref{lem:risk-sc} applies to $\Shat$. The equality with $\eopt+\eapp$ is~\eqref{eq:excess-decomp}.
\end{proof}
Both sides of the bound are population objects, so no concentration is needed at this stage.

\section{Proof of Step 2 (Lemma~\ref*{lem:step2})}\label{app:step2}

\begin{lemma}[Step 2, restated]\label{lem:step2-restated}
Under Assumption~\ref{ass:bounded}, with probability at least $1-n^{-1}$ over the random draw of $\Hcal$, simultaneously for all $S\in\Scal(A,\Tcal)$,
\[
\mathrm{ScoreMSE}_{\mathrm{full}} \;\le\; \mathrm{ScoreMSE}_\Hcal+\Vstat,\quad \Vstat := \frac{C_1B^2}{\sqrt\tau}\sqrt{\frac{d\log(nBR\Wmax)}{n}}+\frac{4B^2}{n},
\]
for a universal $C_1>0$.
\end{lemma}

\begin{proof}
We first bound $\mathrm{ScoreMSE}_{\mathrm{all}}-\mathrm{ScoreMSE}_\Hcal$ where $\mathrm{ScoreMSE}_{\mathrm{all}}:=\frac{1}{N}\sum_{i<j}(s_{ij}-m^\star_{c_ic_j})^2$ averages over off-diagonal pairs, then add a diagonal correction. With $\Zcal$ as in App.~\ref{app:step0}, define $g_{Z,W,b}(i,j):=(b+z_i^\top W z_j-m^\star_{c_ic_j})^2$ and the held-out and full averages
\[
\bar g_\Hcal(Z,W,b):=\frac{1}{|\Hcal|}\sum_{(i,j)\in\Hcal}g_{Z,W,b}(i,j),\qquad \bar g(Z,W,b):=\frac{1}{N}\sum_{i<j}g_{Z,W,b}(i,j)=\mathrm{ScoreMSE}_{\mathrm{all}}.
\]
The range bound $g\in[0,D]$ with $D:=4B^2$ follows from $|b+z_i^\top W z_j|\le B_0+\Wmax R^2\le B$ on $\Zcal$ (Cauchy--Schwarz and Assumption~\ref{ass:bounded}) and $|m^\star_{c_ic_j}|\le B^\star\le B$. Since $\Hcal$ is a uniform random subset of $\{(i,j):i<j\}$ of size $|\Hcal|=\lceil\tau N\rceil$, we have $\mathbb{E}[\bar g_\Hcal]=\bar g$ for any fixed $(Z,W,b)$.

\smallskip\noindent\textbf{(a) Hoeffding--Serfling.} For sampling without replacement~\citep{serfling1974probability}, every fixed $(Z,W,b)\in\Zcal$ satisfies
\begin{equation}\label{eq:hoeffding-stepD}
\mathbb{P}\bigl[|\bar g_\Hcal(Z,W,b)-\bar g(Z,W,b)|>t\bigr] \;\le\; 2\exp\!\Bigl(-\frac{2|\Hcal|\,t^2}{D^2(1-f^*)}\Bigr),\qquad f^*:=\frac{|\Hcal|-1}{N}.
\end{equation}
Equivalently, the moment generating function of $|\Hcal|(\bar g_\Hcal-\bar g)$ is dominated by that of the i.i.d.\ analogue (Hoeffding's reduction theorem,~\citealp{hoeffding1963probability}).

\smallskip\noindent\textbf{(b) Covering.} The same $\mathcal{N}_\varepsilon$ as in App.~\ref{app:step0} (Eq.~\eqref{eq:cover-stepB}) covers $\Zcal$.

\smallskip\noindent\textbf{(c) Lipschitz.} The score Lipschitz argument of App.~\ref{app:step0}~(c) gives $|s_{ij}-s_{ij}'|\le(1+2R\Wmax+R^2)\varepsilon$ for $(Z,W,b),(Z',W',b')$ within $\varepsilon$ in each component. Combined with $g-g'=(s-s')(s+s'-2m^\star)$ and $|s+s'-2m^\star|\le 4B$,
\begin{equation}\label{eq:lipsch-stepD}
\sup_{i<j}|g_{Z,W,b}(i,j)-g_{Z',W',b'}(i,j)| \;\le\; 4B(1+2R\Wmax+R^2)\varepsilon =: L\varepsilon,
\end{equation}
hence $\max\{|\bar g_\Hcal-\bar g_\Hcal'|,|\bar g-\bar g'|\}\le L\varepsilon$.

\smallskip\noindent\textbf{(d) Assembly.} Use the net $\mathcal N_\varepsilon$ of App.~\ref{app:step0}~(d), with $\varepsilon=1/((1+2R\Wmax+R^2)n^2)$ and the same conventions, so that $\log|\mathcal N_\varepsilon| = O(nd\log(nBR\Wmax))$ and $L\varepsilon=4B/n^2$. As there, the case $d\ge2n$ is trivial, since $\mathrm{ScoreMSE}_{\mathrm{full}}\le4B^2$ always. Set
\[
t \;:=\; D\sqrt{\frac{(1-f^*)\bigl(\log|\mathcal{N}_\varepsilon|+\log(2n)\bigr)}{2|\Hcal|}}.
\]
A union bound of~\eqref{eq:hoeffding-stepD} over $\mathcal{N}_\varepsilon$ gives total failure probability $\le n^{-1}$. On this event, for any $(Z,W,b)\in\Zcal$ with nearest net point in $\mathcal{N}_\varepsilon$, $|\bar g_\Hcal(Z,W,b)-\bar g(Z,W,b)|\le t+2L\varepsilon$ by triangle inequality and (c). With $|\Hcal|\ge\tau n^2/4$, $D=4B^2$, and $1-f^*\le 1$,
\begin{equation}\label{eq:2t-stepD}
t \;\le\; \frac{C_0''B^2}{\sqrt\tau}\sqrt{\frac{d\log(nBR\Wmax)}{n}},\qquad 2L\varepsilon \;=\; \frac{8B}{n^2} \;\le\; \frac{8B^2}{\sqrt\tau}\sqrt{\frac{d\log(nBR\Wmax)}{n}},
\end{equation}
where the second bound uses $B\ge\log 2$ and $\log(nBR\Wmax)\ge\log(2\log2)$ for $n\ge2$. Hence $|\bar g_\Hcal-\bar g|\le t+2L\varepsilon$ is at most $(C_0''+8)B^2\sqrt{d\log(nBR\Wmax)/(\tau n)}$ uniformly on $\Zcal$, for every $n\ge2$. The $1/\sqrt\tau$ rate (rather than $1/\tau$) comes from $|\Hcal|\sim\tau N$ in the denominator together with $1-f^*\le 1$.

\smallskip\noindent\textbf{Diagonal correction.} Using $|s_{ii}-m^\star_{c_ic_i}|\le 2B$ and $|\{(i,i):i\in[n]\}|/n^2=1/n$,
\[
\mathrm{ScoreMSE}_{\mathrm{full}} \;\le\; \mathrm{ScoreMSE}_{\mathrm{all}}+\frac{4B^2}{n},
\]
giving the claim with $C_1:=C_0''+8$. Since the bound on $|\bar g_\Hcal-\bar g|$ holds simultaneously on $\Zcal$, it covers every $S\in\Scal(A,\Tcal)$, in particular the trained $(\widehat Z,\What,\bhat)$.
\end{proof}

\section{Proof of Step 3 (Proposition~\ref*{prop:step3}) and Plug-in Variant}\label{app:step3}

\subsection{Score Profiles, Prototypes, and Node-wise Error}

For each $i\in[n]$, the \emph{full score profile} is $\tilde s_i:=(s_{ij})_{j\in[n]}\in\Real^n$, where the off-diagonal entries $\{s_{ij}\}_{j\ne i}$ are the LP scores from~\eqref{eq:score-def} and the diagonal entry $s_{ii}=\bhat+z_i^\top W z_i$ is the formal extension introduced in Sec.~\ref{sec:setting}. This convention does not affect the LP objective or held-out evaluation; it ensures $\tilde s_i\in\Real^n$ matches $Z\in\Real^{n\times d}$ in dimension, giving the clean bilinear identity (with $W=W^\top$, so $z_i^\top W z_j=z_j^\top W z_i$)
\begin{equation}\label{eq:profile-linear}
\tilde s_i \;=\; \bhat\,\mathbf 1_n + ZW z_i,
\end{equation}
which underlies the equivalence between nearest-prototype classification and linear classification on $z_i$ (Lemma~\ref{lem:linear}). The \emph{oracle prototype} for community $a$ is $\mu_a:=(m^\star_{a,c_j})_{j\in[n]}\in\Real^n$, and the \emph{node-wise squared error} is $\tilde e_i^2:=n^{-1}\|\tilde s_i-\mu_{c_i}\|^2$.

\begin{lemma}[Average node-wise error equals full ScoreMSE]\label{lem:node-error-mse}
\begin{equation}\label{eq:node-error-mse}
\frac{1}{n}\sum_{i=1}^n\tilde e_i^2 \;=\; \mathrm{ScoreMSE}_{\mathrm{full}}.
\end{equation}
\end{lemma}

\begin{proof}
\[
\frac{1}{n}\sum_{i=1}^n\tilde e_i^2 \;=\; \frac{1}{n}\sum_{i=1}^n\frac{1}{n}\sum_{j=1}^n(s_{ij}-m^\star_{c_ic_j})^2 \;=\; \frac{1}{n^2}\sum_{i,j}(s_{ij}-m^\star_{c_ic_j})^2 \;=\; \mathrm{ScoreMSE}_{\mathrm{full}}. \qedhere
\]
\end{proof}

\subsection{Nearest-Prototype is a Linear Classifier}

Define the \emph{oracle classifier} as nearest-prototype on the score profile,
\[
h^\star(z_i) \;:=\; \arg\min_{a\in[K]}\|\tilde s_i - \mu_a\|,
\]
and the \emph{plug-in classifier} $\hat h^{\mathrm{plug}}(z_i) := \arg\min_{a}\|\tilde s_i - \hat\mu_a\|$ with sample-mean prototypes $\hat\mu_a := m_a^{-1}\sum_{i\in L_a}\tilde s_i$ from a labeled set $L = \sqcup_a L_a$.

\begin{lemma}[Nearest-prototype is linear in $z_i$]\label{lem:linear}
For any prototypes $\nu_1,\ldots,\nu_K\in\Real^n$,
\[
\arg\min_{a\in[K]}\|\tilde s_i - \nu_a\| \;=\; \arg\max_{a\in[K]}\bigl(w_a^\top z_i + b_a\bigr),\qquad w_a := WZ^\top\nu_a\in\Real^d,\ \ b_a := \bhat\,\nu_a^\top\mathbf 1_n - \tfrac12\|\nu_a\|^2.
\]
In particular, $h^\star$ and $\hat h^{\mathrm{plug}}$ are linear classifiers on $z_i$ with $\nu_a=\mu_a$ and $\nu_a=\hat\mu_a$ respectively.
\end{lemma}

\begin{proof}
Expand $\|\tilde s_i-\nu_a\|^2 = \|\tilde s_i\|^2 - 2\nu_a^\top\tilde s_i + \|\nu_a\|^2$; the first term is $a$-independent, so
\[
\arg\min_{a}\|\tilde s_i-\nu_a\|^2 \;=\; \arg\max_{a}\bigl(\nu_a^\top\tilde s_i - \tfrac12\|\nu_a\|^2\bigr).
\]
Substituting~\eqref{eq:profile-linear} and using $W=W^\top$, $\nu_a^\top\tilde s_i = \bhat\,\nu_a^\top\mathbf 1_n + (WZ^\top\nu_a)^\top z_i$, hence $\nu_a^\top\tilde s_i - \tfrac12\|\nu_a\|^2 = w_a^\top z_i + b_a$.
\end{proof}

\subsection{Oracle Classification Error}

\begin{lemma}\label{lem:oracle-error}
$\frac{1}{n}\sum_i\indic\{h^\star(z_i)\ne c_i\}\le 4\,\mathrm{ScoreMSE}_{\mathrm{full}}/\Deltas^2$.
\end{lemma}

\begin{proof}
$\|\mu_a-\mu_b\|^2=n\sum_c\pi_c(m^\star_{ac}-m^\star_{bc})^2$, so $\min_{a\ne b}\|\mu_a-\mu_b\|=\sqrt n\,\Deltas$. Node $i$ with $c_i=a$ is correctly classified when $\tilde e_i<\Deltas/2$. Markov on $\tilde e_i^2$ at threshold $\Deltas^2/4$ and~\eqref{eq:node-error-mse} give
\[
\frac{1}{n}\sum_i\indic\{h^\star(z_i)\ne c_i\}\le\frac{1}{n}\sum_i\indic\{\tilde e_i\ge\Deltas/2\}\le\frac{4}{\Deltas^2}\,\mathrm{ScoreMSE}_{\mathrm{full}}. \qedhere
\]
\end{proof}

This proves Proposition~\ref{prop:step3}.

\subsection{Plug-in Prototype Estimation via Direct \texorpdfstring{$\eta_L$}{eta\_L} Bernstein}

For each $a\in[K]$, sample $L_a\subset\mathcal{C}_a$ uniformly without replacement, $|L_a|=m_a\ge 1$, $m_{\min}:=\min_a m_a$. Set $\eta_L:=\max_a n^{-1/2}\|\hat\mu_a-\mu_a\|$.

\begin{lemma}[Direct $\eta_L$ Bernstein]\label{lem:prototype}
With probability $\ge 1-\delta$ over the draw of $L$ (independent of the trained model),
\begin{equation}\label{eq:eta-bound}
\eta_L \;\le\; \sqrt{\frac{\mathrm{ScoreMSE}_{\mathrm{full}}}{\pi_{\min}}}+\sqrt{\frac{2\,\mathrm{ScoreMSE}_{\mathrm{full}}\,\lambda_\delta}{m_{\min}\pi_{\min}}}+\frac{4B\,\lambda_\delta}{3m_{\min}},\quad \lambda_\delta:=\log(2K/\delta).
\end{equation}
\end{lemma}

\begin{proof}
Triangle inequality: $n^{-1/2}\|\hat\mu_a-\mu_a\|\le m_a^{-1}\sum_{i\in L_a}\tilde e_i$. Set $y_i:=\tilde e_i\in[0,2B]$ and $\bar y_a:=n_a^{-1}\sum_{\mathcal{C}_a}y_i$. Cauchy--Schwarz gives $\bar y_a\le\sqrt{E_a}$ with $E_a:=n_a^{-1}\sum_{\mathcal{C}_a}\tilde e_i^2\le\mathrm{ScoreMSE}_{\mathrm{full}}/\pi_{\min}$, and the variance $v_a\le E_a$. Sampling without replacement is dominated by i.i.d.\ sampling in the moment generating function (Hoeffding's reduction), so Bernstein's inequality gives
\[
\mathbb{P}\Bigl[m_a^{-1}\textstyle\sum_{L_a}y_i>\bar y_a+\sqrt{2v_a\lambda/m_a}+\tfrac{4B\lambda}{3m_a}\Bigr]\le e^{-\lambda}.
\]
Setting $\lambda=\lambda_\delta$ and union-bounding over $a\in[K]$, then combining with the triangle bound and using $m_a\ge m_{\min}$, gives~\eqref{eq:eta-bound}.
\end{proof}

\begin{remark}[Linear vs.\ quadratic label dependence]\label{rmk:label-dep}
Bounding $\eta_L$ directly (vs.\ $\eta_L^2$) gives $m_{\min}\gtrsim B\log(K/\delta)/\Deltas$ (linear in $1/\Deltas$) once $\mathrm{ScoreMSE}_{\mathrm{full}}\le\pi_{\min}\Deltas^2/144$ (Cor.~\ref{cor:simplified}).
\end{remark}

\subsection{Plug-in Classification Error}

\begin{lemma}\label{lem:plug-in-error}
If $\eta_L<\Deltas/2$,
\[
\frac{1}{n}\sum_i\indic\{\hat h^{\mathrm{plug}}(z_i)\ne c_i\} \;\le\; \frac{4\,\mathrm{ScoreMSE}_{\mathrm{full}}}{(\Deltas-2\eta_L)^2}.
\]
\end{lemma}

\begin{proof}
Triangle: $\|\hat\mu_a-\hat\mu_b\|\ge\sqrt n(\Deltas-2\eta_L)$. For $c_i=a$, $\|\tilde s_i-\hat\mu_a\|\le\sqrt n(\tilde e_i+\eta_L)$ and $\|\tilde s_i-\hat\mu_b\|\ge\sqrt n(\Deltas-\tilde e_i-\eta_L)$; correct classification requires $\tilde e_i<(\Deltas-2\eta_L)/2$. Markov on $\tilde e_i^2$ with~\eqref{eq:node-error-mse} gives the bound.
\end{proof}

\section{Main Theorem, Mahalanobis Form, and Unsupervised Recovery}\label{app:main-results}

\subsection{Proof of Theorem~\ref*{thm:main}}\label{app:main-proof}

Let $\Omega_n$ be the intersection of the high-probability events of Lemma~\ref{lem:step0-restated} (Step 0) and Lemma~\ref{lem:step2-restated} (Step 2); Step 1 (Lemma~\ref{lem:step1-restated}) is deterministic on the bounded class. Union bound: $\mathbb P(\Omega_n)\ge 1-2n^{-1}$. On $\Omega_n$, chaining Steps 1--2 with the excess-risk identity~\eqref{eq:excess-decomp} gives $\mathrm{ScoreMSE}_{\mathrm{full}}\le \En$.

\smallskip\noindent\emph{Part I.} Lemma~\ref{lem:oracle-error} applied with $\mathrm{ScoreMSE}_{\mathrm{full}}\le\En$ gives the oracle error bound $4\En/\Deltas^2$.

\smallskip\noindent\emph{Part II.} Conditional on $\Omega_n$, the label set $L$ is independent of the trained $(\Shat,\hat Z,\What,\bhat)$, so Lemma~\ref{lem:prototype} (with $\delta=n^{-1}$, hence $\lambda_\delta=\log(2Kn)=\lambda_n$) gives the $\eta_L$ Bernstein bound with conditional probability $\ge 1-n^{-1}$ over $L$. On $\eta_L<\Deltas/2$, Lemma~\ref{lem:plug-in-error} with $\mathrm{ScoreMSE}_{\mathrm{full}}\le\En$ gives the plug-in error $4\En/(\Deltas-2\eta_L)^2$. The explicit $(\hat w_a,\hat b_a)$ form follows from Lemma~\ref{lem:linear} with $\nu_a=\hat\mu_a$. Joint failure probability over $(A,\Hcal,L)$ is $\le 2n^{-1}+n^{-1}=3n^{-1}$, matching the theorem.

\subsection{Proof of Proposition~\ref*{prop:mahalanobis}}\label{app:mahalanobis}

By~\eqref{eq:profile-linear}, $\hat\mu_a = |L_a|^{-1}\sum_{i\in L_a}\tilde s_i = \bhat\mathbf 1_n+ZW\bar z_a$. For any test embedding $z\in\Real^d$, its score profile $\tilde s(z):=\bhat\mathbf 1_n+ZWz$ satisfies
\[
\tilde s(z)-\hat\mu_a \;=\; ZW(z-\bar z_a),
\]
and
\[ \|\tilde s(z)-\hat\mu_a\|^2 \;=\; (z-\bar z_a)^\top WZ^\top ZW(z-\bar z_a) \;=\; (z-\bar z_a)^\top G(z-\bar z_a),
\]
using $W=W^\top$. The first equality of Proposition~\ref{prop:mahalanobis} follows from $\hat h^{\mathrm{plug}}(z)=\arg\min_a\|\tilde s(z)-\hat\mu_a\|^2$ (Lemma~\ref{lem:linear} read as a nearest-prototype rule); the second is the standard $\arg\min$-to-$\arg\max$ expansion (drop $a$-independent $z^\top Gz$). PSD: $v^\top Gv = \|ZWv\|^2\ge 0$ for all $v\in\Real^d$. \qed

\subsection{Unsupervised Recovery by $K$-means}\label{app:kmeans}

For an assignment $\hat c:[n]\to[K]$ and centers $c_1,\ldots,c_K\in\Real^d$, define the $K$-means objective under $G$ as
\[
\mathcal J_G(\hat c;c_1,\ldots,c_K) \;:=\; \sum_{i=1}^n (z_i-c_{\hat c(i)})^\top G\,(z_i-c_{\hat c(i)}).
\]
For $\kappa\ge0$, an assignment $\hat c$ is a \emph{$(1+\kappa)$-approximate $K$-means clustering of the embeddings under $G$} if $\mathcal J_G(\hat c;c_1,\ldots,c_K)\le(1+\kappa)\min\mathcal J_G$ for some centers, where the minimum is over all assignments and centers. Exact $K$-means corresponds to $\kappa=0$.

\begin{corollary}[Unsupervised recovery by $K$-means]\label{cor:kmeans}
Under the setting of Theorem~\ref{thm:main}, work on the event $\Omega_n$ of App.~\ref{app:main-proof}, and let $\hat c$ be any $(1+\kappa)$-approximate $K$-means clustering of $\{z_i\}_{i=1}^n$ under $G$. If $(16+8\kappa)\En/\Deltas^2<\pi_{\min}$, then there exists a permutation $\sigma$ of $[K]$ such that
\[
\frac1n\sum_{i=1}^n\indic\{\hat c(i)\ne\sigma(c_i)\}\;\le\;\frac{(16+8\kappa)\,\En}{\Deltas^2}.
\]
\end{corollary}

\begin{proof}
\emph{Step 1: $K$-means under $G$ is Euclidean $K$-means on score profiles.} For $c\in\Real^d$ set $\nu(c):=\bhat\mathbf 1_n+ZWc\in\Real^n$. By~\eqref{eq:profile-linear} and $W=W^\top$, $\tilde s_i-\nu(c)=ZW(z_i-c)$, so
\[
(z_i-c)^\top G\,(z_i-c)\;=\;\|\tilde s_i-\nu(c)\|^2.
\]
Hence $\mathcal J_G(\hat c;c_1,\ldots,c_K)$ equals the Euclidean $K$-means objective $\sum_i\|\tilde s_i-\nu_{\hat c(i)}\|^2$ of the score profiles with centers $\nu_k:=\nu(c_k)$. Conversely, for a fixed assignment the optimal Euclidean centers on $\{\tilde s_i\}$ are the cluster means, and the mean of the $\tilde s_i$ over a cluster equals $\nu(\cdot)$ evaluated at the mean of the corresponding $z_i$. The two problems therefore have the same optimal value, and a $(1+\kappa)$-approximate solution of one is a $(1+\kappa)$-approximate solution of the other with the same assignment $\hat c$.

\emph{Step 2: the $K$-means error bound of \citet{lei2015consistency}.} Let $\widehat U\in\Real^{n\times n}$ have rows $\tilde s_i^\top$, and let $U:=\Theta X$, where $\Theta\in\{0,1\}^{n\times K}$ is the membership matrix of the labels $c$ and $X\in\Real^{K\times n}$ has rows $\mu_1^\top,\ldots,\mu_K^\top$. By Lemma~\ref{lem:node-error-mse}, on $\Omega_n$,
\[
\|\widehat U-U\|_F^2\;=\;\sum_{i=1}^n\|\tilde s_i-\mu_{c_i}\|^2\;=\;n^2\,\mathrm{ScoreMSE}_{\mathrm{full}}\;\le\;n^2\En.
\]
The proof of Lemma~\ref{lem:oracle-error} gives $\min_{a\ne b}\|\mu_a-\mu_b\|=\sqrt n\,\Deltas=:\delta$, and each community has $n_k\ge\pi_{\min}n$ nodes. Lemma~5.3 of \citet{lei2015consistency} states the following for any $(1+\kappa)$-approximate $K$-means solution $\bar U$ of $\widehat U$. The sets $S_k:=\{i:c_i=k,\ \|\bar U_{i}-U_{i}\|\ge\delta/2\}$ satisfy
\[
\sum_{k=1}^K|S_k|\,\delta^2\;\le\;(16+8\kappa)\,\|\widehat U-U\|_F^2,
\]
and if $(16+8\kappa)\|\widehat U-U\|_F^2/\delta^2<n_k$ for all $k$, then all nodes outside $\bigcup_kS_k$ are clustered correctly up to a permutation of the cluster labels. The lemma is stated with $K$ columns, but its proof uses only the triangle inequality, the approximate optimality of $\bar U$ against the feasible competitor $U$, and a pigeonhole argument over the $K$ communities, so it holds verbatim with $n$ columns. Here $(16+8\kappa)\|\widehat U-U\|_F^2/\delta^2\le(16+8\kappa)\,n\En/\Deltas^2<\pi_{\min}n\le n_k$ by assumption. The number of misclassified nodes is therefore at most
\[
\Bigl|\bigcup_kS_k\Bigr|\;\le\;\frac{(16+8\kappa)\,\|\widehat U-U\|_F^2}{\delta^2}\;\le\;\frac{(16+8\kappa)\,n\,\En}{\Deltas^2},
\]
and dividing by $n$ gives the claim.
\end{proof}

\begin{remark}[Solvers, implementation, and relation to the plug-in classifier]\label{rmk:kmeans}
Exact $K$-means is NP-hard, so Corollary~\ref{cor:kmeans} is stated for any solver with a multiplicative approximation guarantee. Lloyd's iterations carry no such guarantee in general. Since $(z_i-c)^\top G(z_i-c)=\|G^{1/2}(z_i-c)\|^2$, $K$-means under $G$ can be run as Euclidean $K$-means on $G^{1/2}z_i\in\Real^d$. The plug-in classifier of Proposition~\ref{prop:mahalanobis} and the $K$-means clustering use the same quadratic form $G$ and differ only in how the centroids are obtained: from the labeled sets $L_a$, or without labels. The price of removing labels is a larger constant, $16+8\kappa$ instead of $4$ for the oracle classifier, and the condition $(16+8\kappa)\En/\Deltas^2<\pi_{\min}$, which ensures that every community keeps a correctly clustered node.
\end{remark}

\subsection{Simplified Plug-in Bound in a Clean Regime}\label{app:simplified}
\begin{corollary}[Simplified bound]\label{cor:simplified}
Under the setting of Theorem~\ref{thm:main}, suppose
\begin{equation}\label{eq:regime}
 \En \le \frac{\pi_{\min}\Deltas^2}{144} \qquad \text{and} \qquad m_{\min} \ge \frac{16\,B\,\log(2K/\delta)}{\Deltas}.
\end{equation}
Then, on the event $\Omega_n$ of App.~\ref{app:main-proof}, $\eta_L\le\Deltas/4$ with probability $\ge 1-\delta$ over $L$, and
\begin{equation}\label{eq:simplified-bound}
\frac{1}{n}\sum_{i=1}^n\indic\bigl\{\hat h^{\mathrm{plug}}(z_i)\ne c_i\bigr\} \;\le\; \frac{16\,\En}{\Deltas^2}.
\end{equation}
\end{corollary}

\begin{proof}
We show each of the three terms of the $\eta_L$ Bernstein bound is at most $\Deltas/12$, so $\eta_L\le\Deltas/4$.

\smallskip\noindent\emph{Term 1.} The first condition of~\eqref{eq:regime} gives $\sqrt{\En/\pi_{\min}}\le\Deltas/12$.

\smallskip\noindent\emph{Term 3.} The second condition gives $4B\lambda_\delta/(3m_{\min}) \le \Deltas/12$.

\smallskip\noindent\emph{Term 2.} Factor $\sqrt{2\En\lambda_\delta/(m_{\min}\pi_{\min})} = \sqrt{2\lambda_\delta/m_{\min}}\cdot\sqrt{\En/\pi_{\min}} \le \sqrt{2\lambda_\delta/m_{\min}}\cdot\Deltas/12$. Under the second condition, $2\lambda_\delta/m_{\min}\le\Deltas/(4B)$. Since $|m^\star_{ac}-m^\star_{bc}|\le 2B$ for all $a,b,c$, we have $\Deltas^2\le\sum_c\pi_c(2B)^2=4B^2$, so $\Deltas/(4B)\le 1/2$ and Term~2 $\le\Deltas/12$.

Combining, $\eta_L\le\Deltas/4$, so $\Deltas-2\eta_L\ge\Deltas/2$, and Theorem~\ref{thm:main} Part~II gives $4\En/(\Deltas/2)^2 = 16\En/\Deltas^2$.
\end{proof}

\subsection{Proof of Corollary~\ref*{cor:half-empirical}}\label{app:half-empirical}

Since $\GammaH$ is a supremum over $\Scal(A,\Tcal)\ni\Shat$, its definition gives $\Rpop(\Shat)\le\widehat R_\Hcal(\Shat)+\GammaH$, and Lemma~\ref{lem:step0-restated} bounds $\GammaH$. Subtracting $\Rpop(\Sopt)$ gives $\Rpop(\Shat)-\Rpop(\Sopt)\le\widehat R_\Hcal(\Shat)-\Rpop(\Sopt)+\GammaH$. Substituting into the Part~I chain (App.~\ref{app:main-proof}: Step~1 $\to$ Step~2 $\to$ Lemma~\ref{lem:oracle-error}) gives the stated bound. Replacing Lemma~\ref{lem:oracle-error} by Lemma~\ref{lem:plug-in-error} in the same chain yields the analogous half-empirical bound for Part~II (plug-in), with $\Deltas^2$ replaced by $(\Deltas-2\eta_L)^2$. \qed

\section{Proof of Corollary~\ref*{cor:train-side}}\label{app:train-side}

Throughout this appendix, define the deterministic relaxed score class
\[
\Scal^{\Zcal} := \bigl\{(s_{ij})_{i,j\in[n]} : s_{ij}=\bhat+z_i^\top W z_j,\ \|z_i\|\le R,\ W=W^\top,\ \|W\|_{\mathrm{op}}\le\Wmax,\ |\bhat|\le B_0\bigr\}.
\]
Boundedness in Assumption~\ref{ass:bounded} gives $\Scal(A,\Tcal)\subseteq\Scal^{\Zcal}$, since any encoder output $z_i=f_\theta(\Aenc,i)$ with $\|z_i\|\le R$ realizes an admissible $(Z,W,\bhat)$ in $\Scal^{\Zcal}$. The class $\Scal^{\Zcal}$ is deterministic (not a function of $\Aenc$); it appears only in the proofs below as a uniform-concentration envelope, and not in the statement of Cor.~\ref{cor:train-side}.

\subsection{Uniform Deviation Rates}\label{app:train-side-rates}

\begin{lemma}[$\GammaT$ rate]\label{lem:gammaT}
Under Assumption~\ref{ass:bounded}, with probability $\ge 1-n^{-1}$,
\[
\GammaT \;\le\; \sup_{S\in\Scal^{\Zcal}}\bigl|\Rempt(S)-\Rpopt(S)\bigr| \;\le\; C_2\,B\sqrt{\frac{d\log(nBR\Wmax)}{(1-\tau)n}}
\]
for a universal $C_2>0$.
\end{lemma}

\begin{proof}
First inequality is by inclusion $\Scal(A,\Tcal)\subseteq\Scal^{\Zcal}$. For the second, we mirror Lemma~\ref{lem:step0-restated} with $\Hcal\leftrightarrow\Tcal$. The exact identity $\ell(s;a)-\ell(s;p)=(p-a)s$ gives
\[
\Rempt(S)-\Rpopt(S) \;=\; \frac{1}{|\Tcal|}\sum_{(i,j)\in\Tcal}(P_{ij}-A_{ij})\,s_{ij}.
\]
Conditional on $(\Tcal,\Hcal)$, the summands are independent (SBM edges are globally independent), mean zero, bounded in $[-B,B]$ on $\Scal^{\Zcal}$. We run App.~\ref{app:step0}~(a)--(d) with $\Tcal$ in place of $\Hcal$: the same net $\mathcal N_\varepsilon$ with $\varepsilon=1/((1+2R\Wmax+R^2)n^2)$, the same conventions, and the same reduction to $d<2n$. With probability $\ge1-n^{-1}$ this gives $\sup_{\Scal^{\Zcal}}|\Rempt-\Rpopt|\le2t$, where $t$ is~\eqref{eq:t-stepB} with $|\Tcal|$ in place of $|\Hcal|$. The step $L\varepsilon=1/n^2\le t$ uses only $|\Tcal|\le n^2$. The only other change is the size of the pair set. Since $\Tcal$ is nonempty (otherwise $\Rempt$ is undefined) and $|\Tcal|=N-\lceil\tau N\rceil>(1-\tau)N-1$, we have $|\Tcal|\ge(1-\tau)N/2$ (directly if $(1-\tau)N\ge2$, and from $|\Tcal|\ge1$ otherwise). With $N\ge n^2/4$ this gives $|\Tcal|\ge(1-\tau)n^2/8$, and the stated rate follows for every $n\ge2$.
\end{proof}

\begin{lemma}[$\GammaTH$ rate]\label{lem:gammaTH}
Under Assumption~\ref{ass:bounded}, with probability $\ge 1-n^{-1}$,
\[
\GammaTH \;\le\; \sup_{S\in\Scal^{\Zcal}}\bigl|\Rpopt(S)-\Rpop(S)\bigr| \;\le\; \frac{C_3B^2}{\sqrt{\tau(1-\tau)}}\sqrt{\frac{d\log(nBR\Wmax)}{n}}
\]
for a universal $C_3>0$.
\end{lemma}

\begin{proof}
First inequality by inclusion. For the second, fix $S\in\Scal^{\Zcal}$ and write
\[
\Rpopt(S)-\Rpop(S) = \Bigl[\frac{1}{|\Tcal|}\sum_{\Tcal}g_S - \bar g_S\Bigr] - \Bigl[\frac{1}{|\Hcal|}\sum_{\Hcal}g_S - \bar g_S\Bigr],
\]
with $g_S(i,j):=\ell(s_{ij};P_{ij})\in[0,2B]$ (since $\ell(s;p)\le |s|+\log 2\le 2B$ for $|s|\le B$ and $B\ge\log 2$ by the convention of Sec.~\ref{sec:setting}) and full-pair mean $\bar g_S:=\frac{1}{N}\sum_{i<j}g_S(i,j)$. Each bracket is the deviation of a sampling-without-replacement average from its full-pair mean, since $\Hcal$ is a uniform subset of the $N$ pairs and so is its complement $\Tcal$. Hence Hoeffding--Serfling~\eqref{eq:hoeffding-stepD} applies to each bracket with $D$ replaced by $2B$, and we use $1-f^*\le1$.

We use the net $\mathcal N_\varepsilon$ and the conventions of App.~\ref{app:step2}~(d), so $\varepsilon=1/((1+2R\Wmax+R^2)n^2)$ and $\log|\mathcal N_\varepsilon|=O(nd\log(nBR\Wmax))$. The case $d\ge2n$ is trivial, since $|\Rpopt(S)-\Rpop(S)|\le2B$ always. BCE is $1$-Lipschitz in $s$, because $|\partial_s\ell(s;p)|=|\sigma(s)-p|\le1$. With the score Lipschitz bound of App.~\ref{app:step0}~(c), this gives $|g_S(i,j)-g_{S'}(i,j)|\le1/n^2$ for the nearest net point $S'$, so each bracket changes by at most $2/n^2$. For $\mathcal E\in\{\Tcal,\Hcal\}$ set
\[
t_{\mathcal E} \;:=\; 2B\sqrt{\frac{\log|\mathcal N_\varepsilon|+\log(4n)}{2|\mathcal E|}}.
\]
A union bound over $\mathcal N_\varepsilon$ and the two brackets gives total failure probability $\le n^{-1}$. On this event, uniformly on $\Scal^{\Zcal}$,
\[
\bigl|\Rpopt(S)-\Rpop(S)\bigr| \;\le\; t_\Tcal+t_\Hcal+\frac{4}{n^2}.
\]
Using $|\Hcal|\ge\tau N$, $|\Tcal|\ge(1-\tau)N/2$ (proof of Lemma~\ref{lem:gammaT}) and $|\Tcal|+|\Hcal|=N$,
\[
\frac{1}{\sqrt{|\Tcal|}}+\frac{1}{\sqrt{|\Hcal|}} \;\le\; \sqrt{\frac{2N}{|\Tcal|\,|\Hcal|}} \;\le\; \frac{2}{\sqrt{\tau(1-\tau)N}},
\]
so with $N\ge n^2/4$ we get $t_\Tcal+t_\Hcal\le C'B\sqrt{d\log(nBR\Wmax)/(\tau(1-\tau)n)}$ for a universal $C'$. The remainder satisfies $4/n^2\le4B^2\sqrt{d\log(nBR\Wmax)/(\tau(1-\tau)n)}$ for $n\ge2$, by $B\ge\log2$, $\tau(1-\tau)\le1/4$ and $\log(nBR\Wmax)\ge\log(2\log2)$. Since $B\le B^2/\log2$, the stated bound holds with $C_3:=C'/\log2+4$, for every $n\ge2$.
\end{proof}

\subsection{Proof of Cor.~\ref*{cor:train-side}}

Union-bound Lemmas~\ref{lem:gammaT},~\ref{lem:gammaTH}: with probability $\ge 1-2n^{-1}$, both deviation bounds hold simultaneously, and the first inequalities transfer the rates to $\GammaT,\GammaTH$ as defined on $\Scal(A,\Tcal)$. We bound $\eopt = \Rpop(\Shat)-\Rdag$ in two steps.

\smallskip\noindent\emph{Upper bound on $\Rpop(\Shat)$.} Since $\Shat\in\Scal(A,\Tcal)$,
\begin{equation}\label{eq:train-upper}
\Rpop(\Shat) \;\le\; \Rpopt(\Shat)+\GammaTH \;\le\; \Rempt(\Shat)+\GammaT+\GammaTH.
\end{equation}

\smallskip\noindent\emph{Lower bound on $\Rdag$.} For $\epsilon>0$, pick $S_\epsilon\in\Scal(A,\Tcal)$ with $\Rpop(S_\epsilon)\le\Rdag+\epsilon$. Then $S_\epsilon$ is admissible in the $\Rempdag$ infimum, and
\[
\Rempdag \;\le\; \Rempt(S_\epsilon) \;\le\; \Rpopt(S_\epsilon)+\GammaT \;\le\; \Rpop(S_\epsilon)+\GammaT+\GammaTH \;\le\; \Rdag+\epsilon+\GammaT+\GammaTH,
\]
using (in order) the infimum, $\GammaT$, $\GammaTH$, and the choice of $S_\epsilon$. Letting $\epsilon\to 0$,
\begin{equation}\label{eq:train-lower}
\Rempdag-\GammaT-\GammaTH \;\le\; \Rdag.
\end{equation}

\smallskip\noindent\emph{Combining.} Subtracting~\eqref{eq:train-lower} from~\eqref{eq:train-upper},
\[
\eopt \;=\; \Rpop(\Shat)-\Rdag \;\le\; \bigl[\Rempt(\Shat)-\Rempdag\bigr]+2(\GammaT+\GammaTH) \;=\; \widehat\rho+2(\GammaT+\GammaTH).
\]
Substituting into Theorem~\ref{thm:main} Part~I gives the displayed classification bound. This bound holds with probability $\ge1-5n^{-1}$, on the intersection of the above event and the event of Theorem~\ref{thm:main}.\qed

\subsection{Why the Relaxed Envelope}\label{app:train-side-design}

The covering arguments use the relaxed class $\Scal^{\Zcal}$ as an envelope rather than $\Scal(A,\Tcal)$ directly for two structural reasons. First, $\Scal^{\Zcal}$ is deterministic, so its $\varepsilon$-net has data-independent metric entropy; the encoder class $\Scal(A,\Tcal)$ depends on $\Aenc$ through $f_\theta$, and a covering would have to be constructed conditionally. Second, and more fundamentally, Hoeffding on $\GammaT$'s summands $(P_{ij}-A_{ij})s_{ij}$ requires $s_{ij}$ to be independent of $\{A_{ij}:(i,j)\in\Tcal\}$; for an encoder-induced $S$ this fails because $\Aenc$ is itself a deterministic function of $\{A_{ij}:(i,j)\in\Tcal\}$. The deviation bound therefore cannot be obtained by a direct concentration argument on $\Scal(A,\Tcal)$ without an auxiliary independence-restoring split (e.g.\ a sub-split of $\Tcal$ into encoder-input pairs and loss-evaluation pairs), which would alter the setting and decouple training from the loss the optimizer actually minimizes. Bounding the sup on $\Scal^{\Zcal}$ instead and transferring via inclusion preserves both the setting and the training objective.
\section{Architectural Realization of $\eapp$: Construction and Proof}\label{app:eapp}

\providecommand{\Pbar}{\bar P}
\providecommand{\Penc}{P^{\mathrm{enc}}}
\providecommand{\PE}{\mathrm{PE}}
\providecommand{\Uhat}{\widehat U_K}
\providecommand{\Ustar}{U_K^\star}
\providecommand{\Pihat}{\widehat \Pi_K}
\providecommand{\Pistar}{\Pi_K^\star}
\providecommand{\Lhat}{\widehat \Lambda_K}
\providecommand{\Lstar}{\Lambda_K^\star}
\providecommand{\Ohat}{\widehat O}
\providecommand{\dPE}{\delta_{\PE}}
\providecommand{\Wenc}{W^{\mathrm{enc}}}
\providecommand{\benc}{b^{\mathrm{enc}}}
\providecommand{\Lip}{\operatorname{Lip}}
\makeatletter
\@ifundefined{rank}{\newcommand{\rank}{\operatorname{rank}}}{}
\@ifundefined{diag}{\newcommand{\diag}{\operatorname{diag}}}{}
\@ifundefined{op}{\newcommand{\op}{\mathrm{op}}}{}
\makeatother

This appendix proves Prop.~\ref{prop:eapp-rate}. App.~\ref{app:eapp:setup}
fixes notation, the encoder, and assumptions \textbf{(A1)}--\textbf{(A5)}.
The Master Lemma (App.~\ref{app:eapp:master}) reduces a bound on $\eapp$ to
three conditions: node features concentrate around community prototypes, a
Lipschitz readout maps the prototypes to target embeddings, and a decoder
reproduces the oracle logits at those embeddings. Steps~A--D verify these
conditions for GIN with spectral PE. Step~A shows that the spectral PE
concentrates at rate $\sqrt{K\log n/n}$ uniformly over nodes. Step~B shows
that its population version is block-constant with separated prototypes.
Step~C shows that a GIN on $\Aenc$ realizes any readout of the PE given by a
2-layer multilayer perceptron (MLP) with rectified linear unit (ReLU)
activations, also with extra PE coordinates and more layers. Step~D constructs the
decoder. App.~\ref{app:eapp:synthesis} combines the steps, and
App.~\ref{app:eapp:discussion} states the scope of the construction.

\subsection{Setup and Notation}\label{app:eapp:setup}

We retain the setting of the main paper. In particular:
$P\in[0,1]^{n\times n}$ with $P_{ij}=p_{c_ic_j}$ is the filled population matrix ($P=\Pi \Theta\Pi^\top$, rank $K$); $A\in\{0,1\}^{n\times n}$ is the
hollow observed adjacency, $A_{ij}\sim\Ber(P_{ij})$ for $i<j$;
$\Hcal\subseteq\{(i,j):i<j\}$ is drawn uniformly among the size-$\lceil\tau N\rceil$ subsets of the $N:=\binom n2$ pairs (matching the fixed-size mask of Sec.~\ref{sec:setting}), and $\Tcal=\{(i,j):i<j\}\setminus\Hcal$;
$\Aenc$ is the symmetric (hollow) encoder input, defined for $i<j$ by $\Aenc_{ij}=A_{ij}\indic\{(i,j)\in \Tcal\}$, extended to $i>j$ by symmetry, with $\Aenc_{ii}=0$.
The oracle logits are $m^\star_{ab}=\logit(p_{ab})$, collected in $M^\star:=(m^\star_{ab})_{a,b\in[K]}$, and $\pi_a=|\{i:c_i=a\}|/n$ are the community proportions.

\paragraph{Spectral objects.}
Let $P=\sum_{k=1}^n\lambda_k^\star u_k^\star(u_k^\star)^\top$ with
$\|u_k^\star\|_2=1$, ordered by absolute value
$|\lambda_1^\star|\ge|\lambda_2^\star|\ge\cdots$. Since $\rank(P)=K$,
exactly $K$ eigenvalues are nonzero (possibly of mixed signs when $\Theta$ is
indefinite) and $\lambda_k^\star=0$ for $k>K$. Define the population
signal matrices
$\Ustar:=[u_1^\star,\ldots,u_K^\star]\in\Real^{n\times K}$,
$\Lstar:=\diag(\lambda_1^\star,\ldots,\lambda_K^\star)$,
$\Pistar:=\Ustar(\Ustar)^\top$, with signal-noise gap
$\gamma_K^\star:=|\lambda_K^\star|-|\lambda_{K+1}^\star|=|\lambda_K^\star|$.
Analogously, let $\hat\lambda_1,\ldots,\hat\lambda_n$ be the eigenvalues
of $\Aenc$ ordered by absolute value, with eigenvectors
$\phi_1,\ldots,\phi_n$, and define $\Uhat:=[\phi_1,\ldots,\phi_K]$,
$\Lhat:=\diag(\hat\lambda_1,\ldots,\hat\lambda_K)$,
$\Pihat:=\Uhat\Uhat^\top$. Once
$\|\Aenc-\Pbar^\circ\|_{\op}$ is below the gap (Lemma~\ref{app:eapp:lem:A3}),
Weyl's inequality separates the top-$K$ absolute-eigenvalue subspace of
$\Aenc$ from the remaining eigenspace; the basis is identified only modulo
an orthogonal factor. We will repeatedly use the sign-agnostic spectral identity
\begin{equation}\label{app:eapp:eq:spec-id}
\Aenc\,\Uhat \;=\; \Uhat\,\Lhat,
\end{equation}
which drives the native message-passing neural network (MPNN) argument in
Step~C.

\paragraph{Positional encoding.}
For each $i\in[n]$,
\begin{equation}\label{app:eapp:eq:pe-def}
\PE(i) := \sqrt n\cdot(\Uhat)_{i,:}\in\Real^K,\qquad
\PE^\star(i) := \sqrt n\cdot(\Ustar)_{i,:}\in\Real^K.
\end{equation}
The $\sqrt n$ rescaling brings entries to $O(1)$ scale
(cf.~\cite{dwivedi2020generalization}). By orthonormality, $\PE(i)$ is
defined only up to a sample-dependent $\Ohat\in O(K)$;
Lemma~\ref{app:eapp:lem:step-a} states concentration modulo $\Ohat$,
which Step~C absorbs into the readout's first linear layer.
The encoder may receive more coordinates: its input is
$X_{\mathrm{PE}}:=\sqrt n\,[\phi_1,\ldots,\phi_{D_{\mathrm{PE}}}]$ for a fixed
$D_{\mathrm{PE}}\ge K$, ordered by $|\hat\lambda_k|$. On the event of
Lemma~\ref{app:eapp:lem:A3}, Weyl's inequality gives
$|\hat\lambda_k|\ge\frac12\tau_0c_\gamma n$ for $k\le K$ and
$|\hat\lambda_k|\le C_{\op}\sqrt{n\log n}$ for $k>K$. For $n$ large the top-$K$
absolute eigenvalues are therefore separated from the rest, the first $K$
columns of $X_{\mathrm{PE}}$ are $\sqrt n\,\Uhat$, and $\PE(i)$ is the
restriction of the $i$-th row of $X_{\mathrm{PE}}$ to these columns.

\paragraph{Encoder architecture.}
We consider depth-$L$ MPNN encoders with node features
$X_{\mathrm{PE}}\in\Real^{n\times D_{\mathrm{PE}}}$ and GIN-style
updates~\cite{xu2019gin}
\begin{equation}\label{app:eapp:eq:mpnn-update}
H^{(l+1)}_{i,:} = \mathrm{MLP}^{(l)}\!\bigl((1+\epsilon^{(l)})H^{(l)}_{i,:} + (\Aenc H^{(l)})_{i,:}\bigr),
\quad H^{(0)}:=X_{\mathrm{PE}},
\end{equation}
where each $\mathrm{MLP}^{(l)}$ is a 2-layer ReLU MLP and $\epsilon^{(l)}$ is a
scalar parameter; a ReLU is applied between consecutive layers but not after
the last one. Write $\rho(x):=\max\{x,0\}$ for the ReLU. A $1$-Lipschitz projection
$\Pi_{B_R(0)}(x):=x\min(1,R/\|x\|)$ enforces $\|z_i\|\le R$, giving
$z_i:=\Pi_{B_R(0)}(H^{(L)}_{i,:})\in B_R(0)\subset\Real^d$. Other MPNN
families that share this aggregation operator may admit analogous
treatments; normalized aggregators (GCN, GraphSAGE) and attention-based
aggregators (GAT) require separate compatibility arguments
(App.~\ref{app:eapp:discussion}, App.~\ref{app:K4-collapse}).

\paragraph{Assumptions for Step A--D.}
\begin{itemize}
    \item \textbf{(A1)} (dense regime) $p_{ab}\in[p_{\min},1-p_{\min}]$ for constant $p_{\min}\in(0,1/2)$.
    \item \textbf{(A2)} (balanced communities) $\pi_a\geq\pi_{\min}$ for constant $\pi_{\min}>0$.
    \item \textbf{(A3)} (spectral regularity and embedding dimension) $\rank(\Theta)=K$; the $K$ nonzero eigenvalues of $P$ (possibly of mixed signs) satisfy $\min_{k\leq K}|\lambda_k^\star|\geq c_\gamma n$ for constant $c_\gamma>0$; embedding dimension $d\geq K$. Equivalently, $\gamma_K^\star = |\lambda_K^\star| - |\lambda_{K+1}^\star| = |\lambda_K^\star|\geq c_\gamma n$, since $\lambda_{K+1}^\star=\cdots=\lambda_n^\star=0$.
    \item \textbf{(A4)} (nontrivial masking) $\tau\in(0,1-\tau_0]$ for constant $\tau_0>0$.
    \item \textbf{(A5)} (bilinear capacity) $\Wmax R^2 \geq \min_{|b|\le B_0}\|M^\star - b\,\mathbf{1}\mathbf{1}^\top\|_{\op}$, with the minimum taken over the admissible bias range $|b|\le B_0$ (cf.\ Prop.~\ref{prop:bilinear-realizability} on community-level decoder realizability).
\end{itemize}

\subsection{Master Lemma: Abstract Feature-to-Score Bound}\label{app:eapp:master}

The following lemma is the organizational backbone of the proof. It
shows that any encoder design satisfying three modular conditions
(feature concentration, Lipschitz readout, exact decoder) yields a
quantitative $\eapp$ bound.

\begin{lemma}[Master Lemma]\label{app:eapp:lem:master}
Let $\phi_i \in \Real^q$ be a sample-dependent feature for each
$i\in[n]$. Let $\mu_1,\ldots,\mu_K \in \Real^q$ be deterministic prototypes
and $g:\Real^q\to\Real^d$ a deterministic Lipschitz readout. Suppose:
\begin{enumerate}[label=(M\arabic*),nosep,leftmargin=2em]
\item \textbf{(Concentration)} On an event $E_n$ with $\mathbb{P}(E_n)\geq 1-n^{-c}$,
\[
\max_{i\in[n]} \|\phi_i - \mu_{c_i}\|_2 \;\leq\; r_n.
\]
\item \textbf{(Readout exactness and Lipschitzness)} For each $a\in[K]$, $g(\mu_a) = u_a$ for some $u_a\in\Real^d$ with $\|u_a\|_2\leq R_u$, and $g$ is $L_g$-Lipschitz on $\bigcup_a B_{r_n}(\mu_a)$.
\item \textbf{(Decoder exactness)} There exist symmetric $W\in\Real^{d\times d}$ with $\|W\|_{\op}\leq\Wmax$ and $b\in\Real$ such that
\[
b + u_a^\top W u_b \;=\; m^\star_{ab}\qquad\forall\,a,b\in[K].
\]
\end{enumerate}
Define $z_i := g(\phi_i)$ and $s_{ij} := b + z_i^\top W z_j$. Then on $E_n$,
\[
\max_{i,j\in[n]}|s_{ij} - m^\star_{c_ic_j}|
\;\leq\;
\Wmax\bigl(2R_u L_g r_n + L_g^2 r_n^2\bigr).
\]
If, in addition, the score matrix $S=(s_{ij})$ belongs to the hypothesis
class $\Scal(A,\Tcal)$ of Sec.~\ref{sec:setting}, then on $E_n$
\[
\eapp \;\leq\; \tfrac{1}{8}\Wmax^2\bigl(2R_u L_g r_n + L_g^2 r_n^2\bigr)^2.
\]
\end{lemma}

\begin{proof}
Write $\Delta_i := z_i - u_{c_i}$; by (M1) and (M2), $\|\Delta_i\|\le L_g r_n$ on $E_n$.
Expanding $s_{ij} = b + (u_{c_i}+\Delta_i)^\top W(u_{c_j}+\Delta_j)$
and using (M3) to cancel $b + u_{c_i}^\top W u_{c_j} = m^\star_{c_ic_j}$,
\[
|s_{ij} - m^\star_{c_ic_j}|
\;=\; |u_{c_i}^\top W\Delta_j + \Delta_i^\top W u_{c_j} + \Delta_i^\top W\Delta_j|
\;\leq\; \Wmax(2R_u L_g r_n + L_g^2 r_n^2) \;=:\; \delta_n.
\]
The BCE loss $\ell(s;p) = -p\log\sigma(s) - (1-p)\log(1-\sigma(s))$ is
$1/4$-smooth on $\Real$ (since $\sigma(s)(1-\sigma(s))\le 1/4$) and
minimized at $s=\logit p$, so
$\ell(s;p) - \ell(\logit p;p) \le \tfrac{1}{8}(s-\logit p)^2$. Since
$S\in\Scal(A,\Tcal)$, it is feasible for the in-class infimum $\Rdag$ in
$\eapp=\Rdag-\Rpop(\Sopt)$. Averaging over $\Hcal$,
\[
\eapp \;\leq\; \Rpop(S) - \Rpop(\Sopt)
\;\leq\; \frac{1}{8|\Hcal|}\sum_{(i,j)\in\Hcal}(s_{ij}-m^\star_{c_ic_j})^2
\;\leq\; \tfrac{1}{8}\delta_n^2.\qedhere
\]
\end{proof}

\subsection{Step A: Spectral PE Concentration}\label{app:eapp:step-a}

The main result of this section provides the $r_n$ in the Master Lemma:
\begin{lemma}[Step A: PE concentration]\label{app:eapp:lem:step-a}
Under \textbf{(A1)}--\textbf{(A4)}, there exist constants $C_A, c_A > 0$
depending only on $(K, p_{\min}, \pi_{\min}, c_\gamma, \tau_0)$ such that,
with probability at least $1 - n^{-c_A}$, there exists a sample-dependent
orthogonal matrix $\Ohat\in O(K)$ satisfying
\begin{equation}\label{app:eapp:eq:step-a}
\max_{i\in[n]} \|\PE(i)\Ohat - \PE^\star(i)\|_2
\;\leq\; r_n := C_A\sqrt{\frac{K\log n}{n}}.
\end{equation}
\end{lemma}

\paragraph{Proof route.}
We use Cape--Tang--Priebe only through their deterministic Procrustean
$2\to\infty$ decomposition~\cite[Theorem~3.7]{cape2019twotoinfinity}. The
fixed-size mask is handled by comparison with an independent mask
(Lemma~\ref{app:eapp:lem:A1}(iv)). Lemma~\ref{app:eapp:lem:A3} gives an
operator-norm event, and Lemma~\ref{app:eapp:lem:A4} gives the row-wise
$E\Ustar$ event needed by the leading term of their bound.

\begin{lemma}[Masked adjacency under fixed-size sampling]\label{app:eapp:lem:A1}
Let $A$ be as in the main paper with $\{A_{ij}\}_{i<j}$ mutually independent, and let $\Hcal$ be uniform on size-$\lceil\tau N\rceil$ subsets of $\{(i,j):i<j\}$, independent of $A$. Set $\eta:=|\Tcal|/N=1-\lceil\tau N\rceil/N=1-\tau+O(1/N)$. Then:
\begin{enumerate}[label=(\roman*),nosep]
\item $\Aenc_{ij}$ has marginal $\Ber(\eta P_{ij})$ for $i<j$.
\item The mask indicators $\{\indic\{(i,j)\in\Tcal\}\}_{i<j}$ are exchangeable but not independent, because $|\Tcal|$ is fixed; they are negatively associated~\citep{joag1983negative}. We never treat $\Aenc$ as an independent-entry matrix. The proofs below handle the mask through~(iv) and the Bernoulli edge noise conditionally on the mask.
\item $\mathbb{E}\Aenc = \Pbar$, where $\Pbar_{ij} = \eta P_{ij}$ for $i\neq j$ and $\Pbar_{ii}=0$. Writing $\Pbar^\circ := (1-\tau)P$ as the rank-$K$ reference, $\|\Pbar - \Pbar^\circ\|_{\op} = O(1)$ (diagonal correction plus the $O(1/N)$ rounding shift).
\item \emph{(Independent-mask comparison.)} Let $\{\xi_{ij}\}_{i<j}$ be independent $\Ber(\eta)$ variables, and write $\mathbb P_{\mathrm{fix}}$ and $\mathbb P_{\mathrm{ind}}$ for the laws of $\{\indic\{(i,j)\in\Tcal\}\}_{i<j}$ and of $\{\xi_{ij}\}_{i<j}$. For every event $\mathcal B$ determined by the mask,
\[
\mathbb P_{\mathrm{fix}}[\mathcal B]\;\le\;(N+1)\,\mathbb P_{\mathrm{ind}}[\mathcal B]\;\le\; n^2\,\mathbb P_{\mathrm{ind}}[\mathcal B].
\]
\end{enumerate}
\end{lemma}

\begin{proof}
(i) $\mathbb{P}[\Aenc_{ij}=1] = \mathbb{P}[A_{ij}=1,(i,j)\in\Tcal] = P_{ij}\cdot|\Tcal|/N = \eta P_{ij}$ by independence of $A$ and $\Hcal$.
(ii) $\Tcal$ is a uniformly random subset of fixed size, so its indicators are exchangeable and sum to $|\Tcal|$; sampling without replacement yields negative association~\citep{joag1983negative}.
(iii) From (i) off-diagonally; the diagonal is zero by $A_{ii}=0$. The norm bound follows from the triangle inequality: the diagonal contribution has operator norm $\le 1$, and the off-diagonal $O(1/N)$ rounding shift contributes $O(1/N)\|P\|_{\op}=O(1)$.
(iv) The uniform law on size-$|\Tcal|$ subsets equals the law of $\{\xi_{ij}\}$ conditioned on $\sum_{i<j}\xi_{ij}=|\Tcal|$. Since $N\eta=|\Tcal|$ is an integer, it is a mode of $\Bin(N,\eta)$, so $\mathbb P_{\mathrm{ind}}[\sum_{i<j}\xi_{ij}=|\Tcal|]\ge(N+1)^{-1}$. Hence $\mathbb P_{\mathrm{fix}}[\mathcal B]=\mathbb P_{\mathrm{ind}}[\mathcal B\mid\sum_{i<j}\xi_{ij}=|\Tcal|]\le(N+1)\,\mathbb P_{\mathrm{ind}}[\mathcal B]$, and $N+1\le n^2$.
\end{proof}

\begin{lemma}[Spectral invariants]\label{app:eapp:lem:A2}
$P$ and $\Pbar^\circ=(1-\tau)P$ share eigenvectors, with eigenvalues
scaled by $(1-\tau)$ (preserving signs). In particular, under
\textbf{(A3)}--\textbf{(A4)}, the $K$ nonzero eigenvalues of
$\Pbar^\circ$ all satisfy
\[
\min_{k\leq K}|\lambda_k(\Pbar^\circ)|
\;\geq\; \tau_0\,c_\gamma\, n \;=\; \Omega(n),
\]
so the signal-noise gap
$|\lambda_K(\Pbar^\circ)| - |\lambda_{K+1}(\Pbar^\circ)| = |\lambda_K(\Pbar^\circ)|\geq \tau_0\,c_\gamma\, n$
is $\Omega(n)$.
\end{lemma}

\begin{proof}
Immediate from $\Pbar^\circ=(1-\tau)P$ with $1-\tau\geq \tau_0$ by (A4)
and (A3) on $|\lambda_k^\star|$.
\end{proof}

In particular, $\Ustar$ remains an orthonormal basis for the rank-$K$ column space of $\Pbar^\circ$. We write $E := \Aenc - \Pbar^\circ$ throughout the rest of Step~A.

\begin{lemma}[Operator-norm concentration]\label{app:eapp:lem:A3}
Under \textbf{(A1)}--\textbf{(A4)}, there exist constants
$C_{\op},c_{\op}>0$ such that
\[
\mathbb{P}\bigl[\|E\|_{\op} = \|\Aenc-\Pbar^\circ\|_{\op} \leq C_{\op}\sqrt{n\log n}\bigr] \;\geq\; 1-n^{-c_{\op}}.\tag{E1}
\]
\end{lemma}

\begin{proof}
Let $M_{ij}:=\indic\{(i,j)\in\Tcal\}$ for $i<j$, extend $M$ symmetrically,
and set $M_{ii}=0$. Let $J:=\mathbf 1\mathbf 1^\top-I_n$ and
$D_P:=\diag(P_{11},\ldots,P_{nn})$. Since $\eta=|\Tcal|/N=1-\tau+O(N^{-1})$,
we decompose
\begin{equation}\label{app:eapp:eq:E-decomp}
E=\Aenc-\Pbar^\circ
= M\circ(A-P) +(M-\eta J)\circ P
+ (\eta-(1-\tau))(J\circ P) -(1-\tau)D_P .
\end{equation}
The last two terms are deterministic given the partition and have operator
norm $O(1)$: indeed $|\eta-(1-\tau)|=O(N^{-1})$, $\|P\|_{\op}=O(n)$, and
$\|D_P\|_{\op}\le 1$.

For the first random term, condition on $M$. The upper-triangular entries of
$M\circ(A-P)$ are independent, centered, bounded by $1$, and have matrix
variance proxy at most $Cn$ because each summand
$M_{ij}(A_{ij}-P_{ij})(e_ie_j^\top+e_je_i^\top)$ has squared expectation
bounded by $M_{ij}P_{ij}(1-P_{ij})(e_ie_i^\top+e_je_j^\top)$ and the maximum
row sum is at most $n$. Matrix Bernstein~\citep{tropp2012user} therefore gives, uniformly in the
realized mask,
\[
\mathbb P\{\|M\circ(A-P)\|_{\op}>C\sqrt{n\log n}\mid M\}\le n^{-c}
\]
for suitable constants $C,c>0$. Integrating over $M$ preserves the same tail.

For the mask-noise term, compare with an independent Bernoulli mask
(Lemma~\ref{app:eapp:lem:A1}(iv)). With $\{\xi_{ij}\}_{i<j}$ as there, set
\[
Q_{\mathrm{ind}}
:=\sum_{i<j}(\xi_{ij}-\eta)P_{ij}(e_ie_j^\top+e_je_i^\top).
\]
The summands are independent, centered, have operator norm at most $1$, and
have variance proxy at most
\[
\Bigl\|\sum_{i<j} \eta(1-\eta)P_{ij}^2(e_ie_i^\top+e_je_j^\top)\Bigr\|_{\op}
\le n .
\]
Thus matrix Bernstein gives
\[
\mathbb P\{\|Q_{\mathrm{ind}}\|_{\op}>C\sqrt{n\log n}\}\le n^{-c-3}
\]
for $C$ large enough. Under the fixed-size mask, $(M-\eta J)\circ P$ is the
same function of the indicators $\{M_{ij}\}_{i<j}$ as $Q_{\mathrm{ind}}$ is of
$\{\xi_{ij}\}_{i<j}$, so Lemma~\ref{app:eapp:lem:A1}(iv) gives
\[
\mathbb P\{\|(M-\eta J)\circ P\|_{\op}>C\sqrt{n\log n}\}
\le n^{2}\cdot n^{-c-3}=n^{-c-1}.
\]
Combining the two random terms with the $O(1)$ deterministic remainder in
\eqref{app:eapp:eq:E-decomp} proves the claim after enlarging $C_{\op}$ and
renaming the exponent.
\end{proof}

\begin{lemma}[Row-wise eigenvector noise]\label{app:eapp:lem:A4}
Let $\Ustar\in\Real^{n\times K}$ be the rank-$K$ population eigenbasis of $\Pbar^\circ$. Under \textbf{(A1)}--\textbf{(A4)}, there exist constants $C_{\mathrm{row}},c_{\mathrm{row}}>0$ such that
\[
\mathbb{P}\bigl[\|E\,\Ustar\|_{2\to\infty} \leq C_{\mathrm{row}}\sqrt{K\log n}\bigr] \;\geq\; 1-n^{-c_{\mathrm{row}}}.\tag{E2}
\]
\end{lemma}

\begin{proof}
Block-constancy and orthonormality (Lemma~\ref{app:eapp:lem:step-b} below) give $\|\Ustar\|_{2\to\infty}\le\sqrt{K/(n\pi_{\min})}$ and column normalization $\sum_{j=1}^n(\Ustar)_{j,\ell}^2=1$ for each $\ell\in[K]$. Fix $i\in[n]$ and $\ell\in[K]$, and decompose
\[
(E\Ustar)_{i,\ell} \;=\; \underbrace{\sum_{j\ne i}(\Aenc_{ij}-\Pbar_{ij})(\Ustar)_{j,\ell}}_{(\mathrm{I})}
\;+\; \underbrace{\sum_{j\ne i}(\Pbar_{ij}-\Pbar_{ij}^\circ)(\Ustar)_{j,\ell}}_{(\mathrm{II})}
\;+\; \underbrace{(-\Pbar^\circ_{ii})(\Ustar)_{i,\ell}}_{(\mathrm{III})}.
\]
Term~(III) is $|(1-\tau)P_{ii}\,(\Ustar)_{i,\ell}|\le\|\Ustar\|_{2\to\infty}=O(1/\sqrt n)$. Term~(II) is $O(1/N)\sum_{j}P_{ij}|(\Ustar)_{j,\ell}|\le O(1/N)\sqrt{n}=O(1/\sqrt n)$ by Cauchy--Schwarz with $\sum_j P_{ij}^2\le n$. Both are $o(\sqrt{\log n})$ uniformly.

\emph{Bounding term (I) via Bernstein.} Decompose $\Aenc_{ij}-\Pbar_{ij} = M_{ij}(A_{ij}-P_{ij}) + (M_{ij}-\eta)P_{ij}$ where $M_{ij}:=\indic\{(\min(i,j),\max(i,j))\in\Tcal\}$. Substituting,
\[
(\mathrm{I}) \;=\; \underbrace{\sum_{j\ne i}M_{ij}(A_{ij}-P_{ij})(\Ustar)_{j,\ell}}_{(\mathrm{Ia})}
\;+\; \underbrace{\sum_{j\ne i}(M_{ij}-\eta)P_{ij}(\Ustar)_{j,\ell}}_{(\mathrm{Ib})}.
\]

\emph{(Ia): conditional Bernstein.} Conditional on the mask $M$, the summands are independent (across $j$), centered, bounded by $\|\Ustar\|_{2\to\infty}\le C/\sqrt n$, with conditional variance
\[
\sum_{j\ne i}M_{ij}P_{ij}(1-P_{ij})(\Ustar)_{j,\ell}^2 \;\le\; \tfrac{1}{4}\sum_{j}(\Ustar)_{j,\ell}^2 \;=\; \tfrac{1}{4}.
\]
Standard Bernstein gives $\mathbb{P}[|(\mathrm{Ia})|>t\mid M]\le 2\exp(-c t^2)$ for $t\le c'\sqrt n$. Marginalizing over $M$ preserves the bound.

\emph{(Ib): independent-mask comparison.} Replace the mask indicators by
independent $\xi_{ij}\sim\Ber(\eta)$ as in Lemma~\ref{app:eapp:lem:A1}(iv).
The resulting sum $\sum_{j\ne i}(\xi_{ij}-\eta)P_{ij}(\Ustar)_{j,\ell}$ has
independent, centered summands bounded by $\|\Ustar\|_{2\to\infty}\le C/\sqrt n$
and variance at most $\eta(1-\eta)\sum_j(\Ustar)_{j,\ell}^2\le\tfrac14$, so
Bernstein's inequality gives the tail $2\exp(-ct^2)$ for $t\le c'\sqrt n$.
Since (Ib) is determined by the mask, Lemma~\ref{app:eapp:lem:A1}(iv)
transfers this tail to the fixed-size mask at the cost of a factor $n^2$:
$\mathbb{P}[|(\mathrm{Ib})|>t]\le 2n^2\exp(-ct^2)$ for $t\le c'\sqrt n$.

\emph{Union bound.} Take $t:=\tfrac13C_{\mathrm{row}}\sqrt{\log n}$, which is
at most $c'\sqrt n$ for $n$ large. Then (Ia) and (Ib) each exceed $t$ with
probability at most $2n^{2-cC_{\mathrm{row}}^2/9}$, while
$(\mathrm{II})+(\mathrm{III})=o(\sqrt{\log n})\le t$ for $n$ large.
Union-bounding over $(i,\ell)\in[n]\times[K]$, with probability at least
$1-4Kn^{3-cC_{\mathrm{row}}^2/9}$ we have
$|(E\Ustar)_{i,\ell}|\le C_{\mathrm{row}}\sqrt{\log n}$ for all $(i,\ell)$, and hence
\[
\|E\Ustar\|_{2\to\infty} = \max_i\sqrt{\sum_{\ell}|(E\Ustar)_{i,\ell}|^2} \;\le\; \sqrt{K}\cdot C_{\mathrm{row}}\sqrt{\log n}.
\]
Choosing $C_{\mathrm{row}}$ large enough gives~\textbf{(E2)} for any prescribed exponent $c_{\mathrm{row}}$.
\end{proof}

\begin{lemma}[$2\to\infty$ subspace concentration]\label{app:eapp:lem:A5}
Under \textbf{(A1)}--\textbf{(A4)}, there exist constants
$C_{\mathrm{CTP}},c_{\mathrm{CTP}}>0$ such that, with probability at least
$1-n^{-c_{\mathrm{CTP}}}$, there exists $\Ohat\in O(K)$ satisfying
\[
\|\Uhat\Ohat-\Ustar\|_{2\to\infty}
\;\leq\; C_{\mathrm{CTP}}\cdot\frac{\sqrt{K\log n}}{n}.
\]
\end{lemma}

\begin{proof}
Write $\Aenc=\Pbar^\circ+E$ and $\sigma_K:=\sigma_K(\Pbar^\circ)=|\lambda_K(\Pbar^\circ)|$. We apply Cape--Tang--Priebe's deterministic Procrustean two-to-infinity bound~\cite[Theorem~3.7]{cape2019twotoinfinity} to this perturbation of the rank-$K$ symmetric matrix $\Pbar^\circ$. Since $\Pbar^\circ$ may be indefinite, write its rank-$K$ eigendecomposition as $\Pbar^\circ=\Ustar\Lambda^\circ(\Ustar)^\top$ with $\Lambda^\circ:=(1-\tau)\Lstar$, and set $S:=\diag(\operatorname{sign}\lambda_1^\star,\ldots,\operatorname{sign}\lambda_K^\star)$. In the singular value decomposition (SVD) notation of Cape--Tang--Priebe one may take the left and right singular-vector matrices to be $\Ustar$ and $\Ustar S$; both projectors then equal $\Pistar$, and right multiplication by $S$ does not change any $2\to\infty$ norm. Since $\rank(\Pbar^\circ)=K$, the term in their decomposition involving $\Pbar^\circ(I-\Pistar)$ vanishes. For the symmetric matrix $\Aenc$, the left singular vectors of the $K$ largest singular values are the eigenvectors $\Uhat$ of the $K$ largest absolute eigenvalues. Absorbing the fixed sign matrix into the Procrustes factors, Theorem~3.7 yields the existence of $\Ohat\in O(K)$ such that, whenever $\sigma_K\ge 2\|E\|_{\op}$,
\begin{equation}\label{app:eapp:eq:cape-decomp}
\begin{aligned}
\|\Uhat\Ohat-\Ustar\|_{2\to\infty}
\le{}& 2\,\frac{\|(I-\Pistar)E\Ustar\|_{2\to\infty}}{\sigma_K}\\
&+2\,\frac{\|(I-\Pistar)E(I-\Pistar)\|_{2\to\infty}}{\sigma_K}\,\|\sin\Theta(\Uhat,\Ustar)\|_{\op}\\
&+\|\sin\Theta(\Uhat,\Ustar)\|_{\op}^2\,\|\Ustar\|_{2\to\infty}.
\end{aligned}
\end{equation}

\emph{Verifying applicability.} On the event \textbf{(E1)} of Lemma~\ref{app:eapp:lem:A3}, $\|E\|_{\op}\le C_{\op}\sqrt{n\log n}$. By Lemma~\ref{app:eapp:lem:A2}, $\sigma_K\ge\tau_0c_\gamma n$. Hence $\|E\|_{\op}/\sigma_K=O(\sqrt{\log n/n})\to0$, so $\sigma_K\ge2\|E\|_{\op}$ for all sufficiently large $n$. Weyl separates the top-$K$ absolute-eigenvalue subspace of $\Aenc$ from the rest, and Wedin's $\sin\Theta$ theorem~\citep{wedin1972perturbation} (applied to the leading-$K$ singular subspaces, which for symmetric matrices are the top-$K$ absolute-eigenvalue subspaces), together with $\sigma_K(\Aenc)\ge\sigma_K-\|E\|_{\op}\ge\sigma_K/2$, gives
\begin{equation*}\label{app:eapp:eq:dk}
\|\sin\Theta(\Uhat,\Ustar)\|_{\op}\le\frac{2\|E\|_{\op}}{\sigma_K}\le C_{\sin}\sqrt{\frac{\log n}{n}}.\tag{E3}
\end{equation*}

\emph{Bounding the three terms in~\eqref{app:eapp:eq:cape-decomp}.}
We use Cape's constrained sub-multiplicativity~\cite[Prop.~6.5]{cape2019twotoinfinity}: $\|CA\|_{2\to\infty}\le\|C\|_\infty\|A\|_{2\to\infty}$ and $\|AB\|_{2\to\infty}\le\|A\|_{2\to\infty}\|B\|_{\op}$, where $\|C\|_\infty$ is the maximum absolute row sum. Block-constancy (Lemma~\ref{app:eapp:lem:step-b}) gives $\|\Ustar\|_{2\to\infty}\le\sqrt{K/(n\pi_{\min})}$, and $\Pistar$ is the block-average projector, so each row of $\Pistar$ has absolute row sum $1$ and $\|I-\Pistar\|_\infty\le2$.

\emph{Term 1.} Combining $\|I-\Pistar\|_\infty\le2$ with the row-wise event \textbf{(E2)} of Lemma~\ref{app:eapp:lem:A4},
\[
\|(I-\Pistar)E\Ustar\|_{2\to\infty}
\le2\|E\Ustar\|_{2\to\infty}
\le2C_{\mathrm{row}}\sqrt{K\log n}.
\]
Hence the first term of~\eqref{app:eapp:eq:cape-decomp} is at most
\[
\frac{4C_{\mathrm{row}}}{\tau_0c_\gamma}\frac{\sqrt{K\log n}}{n}.
\]

\emph{Term 2.} Since every entry of $E$ is bounded by $1$ in absolute value, $\|E\|_{2\to\infty}\le\sqrt n$. Then
\[
\|(I-\Pistar)E(I-\Pistar)\|_{2\to\infty}
\le \|I-\Pistar\|_\infty\|E(I-\Pistar)\|_{2\to\infty}
\le2\|E\|_{2\to\infty}\|I-\Pistar\|_{\op}
\le2\sqrt n.
\]
Combined with $\sigma_K\ge\tau_0c_\gamma n$ and~\eqref{app:eapp:eq:dk},
\[
\frac{\|(I-\Pistar)E(I-\Pistar)\|_{2\to\infty}}{\sigma_K}\|\sin\Theta\|_{\op}
\le \frac{2\sqrt n}{\tau_0c_\gamma n}\cdot C_{\sin}\sqrt{\frac{\log n}{n}}
=O\left(\frac{\sqrt{\log n}}{n}\right).
\]

\emph{Term 3.}
\[
\|\sin\Theta\|_{\op}^2\|\Ustar\|_{2\to\infty}
\le C_{\sin}^2\frac{\log n}{n}\sqrt{\frac{K}{n\pi_{\min}}}
=O\left(\frac{\sqrt K\log n}{n^{3/2}}\right)
=o\left(\frac{\sqrt{K\log n}}{n}\right).
\]

\emph{Combining.} Terms~1 and~2 are $O(\sqrt{K\log n}/n)$ and Term~3 is of lower order, so
\[
\|\Uhat\Ohat-\Ustar\|_{2\to\infty}\le C_{\mathrm{CTP}}\frac{\sqrt{K\log n}}{n},
\]
where $C_{\mathrm{CTP}}$ depends only on $(K,p_{\min},\pi_{\min},c_\gamma,\tau_0)$. The probability statement follows from union-bounding the events of Lemmas~\ref{app:eapp:lem:A3} and~\ref{app:eapp:lem:A4}.
\end{proof}

\paragraph{Assembly.}

\begin{proof}[Proof of Lemma~\ref{app:eapp:lem:step-a}]
On the event of Lemma~\ref{app:eapp:lem:A5},
\begin{align*}
\max_i\|\PE(i)\Ohat-\PE^\star(i)\|_2
&= \sqrt n\cdot\max_i\|(\Uhat\Ohat)_{i,:}-(\Ustar)_{i,:}\|_2\\
&\leq \sqrt n\cdot C_{\mathrm{CTP}}\,\frac{\sqrt{K\log n}}{n}
= C_{\mathrm{CTP}}\sqrt{\frac{K\log n}{n}}.
\end{align*}
This proves the lemma with $C_A:=C_{\mathrm{CTP}}$ and $c_A:=c_{\mathrm{CTP}}$.
\end{proof}

\subsection{Step B: Population PE Separates Communities}\label{app:eapp:step-b}

\begin{lemma}[Block-constancy and separation]\label{app:eapp:lem:step-b}
Under \textbf{(A1)}--\textbf{(A3)}, the population PE $\PE^\star(\cdot)$
takes $K$ distinct values:
\[
\PE^\star(i) = \mu_{c_i}^\star\qquad\text{for some }\mu_1^\star,\ldots,\mu_K^\star\in\Real^K.
\]
The pairwise distances satisfy, exactly,
\begin{equation}\label{app:eapp:eq:sep-exact}
\|\mu_a^\star - \mu_b^\star\|_2^2 \;=\; \frac{1}{\pi_a} + \frac{1}{\pi_b}\qquad\forall\,a\neq b,
\end{equation}
and in particular
\begin{equation}\label{app:eapp:eq:sep-bound}
\dPE := \min_{a\neq b}\|\mu_a^\star - \mu_b^\star\|_2 \;\geq\; \sqrt 2.
\end{equation}
\end{lemma}

\begin{proof}
Write $D_\pi:=\diag(\pi_1,\ldots,\pi_K)$, so $\Pi^\top\Pi=nD_\pi$ and
$D_\pi\succ 0$ by (A2). By (A3), $\rank(\Theta)=K$, hence $\rank(P)=K$ and
$\lambda_k^\star\ne 0$ exactly for $k\in[K]$ (signs unconstrained; the
argument below is sign-agnostic). For each $k\in[K]$,
\[
u_k^\star = \frac{1}{\lambda_k^\star}Pu_k^\star = \frac{1}{\lambda_k^\star}\Pi \Theta\Pi^\top u_k^\star = \Pi v_k,
\qquad v_k := \frac{1}{\lambda_k^\star}\Theta\Pi^\top u_k^\star\in\Real^K,
\]
so $u_k^\star(i)=(v_k)_{c_i}$ depends only on $c_i$. Stacking
$V:=[v_1,\ldots,v_K]\in\Real^{K\times K}$:
$\PE^\star(i)=\sqrt n V_{c_i,:}=:\mu_{c_i}^\star$, proving block-constancy.

Orthonormality $\langle u_k^\star,u_l^\star\rangle=\delta_{kl}$ with
$u_k^\star=\Pi v_k$ yields $V^\top D_\pi V=\tfrac{1}{n}I_K$. Hence
\begin{equation}\label{app:eapp:eq:Qorth}
Q := \sqrt n\,D_\pi^{1/2}V\quad\text{satisfies}\quad Q^\top Q=I_K,\ \text{i.e.,}\ Q\in O(K),
\end{equation}
and we obtain $V=D_\pi^{-1/2}Q/\sqrt n$ and hence
$\mu_a^\star=\sqrt n V_{a,:}=Q_{a,:}/\sqrt{\pi_a}$. Since $QQ^\top=I_K$,
the rows of $Q$ are orthonormal, so for $a\ne b$,
\[
\|\mu_a^\star-\mu_b^\star\|^2
=\frac{\|Q_{a,:}\|^2}{\pi_a}-\frac{2\langle Q_{a,:},Q_{b,:}\rangle}{\sqrt{\pi_a\pi_b}}+\frac{\|Q_{b,:}\|^2}{\pi_b}
=\frac{1}{\pi_a}+\frac{1}{\pi_b},
\]
proving~\eqref{app:eapp:eq:sep-exact}. Since $\pi_a,\pi_b\le 1$,
$\dPE^2\ge 2$, giving~\eqref{app:eapp:eq:sep-bound}.
\end{proof}

\subsection{Step C: Native MPNN Realization}\label{app:eapp:step-c}

The spectral identity $\Aenc\Uhat=\Uhat\Lhat$ lets one GIN layer implement an
MLP on the spectral PE. We first treat one layer with exactly $K$ PE
coordinates and then extend the construction to the encoders used in practice.

\paragraph{Spectral Aggregation Identity.}

\begin{lemma}[MPNN aggregation acts as eigenvalue scaling]\label{app:eapp:lem:mpnn-spectral}
Let $X = \sqrt n\,\Uhat\in\Real^{n\times K}$ be the spectral PE matrix.
Then
\[
\Aenc\,X \;=\; X\,\Lhat,
\]
i.e., for each row $i$ and each column $k$:
\[
(\Aenc X)_{i,k} = \hat\lambda_k\cdot X_{i,k}.
\]
\end{lemma}

\begin{proof}
Each column of $X$ is $\sqrt n\cdot\phi_k$, an eigenvector of $\Aenc$ with
eigenvalue $\hat\lambda_k$. Hence $\Aenc(\sqrt n\,\phi_k) = \hat\lambda_k\cdot\sqrt n\,\phi_k$,
which in matrix form is $\Aenc X = X\Lhat$.
\end{proof}

\paragraph{Single-Layer GIN Realizes MLP-on-PE.}

\begin{lemma}[Native MPNN realizability]\label{app:eapp:lem:mpnn-realize}
Let $D_{\mathrm{PE}}=K$, so that $H^{(0)}=X=\sqrt n\,\Uhat$, and consider a
1-layer GIN of the form~\eqref{app:eapp:eq:mpnn-update} with internal 2-layer
ReLU MLP
$\mathrm{MLP}^{(0)}(\xi)=W_2^{(0)}\rho(W_1^{(0)}\xi+b_1^{(0)})+b_2^{(0)}$
of width $m=O(K)$. Choose $\epsilon^{(0)}>-1$ such that
$\Gamma:=(1+\epsilon^{(0)})I_K+\Lhat$ is invertible (this holds with high probability for
any fixed $\epsilon^{(0)}$ once $n$ is large, since by Weyl plus
Lemma~\ref{app:eapp:lem:A3}, $\min_k|\hat\lambda_k|\ge\frac{1}{2}\tau_0 c_\gamma n$
with high probability). Then for any target 2-layer ReLU MLP
$\tilde g(\xi)=\tilde W_2\rho(\tilde W_1\xi+\tilde b_1)+\tilde b_2$
of width $m$ and any $\Ohat\in O(K)$, there exist GIN parameters such
that $H^{(1)}_{i,:}=\tilde g(\PE(i)\Ohat)$ for all $i\in[n]$.
\end{lemma}

\begin{proof}
By~\eqref{app:eapp:eq:mpnn-update} and Lemma~\ref{app:eapp:lem:mpnn-spectral},
$H^{(1)}_{i,:}=\mathrm{MLP}^{(0)}((1+\epsilon^{(0)})X_{i,:}+X_{i,:}\Lhat)
=\mathrm{MLP}^{(0)}(X_{i,:}\Gamma)$.
Choose $W_1^{(0)}:=\tilde W_1\Ohat^\top\Gamma^{-1}$,
$b_1^{(0)}:=\tilde b_1$, $W_2^{(0)}:=\tilde W_2$, $b_2^{(0)}:=\tilde b_2$.
Using column-vector inputs ($(X_{i,:}\Gamma)^\top=\Gamma X_{i,:}^\top$, since
$\Gamma$ is diagonal) and $\Ohat^\top X_{i,:}^\top=(X_{i,:}\Ohat)^\top$,
$W_1^{(0)}(X_{i,:}\Gamma)^\top=\tilde W_1\Ohat^\top\Gamma^{-1}\Gamma X_{i,:}^\top=\tilde W_1(X_{i,:}\Ohat)^\top$,
so
$\mathrm{MLP}^{(0)}(X_{i,:}\Gamma)=\tilde W_2\rho(\tilde W_1(X_{i,:}\Ohat)^\top+\tilde b_1)+\tilde b_2=\tilde g(\PE(i)\Ohat)$.
\end{proof}

\paragraph{Extra PE coordinates and depth.}
Lemma~\ref{app:eapp:lem:mpnn-realize} uses one layer and exactly $K$ PE
coordinates. The next lemma shows that the same output is realizable by the
encoders used in practice, which have more PE coordinates and more layers.

\begin{lemma}[Extra PE coordinates and deeper GIN]\label{app:eapp:lem:extension}
Let $H^{(0)}=X_{\mathrm{PE}}$ with $D_{\mathrm{PE}}\ge K$, and consider an
$L$-layer GIN of the form~\eqref{app:eapp:eq:mpnn-update} with $L\ge1$ and
internal MLPs and intermediate layers of width at least $2K$. On the event of
Lemma~\ref{app:eapp:lem:A3} and for $n$ large, for any target 2-layer ReLU
MLP $\tilde g$ of width $K$ and any $\Ohat\in O(K)$, there exist parameters
such that $H^{(L)}_{i,:}=\tilde g(\PE(i)\Ohat)$ for all $i\in[n]$.
\end{lemma}

\begin{proof}
Every column of $X_{\mathrm{PE}}$ is an eigenvector of $\Aenc$, so
$\Aenc X_{\mathrm{PE}}=X_{\mathrm{PE}}\widehat\Lambda$ with
$\widehat\Lambda:=\diag(\hat\lambda_1,\ldots,\hat\lambda_{D_{\mathrm{PE}}})$.
Let $S\in\Real^{D_{\mathrm{PE}}\times K}$ select the first $K$ coordinates,
so $X_{\mathrm{PE}}S=X=\sqrt n\,\Uhat$. Take $\epsilon^{(l)}=\epsilon$ in every
layer and let $\Gamma:=(1+\epsilon)I_K+\Lhat$, invertible as in
Lemma~\ref{app:eapp:lem:mpnn-realize}.

\emph{$L=1$.} Proceed as in Lemma~\ref{app:eapp:lem:mpnn-realize} with
first-layer weight $\tilde W_1\Ohat^\top\Gamma^{-1}S^\top$. Since
$S^\top\bigl((1+\epsilon)I+\widehat\Lambda\bigr)=\Gamma S^\top$, the first
linear map of the MLP sends the aggregated input of node $i$ to
$\tilde W_1(\PE(i)\Ohat)^\top$, and the extra $D_{\mathrm{PE}}-K$
coordinates receive zero weight.

\emph{$L\ge2$.} Let the first layer output the $2K$ channels
$[\rho(X),\rho(-X)]$, with all other channels zero: take first-layer weight
$[\Gamma^{-1}S^\top;\,-\Gamma^{-1}S^\top]$, zero biases, and an identity
second weight. The ReLU between layers leaves these nonnegative channels
unchanged. Since aggregation is linear in the node dimension and
$\rho(x)-\rho(-x)=x$, applying $[I_K,\,-I_K]$ to the aggregated input of the
next layer gives $(1+\epsilon)X+\Aenc X=X\Gamma$. An intermediate layer
therefore uses first weight $[\Gamma^{-1};\,-\Gamma^{-1}][I_K,\,-I_K]$ and
again outputs $[\rho(X),\rho(-X)]$. The last layer uses first weight
$\tilde W_1\Ohat^\top\Gamma^{-1}[I_K,\,-I_K]$ and the remaining parameters
of $\tilde g$, which gives $H^{(L)}_{i,:}=\tilde g(\PE(i)\Ohat)$. The
construction uses widths at most $2K$; any extra units receive zero weights.
\end{proof}

\paragraph{Lipschitz target and Step C conclusion.}
To deploy Lemmas~\ref{app:eapp:lem:mpnn-realize}
and~\ref{app:eapp:lem:extension} in the Master Lemma, we choose $\tilde g$
to interpolate prototypes to embeddings.

\begin{lemma}[Step C: target Lipschitz interpolation]\label{app:eapp:lem:step-c}
Under Lemma~\ref{app:eapp:lem:step-b} (separation $\dPE > 0$), for any prescribed
embedding targets $\{u_a^\star\}_{a\in[K]}\subset\Real^d$ with $\|u_a^\star\|\leq R$,
there exists a 2-layer ReLU MLP $\tilde g:\Real^K\to\Real^d$ of width $O(K)$
satisfying:
\begin{enumerate}[label=(\roman*),nosep]
\item $\tilde g(\mu_a^\star) = u_a^\star$ for all $a\in[K]$.
\item $\tilde g$ is $L_g$-Lipschitz on $\bigcup_a B_{r_n}(\mu_a^\star)$ with $L_g \leq 2R$.
\end{enumerate}
\end{lemma}

\begin{proof}
We adopt column-vector convention for MLP inputs throughout this proof.
In particular, via Step~B (Lemma~\ref{app:eapp:lem:step-b}) we identify
\[
\mu_a^\star \;=\; (Q_{a,:})^\top/\sqrt{\pi_a} \;=\; Q^\top e_a/\sqrt{\pi_a}\;\in\;\Real^K,
\]
where $e_a\in\Real^K$ is the $a$-th standard basis vector and $Q\in O(K)$
is the orthogonal matrix from~\eqref{app:eapp:eq:Qorth}.

\textbf{Construction.}
Set hidden width $m := K$ and choose MLP parameters
\begin{equation}\label{app:eapp:eq:g-params}
\tilde W_1 := Q\in\Real^{K\times K},\qquad
\tilde b_1 := -t\in\Real^K\ \text{with}\ t_a := \tfrac{1}{2\sqrt{\pi_a}},
\end{equation}
\[
\tilde W_2 := \bigl[\,2\sqrt{\pi_1}\,u_1^\star \,\big|\,\cdots\,\big|\,2\sqrt{\pi_K}\,u_K^\star\,\bigr]\;\in\;\Real^{d\times K},\qquad
\tilde b_2 := 0,
\]
so that
\begin{equation}\label{app:eapp:eq:g-formula}
\tilde g(\xi) \;=\; \tilde W_2\,\rho(\tilde W_1\xi + \tilde b_1) + \tilde b_2
\;=\; \sum_{a=1}^K 2\sqrt{\pi_a}\,u_a^\star\,\rho\!\bigl((Q\xi)_a - t_a\bigr),
\end{equation}
where $(Q\xi)_a$ denotes the $a$-th coordinate of $Q\xi$.

\emph{(i) Exact interpolation.}
Since $QQ^\top=I_K$, $Q\mu_a^\star=e_a/\sqrt{\pi_a}$, i.e.\
$(Q\mu_a^\star)_b=\delta_{ab}/\sqrt{\pi_a}$. Fix $a\in[K]$: at the
``on'' neuron $b=a$, the pre-activation is
$1/\sqrt{\pi_a}-t_a=1/(2\sqrt{\pi_a})>0$ so
$\rho=1/(2\sqrt{\pi_a})$; at any ``off'' neuron $b\ne a$, the
pre-activation is $0-t_b<0$ so $\rho=0$. Plugging
into~\eqref{app:eapp:eq:g-formula},
$\tilde g(\mu_a^\star)=2\sqrt{\pi_a}u_a^\star\cdot\tfrac{1}{2\sqrt{\pi_a}}=u_a^\star$.

\emph{(ii) Lipschitz constant.}
As a composition of the orthogonal $Q$ ($\|Q\|_{\op}=1$),
1-Lipschitz ReLU, and the linear map $\tilde W_2$,
$L_g\le\|\tilde W_2\|_{\op}$. For any $v\in\Real^K$, by triangle
inequality, $\|u_a^\star\|\le R$, and Cauchy--Schwarz with $\sum_a\pi_a=1$,
\[
\|\tilde W_2\,v\|=\Bigl\|\sum_a 2\sqrt{\pi_a}\,v_a u_a^\star\Bigr\|
\le 2R\sum_a\sqrt{\pi_a}\,|v_a|
\le 2R\sqrt{\sum_a\pi_a}\,\|v\|=2R\|v\|.
\]
Hence $L_g\le 2R$. The hidden width $m=K$ matches the $O(K)$ requirement of Lemma~\ref{app:eapp:lem:mpnn-realize}.
\end{proof}

\paragraph{Combined Step A + Step C bound.}
Fix the event of Lemma~\ref{app:eapp:lem:step-a}, take $\Ohat$ from Step~A,
and take $\tilde g$ from Lemma~\ref{app:eapp:lem:step-c}.
Lemma~\ref{app:eapp:lem:mpnn-realize} (for $L=1$ and $D_{\mathrm{PE}}=K$) or
Lemma~\ref{app:eapp:lem:extension} (in general) gives GIN parameters whose
pre-projection output is $H^{(L)}_{i,:}=\tilde g(\PE(i)\Ohat)$. The readout
used in the Master Lemma is therefore
\[
g:=\Pi_{B_R(0)}\circ\tilde g.
\]
Since $\Pi_{B_R(0)}$ is $1$-Lipschitz and fixes every target prototype
$u_a^\star\in B_R(0)$, we have $g(\mu_a^\star)=u_a^\star$,
$\operatorname{Lip}(g)\le\operatorname{Lip}(\tilde g)\le2R$, and
$z_i=g(\PE(i)\Ohat)$ satisfies
$\|z_i-u_{c_i}^\star\|\le L_g\|\PE(i)\Ohat-\mu_{c_i}^\star\|\le L_g r_n$.
This feeds the Master Lemma with $\phi_i:=\PE(i)\Ohat$, prototypes
$\mu_a^\star$, readout $g$, and concentration radius $r_n$.

\subsection{Step D: Decoder Realizability}\label{app:eapp:step-d}

\begin{lemma}[Step D: bilinear decoder realization]\label{app:eapp:lem:step-d}
Under \textbf{(A5)}, there exist target embeddings
$\{u_a^\star\}_{a\in[K]}\subset\Real^d$ (with $d\geq K$) satisfying
$\|u_a^\star\|\leq R$, symmetric $W^\star\in\Real^{d\times d}$ with
$\|W^\star\|_{\op}\leq\Wmax$, and $b^\star\in\Real$ with
$|b^\star|\leq B_0$, such that
\[
b^\star + (u_a^\star)^\top W^\star u_b^\star \;=\; m^\star_{ab}\qquad\forall\,a,b\in[K].
\]
\end{lemma}

\begin{proof}
Let
\[
b^\star \;\in\; \arg\min_{|b|\le B_0}\|M^\star - b\,\mathbf 1\mathbf 1^\top\|_{\op},
\qquad
\tilde M \;:=\; M^\star - b^\star\,\mathbf 1\mathbf 1^\top\in\Real^{K\times K}.
\]
The argmin exists since the objective is continuous and the feasible set $[-B_0,B_0]$ is compact; by construction $|b^\star|\le B_0$. Since $\tilde M$ is symmetric, spectrally decompose
\[
\tilde M \;=\; \sum_{k=1}^K \nu_k\,\psi_k \psi_k^\top,
\qquad \{\psi_k\}_{k=1}^K\subset\Real^K\text{ orthonormal},\ \nu_k\in\Real.
\]
Stack the eigenvectors into the orthogonal matrix
$\Psi:=[\psi_1,\ldots,\psi_K]\in\Real^{K\times K}$ and embed into $\Real^d$ by zero-padding:
\[
\tilde u_a := \bigl((\Psi_{a,:})^\top,\,0,\ldots,0\bigr)^\top\in\Real^d,
\qquad
\tilde W := \diag(\nu_1,\ldots,\nu_K,\underbrace{0,\ldots,0}_{d-K})\in\Real^{d\times d}.
\]
Then
\[
(\tilde u_a)^\top\tilde W\tilde u_b
=\sum_{k=1}^K\nu_k\Psi_{a,k}\Psi_{b,k}
=\tilde M_{ab},
\]
so $b^\star+(\tilde u_a)^\top\tilde W\tilde u_b=m^\star_{ab}$. Moreover $\|\tilde u_a\|=\|\Psi_{a,:}\|=1$ because $\Psi$ is orthogonal, and $\|\tilde W\|_{\op}=\|\tilde M\|_{\op}$.

Set
\[
u_a^\star:=R\tilde u_a,
\qquad
W^\star:=R^{-2}\tilde W.
\]
Then $\|u_a^\star\|=R$ and the bilinear identity is unchanged. By \textbf{(A5)},
\[
\|W^\star\|_{\op}=R^{-2}\|\tilde M\|_{\op}
=R^{-2}\min_{|b|\le B_0}\|M^\star-b\mathbf 1\mathbf 1^\top\|_{\op}
\le \Wmax.
\]
Thus $(u_a^\star,W^\star,b^\star)$ is admissible for the bounded encoder--decoder class and exactly realizes $M^\star$ at the community prototypes.
\end{proof}

\subsection{Proof of Proposition~\ref{prop:eapp-rate}}\label{app:eapp:synthesis}

\begin{proof}[Proof of Proposition~\ref{prop:eapp-rate}]
Combine the four steps via the Master Lemma. Step~D (Lemma~\ref{app:eapp:lem:step-d}) supplies $\{u_a^\star\},W^\star,b^\star$ with $\|u_a^\star\|\le R$, $\|W^\star\|_{\op}\le\Wmax$, and $|b^\star|\le B_0$. Step~A (Lemma~\ref{app:eapp:lem:step-a}) gives, with probability at least $1-n^{-c_A}$, an orthogonal matrix $\Ohat\in O(K)$ satisfying
\[
\max_i\|\PE(i)\Ohat-\mu_{c_i}^\star\|\le r_n=C_A\sqrt{K\log n/n}.
\]
Step~B (Lemma~\ref{app:eapp:lem:step-b}) gives the population prototypes $\mu_a^\star$ and their separation. Step~C combines Lemmas~\ref{app:eapp:lem:mpnn-realize}, \ref{app:eapp:lem:extension}, and~\ref{app:eapp:lem:step-c}: for the realized $\Aenc$, the GIN parameters can absorb $\Ohat$ and $\Lhat$ and implement the readout
\[
g:=\Pi_{B_R(0)}\circ\tilde g,
\qquad
 g(\mu_a^\star)=u_a^\star,
\qquad
\operatorname{Lip}(g)\le L_g\le 2R.
\]
The resulting embeddings satisfy $\|z_i\|\le R$, and $(W^\star,b^\star)$ satisfies the constraints of Assumption~\ref{ass:bounded}, so the score matrix lies in $\Scal(A,\Tcal)$. This is a pathwise witness for the in-class infimum defining $\eapp$; no optimization claim is made.

Apply Lemma~\ref{app:eapp:lem:master} with $\phi_i:=\PE(i)\Ohat$, prototypes $\mu_a^\star$, readout $g$, decoder $(W^\star,b^\star)$, and $R_u=R$. On the Step-A event,
\[
\eapp\le\tfrac{1}{8}\Wmax^2\bigl(2 R L_g r_n+L_g^2 r_n^2\bigr)^2.
\]
Since $L_g\le2R$ and $r_n=o(1)$,
\[
\eapp
\le C\Wmax^2R^4 r_n^2
\le C'\Wmax^2R^4\,\frac{K\log n}{n}
=\frac{C''\log n}{n},
\]
where the final constant absorbs fixed $K$ and depends only on $(K,p_{\min},\pi_{\min},c_\gamma,\tau_0,\Wmax,R)$.
\end{proof}

\subsection{Scope of the Construction}\label{app:eapp:discussion}

Proposition~\ref{prop:eapp-rate} is an existence statement for one
subclass: GIN with spectral PE and a general bilinear decoder. By
Lemma~\ref{app:eapp:lem:extension}, this subclass includes the 2-layer GIN
with $D_{\mathrm{PE}}\!=\!8$ and hidden width 64 used in
Sec.~\ref{sec:experiments}, for every $K\le4$. Because $\eapp$ is an
in-class infimum, the same bound transfers to any larger encoder--decoder
family that contains this construction.

The proof uses only that the aggregation operator is fixed and diagonalized
by the PE, so that aggregation acts on the PE as a diagonal scaling
(Lemma~\ref{app:eapp:lem:mpnn-spectral}). Attention-based aggregators such
as GAT are input-dependent. Normalized aggregators such as GCN and GraphSAGE
are fixed, but in general they are not diagonalized by the spectral PE of
$\Aenc$, nor by the Laplacian PE that our GCN and GAT runs use
(App.~\ref{app:hyperparams}). These configurations require separate
compatibility arguments (App.~\ref{app:K4-collapse}). The dense-regime
assumption is also essential: in sparse SBMs the $2\to\infty$ concentration
in Lemma~\ref{app:eapp:lem:A5} can be too weak for $r_n\to0$ without a
different spectral analysis.
\section{Experimental Setup, Full Sweeps, and Additional Architectures}
\label{app:setup-and-K}

This appendix records the full experimental setup, the seed coverage of each
cell, the $K\!\in\!\{2,4\}$ sweeps for the architectures that recover the
communities, and the Step-1 ratio across all sweeps. The $K\!=\!4$ failures
of GCN+PE and GAT+PE are discussed in App.~\ref{app:failure-modes}.

\subsection{Hyperparameters and Training}\label{app:hyperparams}

\paragraph{Encoders.}
GIN \citep{xu2019gin}: 2 layers, hidden width 64, learnable $\epsilon$
initialized at $0$.
GCN \citep{kipf2017gcn}: 2 layers, hidden width 64, symmetric normalization
with self-loops.
GAT \citep{velickovic2018gat}: 2 layers, 4 attention heads of width 16,
concatenated in the hidden layer and averaged in the output layer.
Centered-power encoder (CP), a power-iteration encoder inspired by
ChebNet~\citep{defferrard2016chebnet}: 10 propagation steps
$H^{(\ell+1)}\!=\!\mathrm{CS}(A^{\mathrm{enc}}H^{(\ell)})$ with
$H^{(0)}\!=\!X$, where $\mathrm{CS}$ centers each column and scales it to
unit variance. The outputs of all 10 steps are concatenated, passed through
layer normalization, and mapped to $\Real^d$ by a 2-layer ReLU MLP of hidden width 64.
The propagation steps have no trainable parameters.
All encoders output embeddings of dimension $d\!=\!16$.

\paragraph{Features.}
GIN+PE uses the spectral PE of Prop.~\ref{prop:eapp-rate}: the
$D_{\mathrm{PE}}\!=\!8$ eigenvectors of $A^{\mathrm{enc}}$ with the largest
$|\lambda|$, rescaled by $\sqrt n$. This setting is covered by
Prop.~\ref{prop:eapp-rate} (Lemma~\ref{app:eapp:lem:extension}).
GCN+PE and GAT+PE use a Laplacian PE: the eigenvectors of
$I-D^{-1/2}A^{\mathrm{enc}}D^{-1/2}$, with $D$ the degree matrix of
$A^{\mathrm{enc}}$, for the 2nd to 9th smallest eigenvalues (the trivial
eigenvector is skipped), each column rescaled to unit standard deviation.
The two PEs differ, so ``+PE'' does not denote the same input across
architectures.
GIN+random and CP+random use i.i.d.\ $\mathcal{N}(0,1)$ features in $16$
dimensions, drawn once per seed.
All graph-derived features are computed from $A^{\mathrm{enc}}$.

\paragraph{Decoder and constraints.}
Symmetric bilinear decoder $s_{ij}\!=\!b\!+\!z_i^\top W z_j$. All
constraints are enforced in the forward pass, so every evaluated model
satisfies Assumption~\ref{ass:bounded}. Embeddings are projected onto the
$\ell_2$-ball of radius $R\!=\!1.5$; $W$ is computed from an unconstrained
matrix $M$ by symmetrizing it and clipping its eigenvalues to
$[-\Wmax,\Wmax]$; and $b\!=\!B_0\tanh(b_{\mathrm{raw}})$. The bias is
initialized at zero, and $M$ at
$0.8\min(\Wmax,1)\,I+0.01\cdot\mathrm{randn}(d,d)$.

\paragraph{Optimizer.}
Adam, learning rate $10^{-3}$, weight decay $10^{-4}$, full-batch BCE on
$\Tcal$ for 2500 epochs (3500 for CP+random), with no early stopping.
GIN+random is trained from 3 random restarts, and we keep the restart with
the lowest final training loss; all other configurations use a single run.
Held-out pairs are never used for stopping or selection.

\paragraph{Norm-rule details.}
For the $n$, $K$, $\Wmax$, and $\rho_n$ sweeps, capacity is set relative to
the SBM logit gap: $\Wmax\!=\!(\mathrm{cap\text{-}ratio})\cdot\mathrm{cap}/R^2$
with $\mathrm{cap}\!=\!|\logit p_{\mathrm{in}}-\logit p_{\mathrm{out}}|$ and
default cap-ratio $2$. For the symmetric SBMs used here,
$\min_b\|M^\star-b\,\mathbf 1\mathbf 1^\top\|_{\op}=\mathrm{cap}$, so
cap-ratio $1$ is exactly the bilinear capacity condition \textbf{(A5)} of
App.~\ref{app:eapp}, which is sufficient. Realizability needs at least cap-ratio
$(K\!-\!1)/K$ (Prop.~\ref{prop:exact-capacity}). These sweeps also set $B_0\!=\!|b^\star|+0.5$, where
$b^\star$ minimizes $\|M^\star-b\,\mathbf 1\mathbf 1^\top\|_{\op}$
(Lemma~\ref{app:eapp:lem:step-d}). Both $\mathrm{cap}$ and $b^\star$ are
recomputed for each SBM cell, so the $\rho_n$ sweep follows the changing
edge density. The $B_0$ sweep sets $B_0$ directly and keeps the default
$\Wmax$. The $\Deltas$ sweep fixes $\Wmax\!=\!1.5$ and $B_0\!=\!3.5$,
keeping capacity fixed while $\Deltas$ changes through $p_{\mathrm{in}}$.

\paragraph{Statistics and reproducibility.}
Each cell uses 10 random seeds, numbered 0--9. Each seed determines the
graph, the mask, the initialization, the labeled sets, and the random
features. We report medians and interquartile ranges (IQRs) over seeds
throughout.

\subsection{Cell Coverage}\label{app:coverage}

\begin{table}[h]
\centering\small
\caption{$n$-sweep cell coverage (number of completed seeds, max $10$).}
\label{tab:coverage}
\begin{tabular}{l|cccccccc}
\toprule
$K=2$ arch & $n=100$ & 200 & 300 & 400 & 500 & 1000 & 1500 & 2000 \\
\midrule
GIN + random   & 10 & 9  & 8  & 9  & 10 & 10 & 10 & 10 \\
GIN + PE       & 10 & 10 & 9  & 10 & 10 & 10 & 10 & 10 \\
CP + random    & 10 & 10 & 10 & 10 & 10 & 10 & 10 & 10 \\
GCN + PE       & 10 & 10 & 10 & 10 & 10 & 10 & 10 & 10 \\
GAT + PE       & 10 & 10 & 10 & 10 & 10 & 10 & 10 & 10 \\
\midrule
$K=3$ arch & & & & & & & & \\
\midrule
GIN + random   & 10 & 10 & 9  & 7  & 10 & 10 & 10 & 10 \\
GIN + PE       & 10 & 10 & 10 & 10 & 10 & 10 & 10 & 10 \\
CP + random    & 10 & 10 & 10 & 10 & 10 & 10 & 10 & 10 \\
GCN + PE       & 10 & 10 & 10 & 10 & 10 & 10 & 10 & 10 \\
GAT + PE       & 10 & 10 & 10 & 10 & 10 & 10 & 10 & 10 \\
\midrule
$K=4$ arch & & & & & & & & \\
\midrule
GIN + random   & 10 & 10 & 9  & 10 & 6  & 10 & 10 & 10 \\
GIN + PE       & 10 & 10 & 10 & 10 & 10 & 10 & 10 & 10 \\
CP + random    & 10 & 10 & 10 & 10 & 10 & 10 & 10 & 10 \\
GCN + PE       & 10 & 10 & 10 & 10 & 10 & 10 & 10 & 10 \\
GAT + PE       & 10 & 10 & 10 & 10 & 10 & 10 & 9  & 10 \\
\bottomrule
\end{tabular}
\end{table}

The $n$-sweep completed $1{,}185$ of its $1{,}200$ runs, and $111$ of its
$120$ cells have all 10 seeds (Table~\ref{tab:coverage}). Across all five
sweeps, $2{,}329$ of $2{,}350$ runs completed. The other $21$ runs stopped
with a runtime error during training and produced no final model: $15$ in
the $n$-sweep ($13$ of them GIN+random) and $6$ in the $\rho_n$ sweep ($5$
at $\rho_n\!=\!0.05$, $n\!=\!1000$, and $1$ at $\rho_n\!=\!0.2$,
$n\!=\!500$). Every completed run is included in the reported statistics,
including runs whose decoder collapses to $\widehat W\!\approx\!\mathbf 0$
(App.~\ref{app:K4-collapse}). App.~\ref{app:recipe} discusses the cause of the errors.

\subsection{$K=2$ and $K=4$ Sweeps for the Recovering Architectures}
\label{app:KsuccessFull}

\begin{figure}[h]
\centering
\includegraphics[width=\linewidth,trim=0 0 0 36bp,clip]{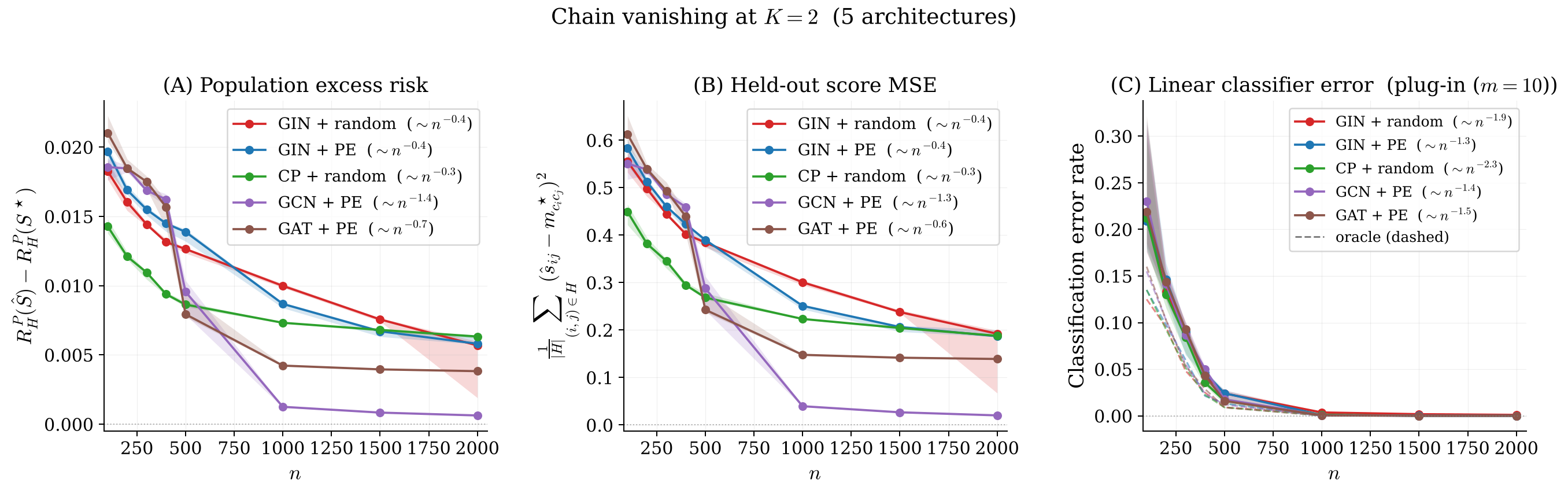}
\caption{\textbf{Theorem quantities at $K\!=\!2$ (all five architectures,
including GAT+PE).} The quantities of Figure~\ref{fig:chain} decrease with
$n$ for all five architectures. By $n\!=\!1000$ every architecture,
including the Laplacian-PE configurations GCN+PE and GAT+PE and the
random-feature configurations, reaches a median plug-in error of at most
$0.004$. Slopes are noisier than at $K\!=\!3$ because the absolute
residuals are smaller.}
\label{fig:K2-chain}
\end{figure}

\begin{figure}[h]
\centering
\includegraphics[width=\linewidth,trim=0 0 0 36bp,clip]{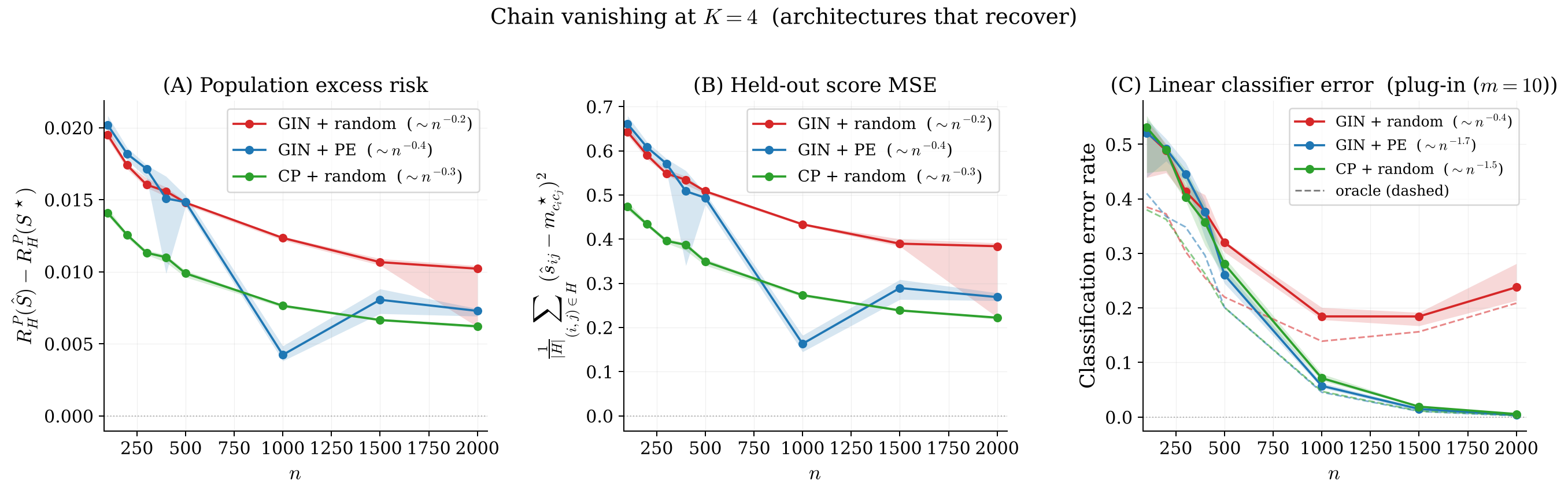}
\caption{\textbf{Theorem quantities at $K\!=\!4$, recovering
architectures.} GIN+PE and CP+random recover: their median plug-in errors
fall to $0.004$ and $0.006$ at $n\!=\!2000$. GIN+random recovers only
partially (error $0.24$). The LP quantities need not decrease
monotonically: the excess of GIN+PE falls to $0.004$ at $n\!=\!1000$ and
returns to about $0.007$ for $n\!\ge\!1500$, while that of CP+random
decreases to $0.006$. At $n\!=\!2000$ the excess of each architecture is
$1.0\times$ to $1.9\times$ its value at $K\!=\!3$, where the separation is
also larger ($\Deltas\!=\!0.80$ vs $0.69$ at $K\!=\!4$). The GCN+PE and
GAT+PE failures at $K\!=\!4$ are discussed in App.~\ref{app:failure-modes}.} 
\label{fig:K4-chain-success}
\end{figure}

At $K\!=\!2$ (Fig.~\ref{fig:K2-chain}), all architectures reach a median error of at most $0.001$ at
$n\!=\!2000$. GAT+PE matches GCN+PE and GIN+PE within the IQR, and the
symmetric $K\!=\!2$ SBM is easy enough that even the random-feature
architectures recover.

At $K\!=\!4$ (Fig.~\ref{fig:K4-chain-success}), the recovering architectures are GIN+PE (excess $0.007$,
error $0.004$ at $n\!=\!2000$), CP+random (excess $0.006$, error $0.006$),
and GIN+random (excess $0.010$, error $0.24$, partial recovery). GCN+PE and
GAT+PE collapse for $n\!\ge\!1500$; App.~\ref{app:failure-modes} discusses
these cases.

\subsection{Cross-Ablation Step-1 Ratio}\label{app:ratio-table}

\begin{table}[h]
\centering\small
\caption{Empirical Step-1 ratio
$\mathrm{ScoreMSE}_\Hcal\,/\,[\Rpop(\widehat S)\!-\!\Rpop(\Sopt)]$ at the
baseline cell of each sweep (GIN+PE, $K\!=\!3$): median and IQR over seeds
of the per-seed ratio. The typical-curvature proxy is
$2/\mu_{\mathrm{eff}}\!\approx\!36$ (App.~\ref{app:mu-eff}).}
\label{tab:ratio}
\begin{tabular}{lccc}
\toprule
Sweep & median ratio & IQR & \# seeds \\
\midrule
$n$-sweep ($n\!=\!2000$)                                   & $39.0$ & $[35.6, 41.6]$ & $10$ \\
$\Deltas$-sweep ($p_{\mathrm{in}}\!=\!0.10$, $n\!=\!1000$) & $36.6$ & $[35.1, 37.5]$ & $10$ \\
cap-ratio sweep (cap-ratio$\!=\!2.0$, $n\!=\!2000$)        & $39.0$ & $[36.6, 41.1]$ & $10$ \\
$B_0$ sweep ($B_0\!=\!3.4$, $n\!=\!2000$)                  & $37.7$ & $[36.2, 40.9]$ & $10$ \\
$\rho_n$ sweep ($\rho_n\!=\!1.0$, $n\!=\!2000$)            & $40.2$ & $[35.8, 41.7]$ & $10$ \\
\midrule
mean across sweeps & \multicolumn{3}{c}{$38.5$ (spread $\approx\!10\%$)} \\
worst-case $2/\muB$ & \multicolumn{3}{c}{$425$--$1940$ at these cells; about $80$ to $2\!\times\!10^6$ over all cells} \\
\bottomrule
\end{tabular}
\end{table}
\section{Empirical Checks of the Proof Steps}\label{app:step-checks}

This appendix checks the steps of Theorem~\ref{thm:main} on the trained
models: the Step-1 curvature constant (App.~\ref{app:mu-eff}), Steps~1 and~2
across $n$ (App.~\ref{app:ratio-by-n}), Part~II through oracle, plug-in, and
direct empirical risk minimization (ERM) classifiers
(App.~\ref{app:three-classifier}), and what LP training adds to its input
(App.~\ref{app:pe-baseline}).

\subsection{Effective vs.\ Worst-Case Hessian}\label{app:mu-eff}

For a pair $(i,j)$ we write $m^\star_{ij}:=m^\star_{c_ic_j}$.

\begin{lemma}[Segment form of the BCE excess]\label{lem:segment-step1}
For any score matrix $S\!\in\!\Scal(A,\Tcal)$,
\begin{equation}\label{eq:segment-form}
\Rpop(S) - \Rpop(\Sopt)
\;=\;
\frac{1}{|\Hcal|}\sum_{(i,j)\in\Hcal} \widetilde\sigma_{ij}\,(s_{ij}-m^\star_{ij})^2,
\end{equation}
where
$\widetilde\sigma_{ij}\!:=\!\int_0^1\!(1-t)\,\sigma'\!\bigl(m^\star_{ij}+t(s_{ij}-m^\star_{ij})\bigr)\,dt$.
\end{lemma}

\begin{proof}
Per pair, $\partial_s^2\,\ell(s;p)\!=\!\sigma'(s)$ is independent of the
probability target $p$, and the linear Taylor term
$\partial_s\ell(m^\star;p)\,(s-m^\star)\!=\!(\sigma(m^\star)\!-\!p)(s-m^\star)$
vanishes because $m^\star\!=\!\logit p$ gives $\sigma(m^\star)\!=\!p$.
The Taylor remainder then gives
\[
\ell(s;p)-\ell(m^\star;p) \;=\; \biggl[\int_0^1 (1-t)\,\sigma'(m^\star+t(s-m^\star))\,dt\biggr]\,(s-m^\star)^2,
\]
and averaging over $\Hcal$ yields \eqref{eq:segment-form}.
\end{proof}

\begin{remark}[$\mu_{\mathrm{eff}}$ as a typical-scale proxy]
Lemma~\ref{lem:segment-step1} gives the exact identity
\[
\mathrm{ScoreMSE}_\Hcal \;=\; \frac{2\,[\Rpop(\widehat S)\!-\!\Rpop(\Sopt)]}{\mu_{\mathrm{err}}(\widehat S)},
\qquad
\mu_{\mathrm{err}}(\widehat S) \;:=\; \frac{2\sum_\Hcal \widetilde\sigma_{ij}\,e_{ij}^2}{\sum_\Hcal e_{ij}^2},
\]
where $e_{ij}\!:=\!s_{ij}\!-\!m^\star_{ij}$. Thus the exact constant is an
error-weighted average of the BCE curvature along the Taylor segments. As
$\widehat S\!\to\!\Sopt$, $\widetilde\sigma_{ij}\!\to\!\tfrac12\sigma'(m^\star_{ij})\!=\!\tfrac12 P_{ij}(1\!-\!P_{ij})$ pointwise, so
\[
\mu_{\mathrm{err}}(\widehat S) \;\longrightarrow\; \frac{\sum_\Hcal P_{ij}(1\!-\!P_{ij})\,e_{ij}^2}{\sum_\Hcal e_{ij}^2}.
\]
This limit is the error-weighted average of $P_{ij}(1\!-\!P_{ij})$. It equals
the simple average
\[
\mu_{\mathrm{eff}} \;:=\; \E_{(i,j)\sim\Hcal}[P_{ij}(1\!-\!P_{ij})]
\]
only when the squared errors are not systematically aligned with block-pair
curvature.

We use $\mu_{\mathrm{eff}}$ as a \emph{typical-scale proxy} because it depends
only on the SBM probabilities $\{p_{ab}\}$, not on the trained model, and it
tracks $\mu_{\mathrm{err}}$ closely in our experiments
(Tables~\ref{tab:ratio},\ref{tab:ratio-by-n}: $2/\mu_{\mathrm{eff}}\!\approx\!36$,
empirical ratio $38.5$). The small gap is consistent with errors putting
slightly more mass on lower-curvature block pairs. For example,
off-diagonal pairs at $p_{\mathrm{out}}\!=\!0.04$ have
$p(1\!-\!p)\!=\!0.038$, compared with $0.090$ on the diagonal at
$p_{\mathrm{in}}\!=\!0.10$.
\end{remark}

The bound $\muB\!=\!\sigma(B)(1\!-\!\sigma(B))$ used in
Lemma~\ref{lem:step1} is the worst-case lower bound of $\sigma'$ over the
\emph{entire} score range $[-B,B]$, where $B\!=\!R^2 W_{\max}\!+\!B_0$.
This can be very conservative. In dense regimes with
$p_{\min}\!\le\!p_{ab}\!\le\!1\!-\!p_{\min}$,
$\mu_{\mathrm{eff}}\!\ge\!p_{\min}(1\!-\!p_{\min})$, but $\muB$ can still go
to zero as $B\!\to\!\infty$.

\paragraph{Empirical implication.}
$\muB$ is still the right constant for a uniform upper bound over all
admissible scores in $[-B,B]$, so Theorem~\ref{thm:main} keeps it.
One cannot simply replace $\muB^{-1}$ by $\mu_{\mathrm{eff}}^{-1}$ in the
formal statement, because the limit
$\mu_{\mathrm{err}}\!\to\!\mu_{\mathrm{eff}}$ requires errors to be balanced
across block pairs. Empirically, however, $\mu_{\mathrm{err}}$ tracks
$\mu_{\mathrm{eff}}$. Across the $\Wmax$ and $B_0$ ablations ($n\!=\!2000$),
$2/\muB$ ranges from $80$ to $2.3\!\times\!10^6$, while the median ratio stays
within $36.5$--$42.2$. For $\rho_n\!\in\![0.1,2]$, the ratio stays
$1$--$16\%$ above $2/\mu_{\mathrm{eff}}\!\in\![19.5,336]$. At
$\rho_n\!=\!0.05$, where $\mathrm{ScoreMSE}_\Hcal\!\approx\!3.7$, it falls below ($375$ vs.\ $669$). A rigorous replacement
of $\muB$ would require controlling
$\mu_{\mathrm{err}}$ over the encoder--decoder class; we leave this open.

\subsection{Step-1 and Step-2 Checks Across $n$}\label{app:ratio-by-n}

\begin{table}[h]
\centering\small
\caption{Step-1 ratio
$\mathrm{ScoreMSE}_\Hcal\,/\,[\Rpop(\widehat S)\!-\!\Rpop(\Sopt)]$
across all $n$ at the $K\!=\!3$ baseline: median over seeds of the
per-seed ratio.}
\label{tab:ratio-by-n}
\begin{tabular}{l|cccccccc}
\toprule
arch / $n$ & 100 & 200 & 300 & 400 & 500 & 1000 & 1500 & 2000 \\
\midrule
GIN+PE       & $32.0$ & $32.6$ & $32.8$ & $32.5$ & $32.5$ & $37.3$ & $36.6$ & $39.0$ \\
GCN+PE       & $31.5$ & $31.7$ & $31.7$ & $31.5$ & $31.5$ & $34.2$ & $34.0$ & $34.1$ \\
GIN+random   & $32.3$ & $33.3$ & $33.3$ & $33.3$ & $33.4$ & $33.6$ & $35.3$ & $36.5$ \\
CP+random    & $32.8$ & $33.7$ & $34.0$ & $34.1$ & $34.3$ & $34.3$ & $34.1$ & $33.8$ \\
\midrule
predicted $2/\mu_{\mathrm{eff}}$ & \multicolumn{8}{c}{$\approx 35.97$} \\
\bottomrule
\end{tabular}
\end{table}

The empirical ratio rises with $n$, from $31.5$--$34.3$ at $n\!\le\!500$
to $33.8$--$39.0$ at $n\!=\!2000$, around the proxy
$2/\mu_{\mathrm{eff}}\!\approx\!36$. The exact limiting ratio is
$2/\mu_{\mathrm{err}}(\widehat S)$, where $\mu_{\mathrm{err}}$ is an
error-weighted curvature. A plausible reason for the lower values at small
$n$ is that $\widehat S$ has not yet concentrated near $\Sopt$, so the
segment $[m^\star,\widehat s]$ samples a wider range of $\sigma'$.
In our operating region, $p_{\mathrm{in}}\!=\!0.10$ and
$p_{\mathrm{out}}\!=\!0.04$ correspond to $\sigma\!<\!0.21$, where
$\sigma'$ is convex. Jensen's inequality then makes the integral average
$\widetilde\sigma_{ij}$ larger than $\tfrac12\sigma'(m^\star_{ij})$ on average,
which lowers the empirical ratio. We do not formalize this explanation.

\paragraph{Step-2 sanity check: held-out vs full ScoreMSE.}
Step~2 of Theorem~\ref{thm:main} (Lemma~\ref{lem:step2}) bounds
$\mathrm{ScoreMSE}_{\mathrm{full}}$ by $\mathrm{ScoreMSE}_\Hcal$ plus a
sampling-without-replacement term $\Vstat\!=\!\Otilde(\sqrt{d/n})$.
Table~\ref{tab:scoremse-compare} shows GIN+PE. For all four architectures
at the $K\!=\!3$ baseline, the medians of the two differ by less than $9\%$
at every $n$ and by less than $4.5\%$ for $n\!\ge\!500$.
$\mathrm{ScoreMSE}_\Hcal$ is therefore a reliable proxy for the
unobservable $\mathrm{ScoreMSE}_{\mathrm{full}}$ in the ratio computations.

\begin{table}[h]
\centering\small
\caption{ScoreMSE$_\Hcal$ vs ScoreMSE$_{\mathrm{full}}$ for GIN+PE at the
$K\!=\!3$ baseline (medians over 10 seeds).}
\label{tab:scoremse-compare}
\begin{tabular}{l|ccc}
\toprule
$n$ & ScoreMSE$_\Hcal$ & ScoreMSE$_{\mathrm{full}}$ &
$(\Hcal{-}\mathrm{full})/\mathrm{full}$ \\
\midrule
$100$  & $0.6410$ & $0.6108$ & $+4.95\%$ \\
$200$  & $0.5638$ & $0.5454$ & $+3.37\%$ \\
$500$  & $0.3538$ & $0.3632$ & $-2.60\%$ \\
$1000$ & $0.1254$ & $0.1270$ & $-1.28\%$ \\
$2000$ & $0.1423$ & $0.1438$ & $-1.02\%$ \\
\bottomrule
\end{tabular}
\end{table}

\subsection{Three-Classifier Comparison and Theorem~Part~II}\label{app:three-classifier}

\begin{figure}[h]
\centering
\includegraphics[width=\linewidth]{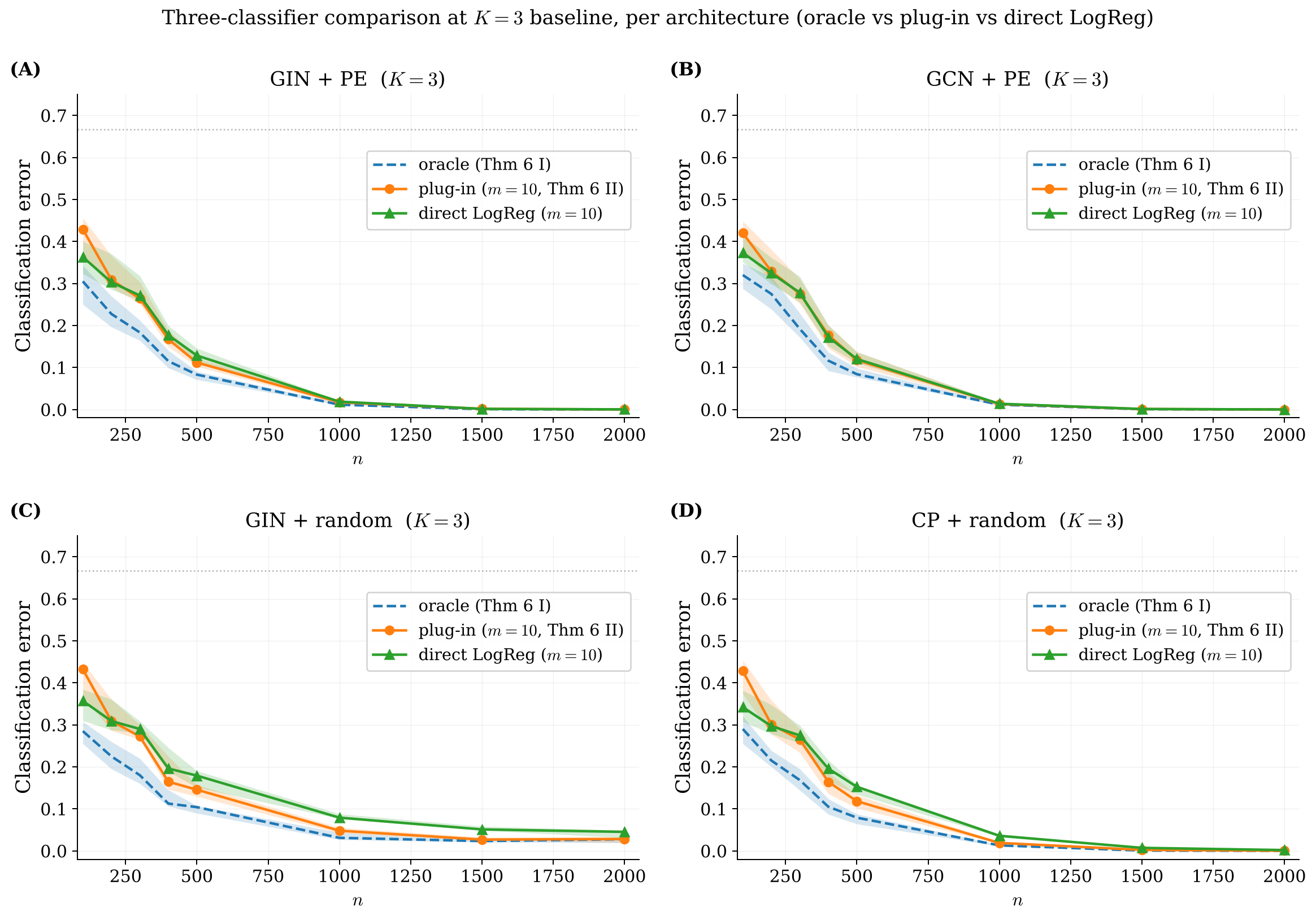}
\caption{\textbf{Three-classifier comparison at $K\!=\!3$, four
architectures.} For each architecture, we plot the oracle classifier
(using ground-truth labels), the plug-in classifier of
Theorem~\ref{thm:main} Part~II (using $m\!=\!10$ labeled prototypes per
class), and a direct ERM classifier (logistic regression trained on the
same $m\!=\!10$ labels per class). Theorem Part~II closes only when
$\En\!\to\!0$ and $m$ grows with $n$, because of the $\log(2Kn)/m$ term in
Lemma~\ref{lem:prototype}. In these runs, fixed $m\!=\!10$ already tracks
the oracle once $n$ is large, so this term is not binding empirically.}
\label{fig:three-classifier}
\end{figure}

Figure~\ref{fig:three-classifier} supports the Part~II prediction that
$(\Deltas\!-\!2\eta_L)^2$ approaches $\Deltas^2$. At $n\!=\!2000$, the
plug-in and oracle classifiers agree within $0.001$ for every
architecture, and direct ERM matches them within $0.002$ except for
GIN+random. At $n\!=\!200$, both plug-in and direct ERM trail the oracle
by $0.05$--$0.09$, and they differ from each other by less than $0.01$.

Table~\ref{tab:three-classifier} gives the same comparison numerically,
sweeping $m\!\in\!\{1,5,10,20\}$ for both plug-in and direct ERM at
$n\!\in\!\{500,1000,2000\}$. Two patterns stand out. First, at
$n\!=\!2000$, for the two PE configurations, all classifiers with
$m\!\geq\!5$, plug-in and direct alike, match the oracle within $0.001$.
Second, at moderate $n$ and small $m$, the plug-in classifier usually
outperforms direct ERM, the exception being GCN+PE at $n\!=\!1000$. For example, at
$n\!=\!1000$ and $m\!=\!1$, plug-in gives $0.055$ versus direct $0.109$ for
GIN+PE, and $0.101$ versus $0.274$ for CP+random. This is consistent with
Theorem Part~II: when $\En$ is already small, the prototype error $\eta_L$
can be smaller than the direct-ERM finite-sample error. Better LP training
then reduces the labels needed through the LP-quality floor
$\sqrt{\En/\pi_{\min}}$ in Lemma~\ref{lem:prototype}.

\begin{table}[h]
\centering\small
\caption{Three-classifier comparison at $K\!=\!3$ baseline:
oracle (using ground-truth labels), plug-in (Theorem Part~II,
$m\!\in\!\{1,5,10,20\}$ labeled prototypes per class), and direct
(logistic regression, LogReg, trained on the same $m$ labels per class), evaluated
on all $n$ nodes. Medians over 10 seeds.}
\label{tab:three-classifier}\setlength{\tabcolsep}{4pt}
\begin{tabular}{ll|c|cccc|cccc}
\toprule
\multirow{2}{*}{arch} & \multirow{2}{*}{$n$} & \multirow{2}{*}{oracle} & \multicolumn{4}{c|}{plug-in (Thm.~\ref{thm:main} II)} & \multicolumn{4}{c}{direct (LogReg)} \\
& & & $m{=}1$ & $m{=}5$ & $m{=}10$ & $m{=}20$ & $m{=}1$ & $m{=}5$ & $m{=}10$ & $m{=}20$ \\
\midrule
GIN+PE     & $500$  & $0.083$ & $0.268$ & $0.140$ & $0.112$ & $0.099$ & $0.361$ & $0.185$ & $0.128$ & $0.101$ \\
           & $1000$ & $0.011$ & $0.055$ & $0.018$ & $0.017$ & $0.015$ & $0.109$ & $0.021$ & $0.019$ & $0.015$ \\
           & $2000$ & $0.000$ & $0.004$ & $0.001$ & $0.000$ & $0.000$ & $0.003$ & $0.000$ & $0.000$ & $0.000$ \\
\midrule
GCN+PE     & $500$  & $0.084$ & $0.300$ & $0.148$ & $0.117$ & $0.107$ & $0.339$ & $0.153$ & $0.120$ & $0.101$ \\
           & $1000$ & $0.011$ & $0.033$ & $0.015$ & $0.013$ & $0.012$ & $0.025$ & $0.014$ & $0.014$ & $0.012$ \\
           & $2000$ & $0.000$ & $0.001$ & $0.000$ & $0.000$ & $0.000$ & $0.001$ & $0.000$ & $0.000$ & $0.000$ \\
\midrule
GIN+random & $500$  & $0.104$ & $0.342$ & $0.188$ & $0.146$ & $0.123$ & $0.425$ & $0.241$ & $0.179$ & $0.135$ \\
           & $1000$ & $0.031$ & $0.163$ & $0.059$ & $0.048$ & $0.038$ & $0.328$ & $0.127$ & $0.079$ & $0.052$ \\
           & $2000$ & $0.028$ & $0.062$ & $0.032$ & $0.028$ & $0.028$ & $0.192$ & $0.066$ & $0.045$ & $0.039$ \\
\midrule
CP+random  & $500$  & $0.079$ & $0.320$ & $0.159$ & $0.118$ & $0.096$ & $0.413$ & $0.208$ & $0.152$ & $0.106$ \\
           & $1000$ & $0.013$ & $0.101$ & $0.029$ & $0.019$ & $0.015$ & $0.274$ & $0.069$ & $0.036$ & $0.020$ \\
           & $2000$ & $0.000$ & $0.020$ & $0.001$ & $0.000$ & $0.000$ & $0.188$ & $0.014$ & $0.002$ & $0.001$ \\
\bottomrule
\end{tabular}
\end{table}

\subsection{Raw-PE Baseline}\label{app:pe-baseline}

To measure what LP training adds to its input, we apply the same
logistic-regression probe, with the same labeled sets, to fixed
representations computed from $\Aenc$ on the same graphs ($K\!=\!3$, 10
seeds). These are the $8$-dimensional spectral PE that GIN+PE receives (raw
PE), the $8$-dimensional Laplacian PE of GCN+PE, the spectral PE truncated
to its top $K$ eigenvectors, and adjacency spectral embedding
$U_K|\Lambda_K|^{1/2}$ (ASE$_K$) \citep{sussman2012consistent}.
Table~\ref{tab:pe-baseline} compares them with the GIN+PE embeddings
(LP+LogReg) and with the plug-in classifier of Theorem~\ref{thm:main}
Part~II (LP+plug-in). LP training lowers the median error relative to the
raw PE at every $n\!\in\!\{500,1000,2000\}$ and $m\!\in\!\{1,5,10,20\}$,
and in $112$ of the $120$ seed-level comparisons. The $K$-dimensional
spectral representations are stronger than LP at $n\!=\!1000$, $m\!=\!1$
($0.033$ for ASE$_K$ vs.\ $0.109$ for LP+LogReg) and on par at $n\!=\!2000$. We therefore claim only that LP training
improves on its own input, not that it outperforms spectral methods
designed for SBMs.

\begin{table}[h]
\centering\footnotesize
\caption{Raw-PE baseline at $K\!=\!3$: classification error, median with
IQR (linear interpolation) over 10 seeds, $m$ labeled nodes per class. Raw and Lap PE have $8$ dimensions, spec PE$_K$ and ASE$_K$ have $K$.
All representations except the last two columns are computed from $\Aenc$
without training.}
\label{tab:pe-baseline}
\setlength{\tabcolsep}{4pt}
\begin{tabular}{cc|cccc|cc}
\toprule
$n$ & $m$ & raw PE & Lap PE & spec PE$_K$ & ASE$_K$ & LP+LogReg & LP+plug-in \\
\midrule
500 & 1 & $0.448$ & $0.443$ & $0.207$ & $0.216$ & $0.361$ & $0.268$ \\
\multicolumn{2}{c|}{} & {\scriptsize$[0.418,0.457]$} & {\scriptsize$[0.422,0.465]$} & {\scriptsize$[0.164,0.243]$} & {\scriptsize$[0.181,0.233]$} & {\scriptsize$[0.328,0.379]$} & {\scriptsize$[0.247,0.309]$} \\[2pt]
500 & 10 & $0.147$ & $0.143$ & $0.104$ & $0.101$ & $0.128$ & $0.112$ \\
\multicolumn{2}{c|}{} & {\scriptsize$[0.139,0.163]$} & {\scriptsize$[0.133,0.157]$} & {\scriptsize$[0.090,0.114]$} & {\scriptsize$[0.088,0.108]$} & {\scriptsize$[0.115,0.146]$} & {\scriptsize$[0.103,0.122]$} \\[2pt]
1000 & 1 & $0.392$ & $0.381$ & $0.045$ & $0.033$ & $0.109$ & $0.055$ \\
\multicolumn{2}{c|}{} & {\scriptsize$[0.369,0.410]$} & {\scriptsize$[0.374,0.393]$} & {\scriptsize$[0.029,0.051]$} & {\scriptsize$[0.026,0.043]$} & {\scriptsize$[0.092,0.130]$} & {\scriptsize$[0.041,0.081]$} \\[2pt]
1000 & 10 & $0.035$ & $0.041$ & $0.018$ & $0.015$ & $0.019$ & $0.017$ \\
\multicolumn{2}{c|}{} & {\scriptsize$[0.033,0.038]$} & {\scriptsize$[0.036,0.044]$} & {\scriptsize$[0.015,0.019]$} & {\scriptsize$[0.013,0.018]$} & {\scriptsize$[0.016,0.020]$} & {\scriptsize$[0.013,0.021]$} \\[2pt]
2000 & 1 & $0.324$ & $0.345$ & $0.002$ & $0.002$ & $0.003$ & $0.003$ \\
\multicolumn{2}{c|}{} & {\scriptsize$[0.311,0.347]$} & {\scriptsize$[0.330,0.368]$} & {\scriptsize$[0.001,0.003]$} & {\scriptsize$[0.001,0.002]$} & {\scriptsize$[0.001,0.007]$} & {\scriptsize$[0.001,0.015]$} \\[2pt]
2000 & 10 & $0.003$ & $0.004$ & $0.000$ & $0.000$ & $0.000$ & $0.000$ \\
\multicolumn{2}{c|}{} & {\scriptsize$[0.003,0.004]$} & {\scriptsize$[0.003,0.005]$} & {\scriptsize$[0.000,0.000]$} & {\scriptsize$[0.000,0.000]$} & {\scriptsize$[0.000,0.001]$} & {\scriptsize$[0.000,0.000]$} \\
\bottomrule
\end{tabular}
\end{table}
\section{Failure Modes and Limitations}\label{app:failure-modes}

This appendix collects four regimes in which Theorem~\ref{thm:main} still
holds, since it is conditional on the LP risk that training attains, but
either training fails or the bound becomes uninformative:
(i) the collapse of GCN+PE and GAT+PE at $K\!=\!4$,
(ii) the looseness of the Markov factor $\Deltas^{-2}$ in Step~3 deep
inside the recovery regime,
(iii) the plateau of the $\Wmax$ ablation below the capacity threshold,
and (iv) the sparse-$\rho_n$ regime.
All results use the training recipe of App.~\ref{app:hyperparams}.
App.~\ref{app:recipe} reruns the failing cells with a second recipe, only
to diagnose (i), (iii), and (iv).

\subsection{Collapse of GCN+PE and GAT+PE at $K\!=\!4$}
\label{app:K4-collapse}

\begin{figure}[h]
\centering
\includegraphics[width=0.95\linewidth,trim=0 0 0 34bp,clip]{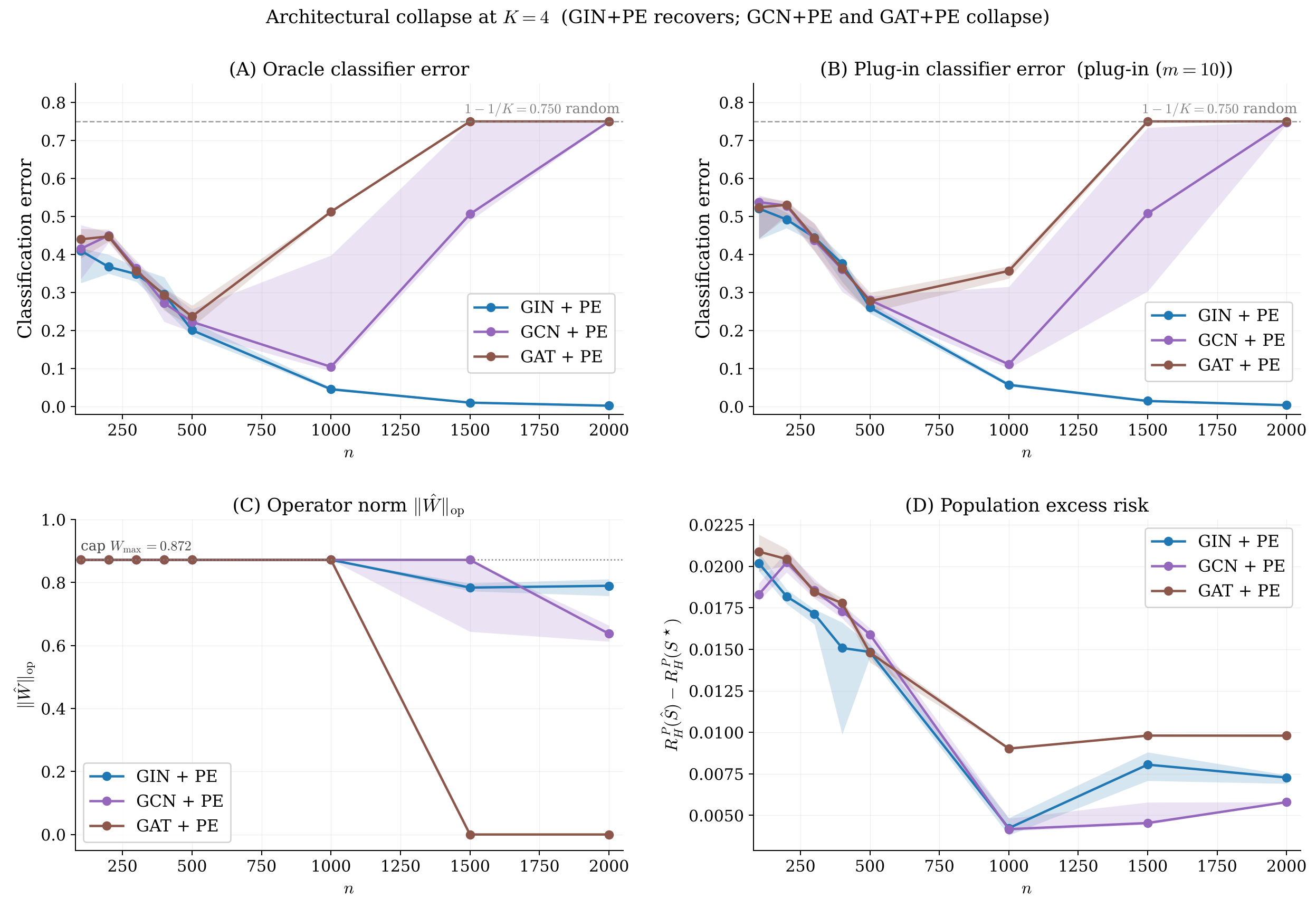}
\caption{\textbf{Collapse at $K\!=\!4$.}
\textbf{(A)} Oracle and \textbf{(B)} plug-in ($m\!=\!10$) classification
error vs $n$ for GIN+PE, GCN+PE, and GAT+PE. GIN+PE recovers; by
$n\!=\!2000$ both GCN+PE (9 of 10 seeds) and GAT+PE (10 of 10 seeds)
collapse to the random baseline $1\!-\!1/K\!=\!0.75$.
\textbf{(C)} Operator norm $\|\widehat W\|_{\op}$ of the trained bilinear
decoder. All three stay at the cap $\Wmax\!=\!0.872$ up to $n\!=\!1000$. At
$n\!=\!2000$, GIN+PE stays close to the cap (median $0.79$), GCN+PE drops
further below it (median $0.64$), and GAT+PE collapses to
$\widehat W\!\approx\!\mathbf 0$, i.e., the model has reverted to the
constant predictor.
\textbf{(D)} Population excess risk. The collapsed GCN+PE runs do not have
a larger excess than GIN+PE.}
\label{fig:K4-collapse}
\end{figure}

\paragraph{A training failure.}
Prop.~\ref{prop:eapp-rate} does not cover GCN+PE and GAT+PE, which change
the aggregator and the PE at the same time
(App.~\ref{app:eapp:discussion}). The collapse is nevertheless not an
approximation limit. Under the second recipe of App.~\ref{app:recipe},
both configurations recover at $K\!=\!4$ (Table~\ref{tab:recipe}), so
their classes contain good solutions that the main recipe does not reach.

\paragraph{Collapse signature.}
We call a run collapsed when its plug-in error exceeds $0.7$. No GCN+PE or
GAT+PE run collapses for $n\!\le\!1000$, although GAT+PE already degrades
at $n\!=\!1000$ (median plug-in error $0.36$, vs $0.06$ for GIN+PE). At $n\!=\!1500$, 4 of 10 GCN+PE
runs and 7 of 9 GAT+PE runs collapse, and at $n\!=\!2000$, 9 of 10 and 10
of 10 (Fig.~\ref{fig:K4-collapse}). For GAT+PE the collapse is complete: $\widehat W\!\approx\!\mathbf 0$
and the embeddings shrink to zero (median of the mean $\|z_i\|$ below
$10^{-3}$), so the only non-trivial output is the bias $b$ and the model is
the constant predictor. GCN+PE shows a softer version at $n\!=\!2000$:
$\widehat W$ falls below the cap (median $0.64$ vs cap $0.87$) without
vanishing, yet the embeddings carry no community information. The failure
lies in the representation, not in our probe: logistic regression and
$k$-means on the same embeddings also stay at chance (errors $0.73$--$0.75$
at $n\!=\!2000$).

\subsection{Markov Factor Tightness in the $\Deltas$ Sweep}\label{app:delta-tight}

\begin{table}[h]
\centering\small
\caption{$\Deltas$ sweep at $K\!=\!3$, GIN+PE: log--log slope $\beta$ of
$\mathrm{err}\!\sim\!\Deltas^\beta$ for $\Deltas\!\in\![0.6,1.9]$, and
coefficient of variation (CV) of the population excess across $\Deltas$,
by $n$. Slopes are least-squares fits of the median oracle classifier error
against $\Deltas$ on a log--log scale, using only the grid points with
nonzero median error (last column); the $n\!=\!1000$ slope therefore rests
on two points. Theory predicts $\beta\!=\!-2$ (the Markov factor); a larger
$|\beta|$ indicates that the Markov bound is loose.}
\label{tab:delta-slopes}
\begin{tabular}{c|ccc}
\toprule
$n$ & slope $\beta$ (err vs $\Deltas$) & CV (excess) across $\Deltas$ & \# points \\
\midrule
$100$  & $-2.19$ & $6.3\%$  & $4$ \\
$200$  & $-4.51$ & $3.0\%$  & $4$ \\
$500$  & $-6.51$ & $14.3\%$ & $3$ \\
$1000$ & $-8.07$ & $10.3\%$ & $2$ \\
\bottomrule
\end{tabular}
\end{table}

The Step~3 factor $\Deltas^{-2}$ is close to tight where errors are still
large (Table~\ref{tab:delta-slopes}). At $n\!=\!100$, where the oracle error lies between $0.05$ and $0.38$ at
four of the five grid points, the slope is $-2.19$, close to the predicted
$-2$.
The factor becomes much looser deep inside recovery, with slopes of $-6.51$
at $n\!=\!500$ and $-8.07$ at $n\!=\!1000$; these two fits use only three
and two points, because the error is exactly zero at larger $\Deltas$. A
plausible reason is the Markov inequality
$\indic\{\widetilde e_i^2\!>\!t\}\!\le\!\widetilde e_i^2/t$ used in Step~3.
It is nearly tight when much of the mass of $\widetilde e_i^2$ lies near the
threshold, but loose when $\widetilde e_i^2$ concentrates well below the
threshold and the indicator is usually zero. This describes the empirical
regime deep inside recovery, although we do not formalize the mechanism.
The CV of the population excess stays below $15\%$ at every $n$, which
suggests that $\eapp$ for GIN+PE is approximately decoupled from the
score-profile geometry.

\subsection{Capacity Threshold and the Plateau in the $\Wmax$ Ablation}
\label{app:cap-saddle}

\begin{table}[h]
\centering\small
\caption{$\Wmax$ ablation (GIN+PE, $K\!=\!3$): floor $F_3$ on $\eapp$
(Prop.~\ref{prop:exact-capacity}) and the number of seeds, out of 10, with
oracle error at most $0.05$. Cap-ratios $0.5$ to $0.75$ come from a finer
rerun of the same configuration, and the last column covers cap-ratios
$1.5$, $2$, $5$, and $10$.}
\label{tab:cap-full}
\setlength{\tabcolsep}{4pt}
\begin{tabular}{l|cccccccc}
\toprule
cap-ratio & $0.25$ & $0.5$ & $0.55$ & $0.6$ & $0.65$ & $0.7$ & $0.75$ & $\ge\!1.5$ \\
\midrule
floor $F_3$ & $0.0026$ & $0.0004$ & $0.0002$ & $0.0001$ & $<\!10^{-5}$ & $0$ & $0$ & $0$ \\
recovered, $n\!=\!1000$ & $0$ & $1$ & $2$ & $4$ & $4$ & $8$ & $7$ & $9$--$10$ \\
recovered, $n\!=\!2000$ & $0$ & $0$ & $1$ & $1$ & $2$ & $5$ & $7$ & $10$ \\
\bottomrule
\end{tabular}
\end{table}

For our SBMs, realizability needs cap-ratio $\ge\!2/3$
(Prop.~\ref{prop:exact-capacity}), and \textbf{(A5)} of
Prop.~\ref{prop:eapp-rate} suffices at cap-ratio $1$.

\paragraph{Below the threshold.}
At cap-ratio $0.25$ for every $n$, and at $0.5$ for $n\!\ge\!500$, every
seed ends with $\|\widehat W\|_{\op}\!=\!\Wmax$ and nearly all
$\|z_i\|\!=\!R$. Its population excess equals that of the constant score
$\bhat-\Wmax R^2$ within $10^{-4}$ ($0.0077$ and $0.0070$). The trained bias
stays between $-2.35$ and $-2.15$, so the bilinear term spends its whole
budget lowering every score, and nothing is left for the community
contrast. The best constant predictor, $b\!=\!\logit\bar p\!=\!-2.75$, lies
in the class and has a smaller excess ($0.0066$), so these runs have
$\eopt\!>\!0$. Under the second recipe of App.~\ref{app:recipe}, the same
class reaches $0.0028$ and $0.0014$ at $n\!=\!2000$, close to the floors
$0.0026$ and $0.0004$, so the plateau is mainly $\eopt$. The embeddings
still separate the communities ($k$-means on $z_i$ has error $0.013$ at
cap-ratio $0.25$, $n\!=\!2000$), but the scores barely use this structure.
The oracle classifier therefore stays at chance, and the plug-in error
falls only to $0.10$.

\paragraph{Near the threshold.}
Between cap-ratios $0.55$ and $0.75$, each seed ends in one of three
states: the constant-score plateau above, a partial stall with oracle
error near $0.33$ (two communities merged), or full recovery. The share
that recovers grows with cap-ratio (Table~\ref{tab:cap-full}). Seeds
recover even below $2/3$, where the floor is below $10^{-3}$ and its
minimizer still separates the communities. The transition is thus set by
optimization rather than by capacity. The LP excess separates the states
only weakly. At $n\!=\!2000$, partially stalled seeds have excess
$0.0052$--$0.0060$ and recovered seeds $0.0040$--$0.0057$. The rerun at
cap-ratios $0.5$ and $0.75$ matches the original runs within $0.03$ in
oracle error for $48$ and $45$ of $50$ seeds. The other seeds switch basin
(e.g., cap-ratio $0.5$, $n\!=\!1000$, seed $5$: $0.66$ in the original,
$0.04$ in the rerun), which we attribute to GPU nondeterminism near the
plateau. Cap-ratio $\ge\!1.5$ removes the split.

\subsection{Sparse-Regime Bound Breakdown}\label{app:rho-impossible}

\begin{table}[h]
\centering\small
\caption{$\rho_n$ ablation (GIN+PE, $K\!=\!3$): SBM-derived parameters and
classifier errors at $n\!=\!2000$, where $\bar q$ is the mean edge
probability. The worst-case prefactor $2/\muB$ grows $32\times$ across the
sweep, while $\Deltas$ changes by only $13\%$.}
\label{tab:rho-full}
\begin{tabular}{l|ccccccc}
\toprule
$\rho_n$ & $\Deltas$ & $|\logit\bar q|$ & $B$ & $\muB$ & $2/\muB$ & oracle err & plug-in err \\
\midrule
$2.0$ & $0.862$ & $1.99$ & $4.72$ & $8.79\!\times\!10^{-3}$ & $228$ & $0.000$ & $0.000$ \\
$1.0$ & $0.801$ & $2.75$ & $5.35$ & $4.71\!\times\!10^{-3}$ & $425$ & $0.000$ & $0.000$ \\
$0.5$ & $0.774$ & $3.48$ & $5.72$ & $3.24\!\times\!10^{-3}$ & $616$ & $0.014$ & $0.017$ \\
$0.2$ & $0.758$ & $4.41$ & $6.95$ & $9.54\!\times\!10^{-4}$ & $2096$ & $0.660$ & $0.414$ \\
$0.1$ & $0.753$ & $5.11$ & $7.86$ & $3.86\!\times\!10^{-4}$ & $5175$ & $0.660$ & $0.615$ \\
$0.05$ & $0.751$ & $5.81$ & $8.21$ & $2.72\!\times\!10^{-4}$ & $7350$ & $0.620$ & $0.647$ \\
\bottomrule
\end{tabular}
\end{table}

For $\rho_n\!\le\!0.1$ (Table~\ref{tab:rho-full}), both the oracle and plug-in classifiers fail
(errors $\!\ge\!0.61$), close to the random baseline
$1\!-\!1/K\!=\!0.667$ at $K\!=\!3$. At $\rho_n\!=\!0.2$, the oracle
classifier also fails (error $0.660$), but the plug-in classifier reaches
$0.414$, clearly below random, so the embeddings still carry community
signal. The two classifiers may differ because the oracle uses fixed-scale
population prototypes, which can mismatch the trained embedding scale once
$\widehat W$ shrinks in the sparse regime, whereas the plug-in classifier
uses empirical prototypes that adapt to the trained representation. We do
not isolate this mechanism formally.

Not all of these failures are information-theoretic. With $a\!=\!n\rho_n
p_{\mathrm{in}}$ and $b\!=\!n\rho_n p_{\mathrm{out}}$ at $n\!=\!2000$, the
Kesten--Stigum condition $(a\!-\!b)^2\!>\!K(a\!+\!(K\!-\!1)b)$, above which
efficient partial recovery is possible \citep{abbe2018community}, holds at
$\rho_n\!=\!0.1$ ($a\!=\!20$, $b\!=\!8$: $144\!>\!108$) and fails at
$\rho_n\!=\!0.05$ ($36\!<\!54$). At $\rho_n\!=\!0.1$ it also holds for the
masked encoder input, which keeps $80\%$ of the pairs ($92\!>\!86$). The
failure at $\rho_n\!=\!0.1$ is therefore a failure of our pipeline, not of
the data, while the failure at $\rho_n\!=\!0.05$ is expected, since no
efficient method is known to succeed below this threshold.

Under the second recipe of App.~\ref{app:recipe}, the plug-in error at
$n\!=\!2000$ falls from $0.41$ to $0.20$ at $\rho_n\!=\!0.2$, but only from
$0.62$ to $0.54$ at $\rho_n\!=\!0.1$. The gap at $\rho_n\!=\!0.2$ is thus
partly an optimization failure, while $\rho_n\!\le\!0.1$ stays near chance
under both recipes. There the bound becomes uninformative. Plausible
contributors include the breakdown of $2\!\to\!\infty$ spectral
concentration (Lemma~\ref{app:eapp:lem:A5}) in the sparse regime, where the
dense-regime assumption \textbf{(A1)} of Prop.~\ref{prop:eapp-rate} fails,
and the simultaneous shrinking of $\muB$ in the bound. We do not separate
these contributions.

\subsection{Diagnosing the Training Failures with a Second Recipe}\label{app:recipe}

The main recipe (App.~\ref{app:hyperparams}) fails in two ways. Some runs
complete but stall: the $K\!=\!4$ collapse of App.~\ref{app:K4-collapse},
the plateau of App.~\ref{app:cap-saddle}, the gap at $\rho_n\!=\!0.2$, and
the partial recovery of GIN+random. Other runs stop with a numerical
error (App.~\ref{app:coverage}).

\paragraph{Likely mechanism for stalled runs.}
The bias starts at $0$ while the target mean logit is about $-2.8$, and
early training compresses the embeddings. From this small-scale state, a
bilinear model learns the community directions one at a time, with
plateaus at saddle points in between. Adam with coupled $L_2$ weight decay
adds $\lambda\theta$ to the gradient before normalization. For a direction
not yet learned, where the data gradient is small, the update is then
about $-\mathrm{lr}\cdot\mathrm{sign}(\theta)$, which drives the direction
to zero and makes the saddle absorbing. The gradient reaching the encoder,
$\partial s_{ij}/\partial z_i\!=\!Wz_j$, scales with $\Wmax$, and
normalized aggregation shrinks the community signal at initialization.
This matches the failing cells (small $\Wmax$, GCN, and GAT). We support
the mechanism with the rerun below but do not test each step separately.

\paragraph{Second recipe.}
We change only two settings: AdamW with decoupled weight decay $10^{-4}$
\citep{loshchilov2019decoupled}, and a bias initialized at the empirical
logit of the training edge density, which uses $\Tcal$ only. We rerun all
five sweeps ($2{,}350$ runs). Table~\ref{tab:recipe} shows that every
stalled cell recovers or improves.

\begin{table}[h]
\centering\small
\caption{Main vs.\ second recipe at $n\!=\!2000$: medians over seeds, with
the number of seeds whose plug-in error ($m\!=\!10$) exceeds $0.1$ in
parentheses. GIN+PE at $K\!=\!3$ unless stated.}
\label{tab:recipe}
\setlength{\tabcolsep}{4pt}
\begin{tabular}{ll|cc}
\toprule
cell & quantity & main & second \\
\midrule
GCN+PE, $K\!=\!4$ & plug-in error & $0.748$ ($10$) & $0.007$ ($0$) \\
GAT+PE, $K\!=\!4$ & plug-in error & $0.750$ ($10$) & $0.008$ ($0$) \\
GIN+random, $K\!=\!4$ & plug-in error & $0.238$ ($10$) & $0.032$ ($0$ of $5$) \\
GIN+random, $K\!=\!3$ & plug-in error & $0.028$ ($0$) & $0.005$ ($0$ of $9$) \\
cap-ratio $0.25$ & excess (floor $0.0026$) & $0.0077$ & $0.0028$ \\
cap-ratio $0.5$ & excess (floor $0.0004$) & $0.0070$ & $0.0014$ \\
$\rho_n\!=\!0.2$ & plug-in error & $0.414$ ($10$) & $0.198$ ($10$) \\
$\rho_n\!=\!0.1$ & plug-in error & $0.615$ ($10$) & $0.536$ ($10$) \\
\bottomrule
\end{tabular}
\end{table}

\paragraph{Runs that stop with an error.}
Under the second recipe, $23$ of the $2{,}350$ runs still stop, all
because \texttt{torch.linalg.eigh} fails to converge. Eigendecompositions
are used both for the spectral PE and for clipping the eigenvalues of $W$.
Of these runs, $13$ are GIN+random, whose three restarts must all succeed.
The five failing seeds at $\rho_n\!=\!0.05$, $n\!=\!1000$ are the same
under both recipes, so these failures depend on the graph rather than on
the optimizer.

\paragraph{Why the main text keeps the main recipe.}
The second recipe overfits at large $n$. At $K\!=\!3$, the excess of GCN+PE
decays with slope $-0.27$ instead of $-1.12$ ($0.0063$ vs.\ $0.0011$ at
$n\!=\!2000$), and that of GIN+PE with slope $-0.45$ instead of $-0.67$.
The classification error at $n\!=\!2000$ is unchanged. Coupled weight
decay thus both creates the absorbing saddles and regularizes the trained
scores. We therefore report the main recipe in Sec.~\ref{sec:experiments}
and use the second one only to diagnose its failures.
\section{Rate of the Guarantee and Classical SBM Recovery}\label{app:rate}

Theorem~\ref{thm:main} controls the classification error through the
held-out LP excess risk, an average over node pairs, and gives the rate
$\Otilde(n^{-1/2})$ in the dense regime. SBM-specific procedures reach
$\exp(-\Omega(n))$ in the same regime
\citep{zhang2016minimax,gao2017achieving}. This appendix shows that the
linear dependence on an average risk cannot be improved, explains what
faster rates would require, and compares the guarantee with classical
recovery thresholds.

\paragraph{An average risk cannot localize errors.}
Assume the bilinear capacity condition, so that there are community
embeddings $u_1,\dots,u_K$ with $\|u_c\|\le R$ and an admissible
$(W^\star,b^\star)$ with $b^\star+u_c^\top W^\star u_{c'}=m^\star_{cc'}$
for all $c,c'$ (Prop.~\ref{prop:bilinear-realizability}). Let communities
$a\ne b$ attain $\Deltas$ and set $\delta_c:=m^\star_{bc}-m^\star_{ac}$, so
that $\sum_c\pi_c\delta_c^2=\Deltas^2$. The embeddings $z_i=u_{c_i}$ give
$S=\Sopt$. Now move a set $\mathcal M\subset\mathcal C_a$ of $\alpha n$
nodes to $z_i=u_b$. The score error $s_{ij}-m^\star_{c_ic_j}$ is
$\delta_{c_j}$ if only $i\in\mathcal M$, $\delta_{c_i}$ if only
$j\in\mathcal M$, $m^\star_{bb}-m^\star_{aa}$ if both are in $\mathcal M$,
and $0$ otherwise. Hence, for $\alpha<\alpha_0:=\Deltas^2/(16B^{\star2})$,
\[
\mathrm{ScoreMSE}_{\mathrm{full}}
=2\alpha\Deltas^2-2\alpha^2\delta_a^2+\alpha^2(m^\star_{bb}-m^\star_{aa})^2
\in\bigl[\tfrac32,\tfrac94\bigr]\,\alpha\Deltas^2 .
\]
In score-profile space, a moved node lies within $2B^\star\sqrt{\alpha n}$
of $\mu_b$, and every other node $i$ lies within $2B^\star\sqrt{\alpha n}$
of $\mu_{c_i}$. For $\alpha<\alpha_0$ this distance is below
$\sqrt n\,\Deltas/2$, half the minimum prototype separation
(App.~\ref{app:step3}). The oracle classifier therefore errs exactly on
$\mathcal M$, so the error is
$\alpha\ge\frac49\,\mathrm{ScoreMSE}_{\mathrm{full}}/\Deltas^2$, within a
factor $9$ of Prop.~\ref{prop:step3}. All scores of this model lie in
$[-B^\star,B^\star]$, so its held-out excess risk lies between
$\frac12\mu_{B^\star}\mathrm{ScoreMSE}_\Hcal$ and
$\frac18\mathrm{ScoreMSE}_\Hcal$ (Lemma~\ref{lem:step1} and
App.~\ref{app:eapp:master}). Since $S$ is fixed,
$\mathrm{ScoreMSE}_\Hcal$ differs from $\mathrm{ScoreMSE}_{\mathrm{full}}$
by $\Otilde(B^{\star2}/(\sqrt\tau\,n))$ with probability at least
$1-n^{-1}$ (Hoeffding--Serfling, as in App.~\ref{app:step2}). With $\alpha=n^{-\gamma}$ and $\gamma\in(0,1)$, the
excess risk is of order $\Deltas^2n^{-\gamma}$ and the error is
$n^{-\gamma}$.

Theorem~\ref{thm:main} covers every encoder family satisfying
Assumption~\ref{ass:bounded}, including families that can output these
embeddings. Any error bound that depends on the model only through its
average LP risk must therefore be at least linear in that risk, with the
factor $\Deltas^{-2}$. In this sense, the linear dependence in
Theorem~\ref{thm:main} is worst-case tight. The certified rate is then the rate at which this risk
decays, which is polynomial: even with known communities, the $K\times K$
block logits are estimated from $\Theta(n^2)$ edges, with squared error of
order $n^{-2}$. The exponent $1/2$ in Theorem~\ref{thm:main} comes from
$\Vstat$.

\paragraph{Faster rates need node-level control.}
A node is misclassified only if $\widetilde e_i\ge\Deltas/2$
(App.~\ref{app:step3}), so the error is zero once
$\max_i\widetilde e_i<\Deltas/2$. A bound on $\max_i\widetilde e_i$, rather
than on the average $n^{-1}\sum_i\widetilde e_i^2=\mathrm{ScoreMSE}_{\mathrm{full}}$,
would give exact recovery as soon as the risk falls below a fixed
threshold. The class contains models with such control. On the Step~A
event of App.~\ref{app:eapp}, which has probability at least $1-n^{-c}$,
the witness of Prop.~\ref{prop:eapp-rate} satisfies
$\max_{i,j}|s_{ij}-m^\star_{c_ic_j}|\le\delta_n=O(\sqrt{\log n/n})$
(Lemma~\ref{app:eapp:lem:master}). Hence $\widetilde e_i\le\delta_n$ for
every $i$, and its oracle classifier makes no error once
$\delta_n<\Deltas/2$. Proving this for a trained model requires fixing both
the architecture and the training algorithm, and showing that optimization
reaches parameters with node-level control. Fixing the architecture alone
does not suffice: Prop.~\ref{prop:eapp-rate} fixes GIN with spectral PE but
says nothing about the parameters that training reaches. The result would
then be an algorithm-specific SBM recovery analysis. We keep
Theorem~\ref{thm:main} at the level of the average risk so that it applies
to any bounded encoder--decoder and any training procedure.

The observed error decays faster than the bound. At $n\!=\!2000$, the
median oracle error of GIN+PE is $0$, while
$\mathrm{ScoreMSE}_{\mathrm{full}}\!=\!0.144$ (Tab.~\ref{tab:scoremse-compare}),
for which $4\,\mathrm{ScoreMSE}_{\mathrm{full}}/\Deltas^2\!=\!0.90$.
Misclassification depends on the upper tail of $\widetilde e_i$, while
Markov's inequality in Step~3 uses only its mean
(App.~\ref{app:delta-tight}).

\paragraph{Classical recovery thresholds.}
For the balanced symmetric $K$-block SBM with fixed $p>q$,
$\Deltas^2=\frac2K(\logit p-\logit q)^2>0$, and (A1)--(A3) of
Prop.~\ref{prop:eapp-rate} hold. Under the capacity condition and the
training antecedent in the Scope paragraph of Sec.~\ref{sec:eapp}, the error
is $\Otilde(\Deltas^{-2}\sqrt{d/n})\to0$ for every fixed $p>q$, so the
guarantee gives weak consistency. An error fraction of order $n^{-1/2}$
still allows $\Otilde(\sqrt n)$ misclassified nodes, so it does not give
exact recovery, which requires a fraction below $1/n$. In the
logarithmic-degree regime, with $p=a\log n/n$ and $q=b\log n/n$, exact
recovery of two communities has the sharp threshold $\sqrt a-\sqrt b=\sqrt2$
\citep{abbe2016exact,mossel2015consistency}. There
$B\ge B^\star=(1+o(1))\log n$, so $\muB=O(\log n/n)$, and the bound is
informative only if $\eopt+\eapp=O(\log n/n)$, the order of the excess risk
of a constant score. Our realization of $\eapp$ also assumes the dense
regime (A1). The reduction therefore gives no sparse threshold, and deriving
one would require analyzing when a specified LP procedure attains
node-level score accuracy at $p_{ab}=\Theta(\log n/n)$.
App.~\ref{app:rho-impossible} shows that our pipeline can fail where
efficient partial recovery is possible.

\end{document}